\documentclass[aps,prx,amsmath,amssymb,twocolumn,superscriptaddress,showpacs,longbibliography]{revtex4-2}
\usepackage{bm,mathrsfs}
\usepackage{graphicx}
\usepackage{appendix}
\usepackage{epsfig}
\usepackage{amsmath,bbm}
\usepackage{amsfonts,amssymb}
\usepackage{times}
\usepackage{dsfont}
\usepackage{enumitem}  
\usepackage{comment}
\usepackage{url}
\usepackage[colorlinks,linkcolor=blue,citecolor=blue]{hyperref}
\usepackage{tikz}
\usepackage{here}
\usetikzlibrary{arrows.meta,positioning,calc}
\usepackage{algpseudocode}

\newtheorem{theorem}{Theorem}[section]
\newtheorem{definition}{Definition}[section]
\newtheorem{assumption}{Assumption}[section]
\newtheorem{proposition}[theorem]{Proposition}
\newtheorem{lemma}[theorem]{Lemma}
\newtheorem{corollary}[theorem]{Corollary}

\newenvironment{proof}[1][Proof.]
  {\par\noindent\textit{#1}\ }
  {\hfill$\square$\\\par}

\newcounter{algorithm}
\renewcommand{\thealgorithm}{\arabic{algorithm}}

\newenvironment{revtexalgorithm}[1][t]
{
    \begin{figure}[#1]
    \refstepcounter{algorithm}
    \begin{minipage}{\columnwidth}
}
{
    \end{minipage}
    \end{figure}
}

\newcommand{\algorithmcaption}[1]{
    \vspace{0.5em}
    \noindent\textbf{Algorithm \thealgorithm.} #1
    \par\vspace{0.5em}
}

\begin{document}

\title{Renormalization Group Flow Matching for Scalable Local Generative Modeling}

\author{Kanta Masuki}
\email{masuki@g.ecc.u-tokyo.ac.jp}
\affiliation{Department of Physics, The University of Tokyo, 7-3-1 Hongo, Bunkyo-ku, Tokyo 113-0033, Japan}

\author{Yuto Ashida}
\email{ashida@phys.s.u-tokyo.ac.jp}
\affiliation{Department of Physics, The University of Tokyo, 7-3-1 Hongo, Bunkyo-ku, Tokyo 113-0033, Japan}
\affiliation{Institute for Physics of Intelligence, The University of Tokyo, 7-3-1 Hongo, Bunkyo-ku, Tokyo 113-0033, Japan}

\date{\today}

\begin{abstract}
Despite their remarkable success in modeling complex data, generative models face a fundamental tradeoff. Global approaches can capture full structural coherence but suffer from high computational costs, while local models are efficient but often fail to reproduce long-range correlations and global coherence. The renormalization group (RG) bridges this gap by seamlessly connecting spatial structures across different length scales, retaining quasi-local descriptions at each step without sacrificing long-range correlations. We introduce renormalization group flow matching (RGFM), a generative framework that systematically structures data generation across different spatial scales. By using an exact RG flow as the probability path in flow matching, RGFM generates data progressively from long- to short-wavelength structures. To reconcile scalability with global structure, we exploit two key properties of the RG: quasi-locality and scale separation. We rigorously show that the RGFM probability flow can be accurately approximated by local velocity fields acting over a spatial range $O\left(\Lambda^{-1}[\ln L+\ln(1/\varepsilon)]\right)$ for RG wavenumber scale $\Lambda$, linear system size $L$, and prescribed error tolerance $\varepsilon$. This property enables local generative modeling with patches of size $O(\ln L)$ and a computational cost that scales nearly linearly with the system volume. We numerically demonstrate that local RGFM reproduces long-range correlations far beyond its receptive field in representative one-dimensional distributions, while conventional local flow matching exhibits substantial errors at long distances. On FFHQ images, RGFM substantially improves global coherence and lowers FID at $64\times64$ resolution and produces higher-quality samples than local flow matching at $256\times256$ resolution. Our results establish RG-guided probability flows as a promising route toward scalable generative modeling that captures long-range structure using only local computation.
\end{abstract}

\maketitle

\section{\label{sec:intro}Introduction}

\begin{figure*}[t]
    \centering
    \includegraphics[width=17.9cm]{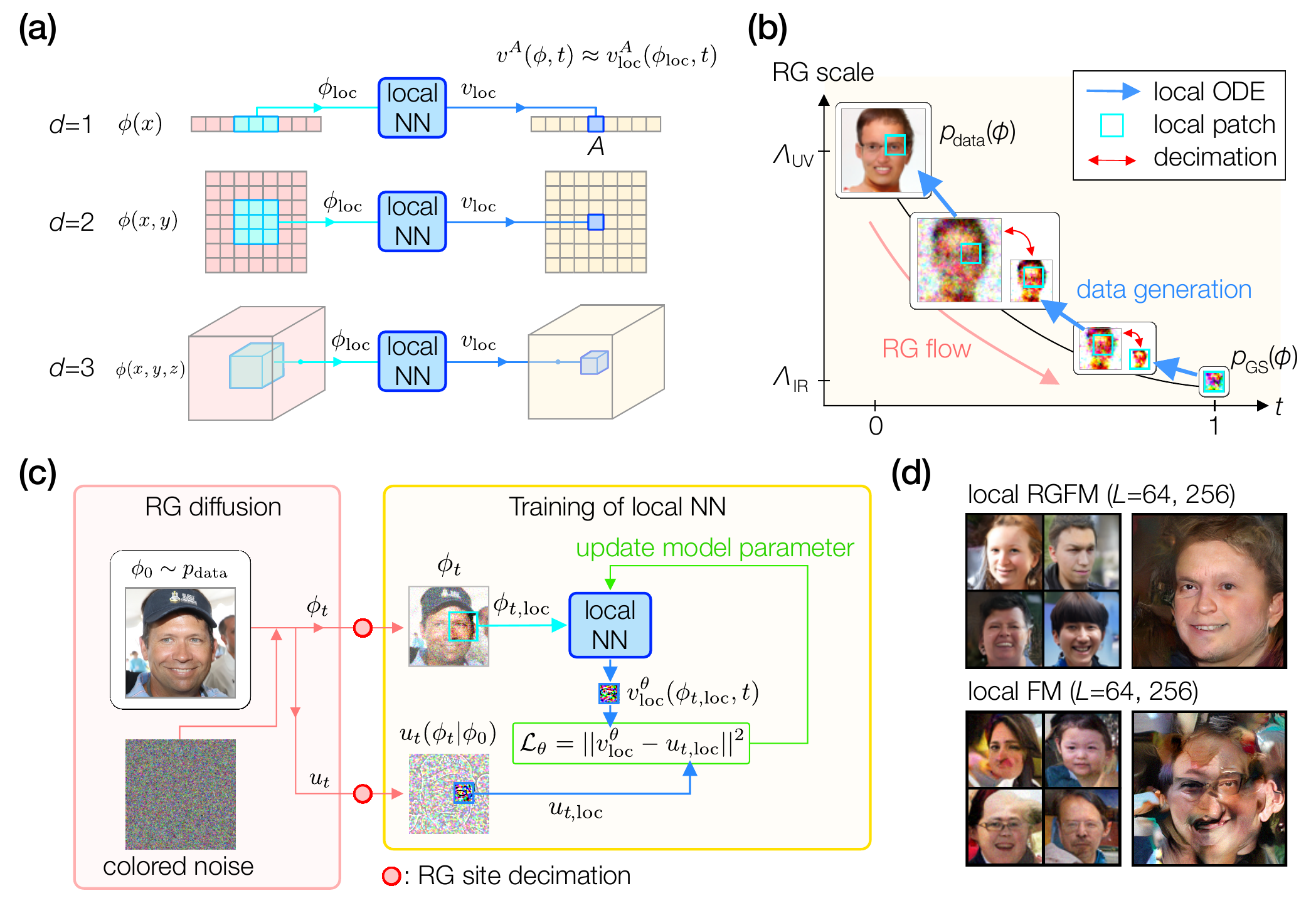}
    \caption{\label{fig:intro}
    (a) Local generative modeling. Exploiting the underlying spatial structure of the data, the velocity field in a target region $A$ (shaded in blue) is approximated using only the configuration $\phi_{\rm loc}$ in its surrounding neighborhood. Data are generated by integrating the ODE flow defined by this locally approximated velocity field, which is learned using a local neural network (NN). 
    (b) Schematic of the probability path used for local generative modeling with renormalization group flow matching (RGFM). The RG wavenumber scale $\Lambda(t)$ is related to the logarithmic time $t$ by $\Lambda(t)=\Lambda_{\rm UV}e^{-t/\tau}$ with $\tau\ll 1$ being a characteristic time scale specified later. Along the forward path from $t=0$ to $t=1$, the coarse-graining diffusion process, which we call the RG diffusion, is alternated with site-decimation transformations. The RG diffusion can be reversed by solving the ODE generated by a local velocity field, while each site-decimation step can be reversed at the level of probability distributions by resampling the discarded Gaussian modes. The entire path can therefore be reversed from $t=1$ to $t=0$ using only local patches.
    (c) Training scheme for local RGFM. Applying the RG diffusion to a data configuration $\phi_0\sim p_{\rm data}$ yields a noisy configuration $\phi_t$ and the corresponding conditional velocity field $u_t(\phi_t|\phi_0)$. After transforming $\phi_t$ and $u_t$ according to the site-decimation step in panel (b), a local velocity field $v_{{\rm loc},\theta}$ is trained by flow matching on local patches.
    (d) Representative $64\times64$ and $256\times256$ image samples generated by local RGFM and the standard local FM. While the local model based on the standard FM fails to generate globally consistent samples, the local RGFM produces substantially more globally coherent samples.
    }
\end{figure*}

\subsection{Background}

Diffusion-based generative models have emerged as a powerful framework for learning complex high-dimensional distributions \cite{sohl-dickstein2015a,song2019,ho2020b,song2021a,song2021b} and have achieved remarkable success across a broad range of generative tasks \cite{yang2023b,cao2024}. Their established applications now extend beyond computer vision \cite{dhariwal2021a,croitoru2023} to audio \cite{kong2021,huang2023}, video \cite{ho2022b,blattmann2023}, and point-cloud generation \cite{luo2021,zhou2021,zeng2022}, as well as the prediction of three-dimensional molecular \cite{jo2022,hoogeboom2022d,cornet2024} and protein structures \cite{jing2023a,abramson2024a}. A key idea underlying these models is to construct a probability path between a complex data distribution and a simple noise distribution. In diffusion models, a forward stochastic process gradually transforms the data distribution into white Gaussian noise, while samples are generated by reversing these dynamics through incremental denoising \cite{sohl-dickstein2015a}. By decomposing the transformation from noise to data into a sequence of simple and gradual denoising steps, diffusion models can generate high-dimensional structured data with remarkable fidelity \cite{song2019,ho2020b,song2021a,song2021b}.

Flow matching (FM) \cite{lipman2023} provides a framework for continuous-time generative modeling that is closely related to diffusion models. In FM and similar flow-based generative models \cite{lipman2023,albergo2023,liu2023a}, a time-dependent velocity field is learned to transport a simple reference distribution to the target data distribution through an ordinary differential equation. While sharing the noise-to-data evolution of diffusion models, FM distinctively drives generative dynamics via a deterministic velocity field, allowing a flexible choice of the probability path connecting the reference and data distributions \cite{lipman2023,albergo2023,liu2023a,pooladian2023,tong2024}. This flexibility broadens the design space of continuous generative dynamics and facilitates the development of flows suited to different modeling objectives. Leveraging this versatility, flow-based models have emerged as a promising framework for extending generative modeling to increasingly large and structured data, including high-resolution images \cite{rombach2022a,peebles2023}, videos \cite{lu2024,ma2025}, and 3D data \cite{xiang2025}, as well as to emerging applications involving complex physical systems such as biomolecular systems \cite{morehead2026,li2026}, fluids \cite{wang2025,oommen2026,li2026a}, and statistical and quantum many-body systems \cite{wang2024,singha2025,tan2026,hu2026a}.

A central challenge for expanding the applications of diffusion or flow-based models is the rapid growth of computational cost with the data dimension \cite{rombach2022a,ma2025a,shen2025}. In general, these models employ a neural network (NN) that takes a data configuration $\phi\in\mathbb{R}^D$ as input and predicts a $D$-dimensional score or velocity field as a function of $\phi$. This network must be evaluated repeatedly during both training and sample generation. Thus, expressive neural architectures, which are necessary for accurately modeling complex high-dimensional distributions, generally lead to substantial computational and memory costs. A representative example is self-attention \cite{vaswani2017}, which is widely used in modern generative models and has computational and memory costs that scale quadratically with the number of input tokens. When the number of tokens is proportional to the data dimension $D$, these costs scale as $O(D^2)$. Since $D$ becomes extremely large for data such as high-resolution images and many-body physical systems, straightforward training becomes infeasible without substantial computational resources. For example, training high-performance pixel-space diffusion models has been reported to require hundreds of GPU days, reaching approximately $150$--$1000$ V100 GPU days for large-scale image-generation models \cite{rombach2022a}. Such computational demands present a major obstacle to scaling conventional diffusion and flow-based generative models to increasingly high-dimensional data.

One direct strategy for reducing this computational cost is local generative modeling \cite{wang2023a,ding2024,bieder2024,kamb2025,niedoba2025}, which exploits the underlying spatial structure of natural data. When the components of $\phi$ are associated with sites in a spatial domain of size $L^d$, local generative modeling approximates the score or velocity field in a target region using only the configuration in its surrounding neighborhood, as illustrated in Fig.~\ref{fig:intro}(a). Combining these local predictions across patches covering the full spatial domain yields a local approximation to the full score or velocity field that can be used for sample generation. Since the computational cost of each local prediction is determined by the patch size rather than by the total data dimension $D$, the total cost of evaluating the field over the entire domain scales only linearly with $D$ when the patch size is fixed. Even when the required patch size grows logarithmically with the system size, the overall scaling remains near-linear up to polylogarithmic factors, representing a qualitative improvement over the quadratic scaling of the conventional self-attention discussed above.

A fundamental limitation of local generative modeling, however, is that a purely local model easily fails to construct globally consistent structures \cite{wang2023a,ding2024}. Because each prediction depends only on a fixed neighborhood, the model cannot coordinate distant regions to produce the long-range correlations and semantic features required for global coherence. Recent detailed analyses of conventional diffusion and FM paths further suggest that this difficulty is concentrated in an intermediate transition-like regime \cite{raya2024,biroli2024,li2024,hu2026,zhang2026}: local approximations may remain effective near the data and Gaussian endpoints, whereas the score or velocity field develops essentially nonlocal dependence at intermediate times. In this regime, a purely local FM typically produces low-quality samples that lack global consistency \cite{wang2023a,hu2026}; see, e.g., bottom images of Fig.~\ref{fig:intro}(d). Accordingly, considerable effort has been devoted to developing additional mechanisms that communicate information or enforce consistency across distant patches while retaining the computational advantages of patchwise modeling \cite{bar-tal2023,lee2023,zheng2024,du2024,skorokhodov2024,ding2025}.

The limitation described above can be viewed as a problem of connecting distinct scales; each prediction is made from a local neighborhood, yet the combined predictions must organize structures extending across the entire system. This problem has a close conceptual parallel with the renormalization group (RG) \cite{wilson1975a,wilson1983} developed in theoretical physics. Starting from a microscopic description, the RG progressively coarse-grains short-distance degrees of freedom and constructs a sequence of effective descriptions at increasingly long distances. In particular, along the coarse-graining flow, the RG retains the information needed to describe the remaining large-scale structure. This systematic connection between microscopic and macroscopic descriptions has made the RG a foundational framework in many areas of physics, such as high-energy physics \cite{gross1973a,politzer1973a}, statistical and condensed matter physics \cite{kadanoff1966a,fisher1974a,shankar1994,schollwock2005a,bulla2008a,metzner2012a}, and biological and nonequilibrium physics \cite{hohenberg1977,vicsek1995,toner1995,toner1998,cavagna2023}.

Importantly, such an RG perspective is also relevant to computational science because structured data often contain both local details and global organization across a broad range of length scales \cite{ruderman1994a,vanderschaaf1996a,saremi2013a}. In a natural image, for example, fine textures and edges appear alongside object-scale features and the overall composition. Treating such multiscale structure through a hierarchy of scale-dependent descriptions is closely aligned with the scale separation inherent in the RG, which organizes fluctuations by their characteristic length scales. One related application is multiscale generative modeling, in which large-scale features are generated first and then progressively refined through learned inverse coarse-graining transformations. By decomposing the generative process across spatial scales, these approaches can capture long-range dependencies and local details more effectively, thus improving generative performance \cite{guth2022a,ho2022c,kadkhodaie2023,daras2023,saharia2023b,marchand2023,ren2023,gerdes2024,masuki2025b,lempereur2026,mukhopadhyay2026,chen2026}.

To turn this RG picture of scale-by-scale data generation into a scalable local generative framework, we exploit another key property of the RG, that is, quasi-locality. When the microscopic description contains only local interactions, coarse-graining up to a length scale $\ell$ generally produces an effective description whose interactions remain concentrated within distances of order $\ell$ \cite{kopper2007,kopper2007a,rosten2012}. Equivalently, at a running wavenumber scale $\Lambda\sim\ell^{-1}$, each stage of the RG is governed primarily by spatial neighborhoods of linear size $O(\Lambda^{-1})$. Iterating these locally controlled stages then communicates structures across progressively longer distances. In physics, this quasi-local property of the RG has been essential because it allows the effective description at each scale to be organized systematically in terms of local interactions \cite{hasenfratz1986} and their spatial derivatives \cite{morris1994b}, keeping the coarse-grained theory both physically interpretable and tractable \cite{wilson1975a,polchinski1984a,morris1994a,kopietz2010,rosten2012}.

In this work, we leverage the scale separation and quasi-locality of the RG to introduce renormalization group flow matching (RGFM), in which an exact RG flow defines the continuous probability path of the FM framework. Along this path, short-wavelength modes eliminated by coarse-graining form a simple Gaussian sector, while the remaining long-wavelength modes retain the nontrivial structure of the data. We successively rescale the lattice so that the characteristic length scale remains of order unity in the rescaled lattice units, independently of the system size. This allows the remaining modes to evolve with a local flow-matching velocity field at every stage. During generation, each rescaling step can be reversed probabilistically by restoring the discarded modes from their known Gaussian distribution. The RGFM thus provides a scalable and continuous coarse-to-fine generative process that combines local computation with the ability to construct correlations far beyond the receptive field of the local model.

From a broader perspective, the relation between RG and machine learning has been investigated in several areas beyond multiscale generative modeling. For instance, neural networks have been used to identify informative coarse-grained variables and to learn real-space RG transformations directly from data \cite{beny2013,mehta2014,lin2017,iso2018,li2018a,efthymiou2019,hu2020}. Conversely, parallels between successive coarse-graining and hierarchical feature extraction have provided a framework for interpreting representations learned by deep neural networks \cite{koch-janusz2018a,lenggenhager2020,chung2021,gokmen2021}. More recently, the interplay between RG and machine learning has expanded to a broader range of computational frameworks, including RG-informed generative methods and neural-network-based approaches to solving functional RG flows \cite{yokota2024,ihssen2025,tan2026a}. Taken together, these developments reflect a broader synergy in which machine learning offers flexible tools for discovering transformations across scales, while the RG provides a conceptual language for understanding hierarchical structure in learning systems. By employing the RG as a design principle for controlling the locality of a probability flow, our work reveals a further connection between the RG and machine learning.

\subsection{Summary of the main results}
Before presenting the details, we here provide a nontechnical summary of the main results. We first formulate the RGFM by taking an exact RG flow as the probability path of the FM. Along the forward RG diffusion, fluctuations are successively transformed into Gaussian noise in order of decreasing wavenumber. At a given RG scale $\Lambda(t)$, the resulting distribution separates into a nontrivial effective distribution for modes below $\Lambda(t)$ and an independent Gaussian sector for the modes that have already been integrated out. We note that the terminal distribution approaches a Gaussian distribution once the RG scale is lowered below the infrared cutoff $\Lambda_{\rm IR}$, which scales with the linear system size $L$ as $\Lambda_{\rm IR}\!\propto\! 1/L$. Reversing the probability flow with a flow-matching ODE, we obtain a generative process that constructs data systematically from coarse to fine scales.

We next augment this direct RGFM path with successive lattice rescalings to enable scalable {\it local} generative modeling (Fig.~\ref{fig:intro}(b)). Along the direct RGFM path, the spatial range required to approximate the flow-matching velocity field would naively grow with the running length scale $\Lambda(t)^{-1}$ and eventually become comparable to the linear system size $L$. To prevent this growth, we successively discard the decoupled Gaussian high-wavenumber modes and represent the remaining low-wavenumber modes on a decimated lattice. Each decimation is designed to restore the locality length to a microscopic scale in the new lattice units. While this transformation is not invertible at the level of an individual configuration, it has a probabilistic inverse due to the scale separation of the RG; the discarded modes can be sampled independently from their known Gaussian distribution. Consequently, the probability path can be reversed by alternating local reverse-ODE evolution with stochastic lifting to a finer lattice. This rescaled probability flow enables local generative modeling with the RGFM, where the ODE flow can be learned and reversed using only local patches of linear size $O(\ln L)$ (see Fig.~\ref{fig:intro}(c)).

To provide a theoretical foundation for local generative modeling with the RGFM and clarify its range of applicability, we analyze the conditions under which the RGFM probability path can be accurately approximated using local velocity fields. Specifically, we consider two classes of distributions that we expect to encompass a broad range of structured multiscale data: local distributions with short-range correlations and conditionally local distributions whose long-range dependencies are mediated only by latent variables. Under physically motivated assumptions such as the quasi-locality of RG-evolved interactions, we prove our main theoretical result: along the RGFM path without lattice rescaling, the locality length required for accurate local approximations of the probability flow scales linearly with the running RG length scale $\Lambda(t)^{-1}$. This scaling provides a mathematical basis for our practical strategy of representing the remaining modes on successively coarser lattices to keep the relevant locality length at $O(L^0)$ throughout the flow. More specifically, we prove the following statement:
\\
\\
\noindent\textit{Locality theorem (informal).---}
{\it For a physical target distribution $p_{\rm data}$ and any prescribed error tolerance $\varepsilon>0$, the original RGFM path can be approximated by a flow generated by local velocity fields with buffer widths $l_{B,t}$ that scale as}
\begin{align}
    l_{B,t}
    =
    O\left(
        \Lambda(t)^{-1}
        \left[\ln L+\ln(1/\varepsilon)\right]
    \right),
\end{align}
{\it where the $2$-Wasserstein distance between the target $p_{\rm data}$ and the distribution $p_{\rm data}^{\rm loc}$ generated by the local RGFM is bounded as}
\begin{align}
    W_2\left(p_{\rm data},p_{\rm data}^{\rm loc}\right)
    \leq \varepsilon.
\end{align}

To prove this result, we first define the local approximation error of the RGFM velocity field and show that it decays exponentially with $\Lambda(t)l_B$. We then derive a general stability bound that controls the $2$-Wasserstein distance between the exact probability flow and its locally approximated counterpart in terms of the time-integrated velocity-field approximation error. 

Finally, we numerically compare the local RGFM with a conventional FM under the same local-network constraints. For a one-dimensional Ising model, the local RGFM reproduces correlation functions over distances far beyond the receptive field of the neural network, whereas the local FM exhibits substantial errors at long distances. For a conditionally local distribution with nondecaying oscillatory correlations, the local RGFM reconstructs both the latent global waveform and the local fluctuations, while the local FM fails to maintain global coherence. We also compare local RGFM and local FM for image generation; typical samples are shown in Fig.~\ref{fig:intro}(d). On FFHQ images at $64\times64$ resolution, the local RGFM generates images with substantially more coherent facial structure and maintains substantially lower FID scores over the range of receptive patch sizes considered. The local formulation also makes training feasible at $256\times256$ resolution without evaluating a global network on the entire image; at this resolution, the RGFM produces much more coherent samples than the local FM, although it still exhibits inconsistencies between spatially separated facial components. In the Discussion, we outline how such inconsistencies may be mitigated by conditioning the local RGFM on precomputed global latent representations or by combining it with a low-cost model that infers the relevant latent variables.

The remainder of this paper is organized as follows. In Sec.~\ref{sec:rgfm}, we review the standard FM and formulate the RGFM using the exact RG probability path. In Sec.~\ref{sec:lgm_with_rgfm}, we construct the rescaled RGFM path and present the local training and sampling procedures. In Sec.~\ref{sec:loc_app}, we develop the theoretical framework for local approximation and establish error bounds for the velocity field and the resulting probability flow. Section~\ref{sec:num_exp} presents numerical experiments on local and conditionally local one-dimensional distributions and on image generation. In Sec.~\ref{sec:discussions}, we discuss the implications, limitations, and future directions of the proposed framework. Technical details and extensions of the theoretical results are provided in the Appendices.

\section{\label{sec:rgfm}Renormalization group flow matching}

\begin{figure*}
    \centering
    \includegraphics[width=17.9cm]{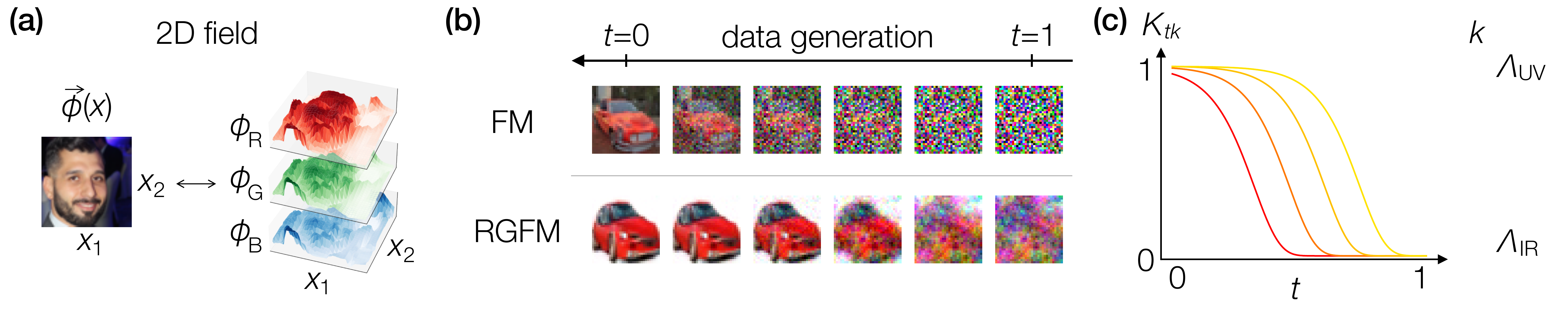}
    \caption{\label{fig:fm_rgfm}
    (a) Correspondence between image data and an $\mathbb R^3$-valued vector field $\vec{\phi} = (\phi_R, \phi_G, \phi_B)$ on the two-dimensional pixel lattice, where $\vec{\phi}(x)$ represents the RGB value of the pixel at position $x$.
    (b) Data generation in the standard flow matching (FM) and the renormalization group flow matching (RGFM).
    (c) Behavior of the RG cutoff function $K_{tk}$ in the Polchinski RG flow~\eqref{eq:pt_diff}.    
    }
\end{figure*}

A wide range of natural data can be viewed as spatially structured field configurations. For example, as shown in Fig.~\ref{fig:fm_rgfm}(a), an RGB image $\phi = (\phi^R, \phi^G, \phi^B)$ can be regarded as an $\mathbb R^3$-valued field configuration on a two-dimensional pixel lattice. Such data typically exhibit multiscale structure, suggesting a generative process that proceeds progressively from coarse features to fine details (i.e., low-to-high wavenumber modes). In this section, we first review the framework of flow matching (FM). We then introduce renormalization group flow matching (RGFM), in which a conditional probability path derived from the exact RG flow enables systematic data generation in a coarse-to-fine manner.

\subsection{Flow matching}

For concreteness, let $\phi$ be a real scalar or vector field defined either on the $d$-dimensional box $[0,L]^d \subset \mathbb R^d$ or on the $d$-dimensional discrete lattice $\{0,1,\ldots,L-1\}^d \subset \mathbb Z^d$. In either case, we denote the underlying spatial domain by $\Omega_L$ and the value of the field at $x\in\Omega_L$ by $\phi_x$. In the continuum, we use $\int_x$ and $\delta/\delta\phi_x$ to denote integration over $\Omega_L$ and the functional derivative with respect to $\phi_x$, respectively. On the discrete lattice, $\int_x$ and $\delta/\delta\phi_x$ are understood as the sum $\sum_{x\in\Omega_L}$ and the derivative $\partial/\partial\phi_x$, respectively.

We define a time-dependent vector field $v(\phi,t)$, referred to as the velocity field, and consider the corresponding deterministic evolution
\begin{align}
    \frac{d\phi_t}{dt} = v(\phi_t,t).\label{eq:fm_ode}
\end{align}
The ordinary differential equation (ODE)~\eqref{eq:fm_ode} induces an evolution of the probability distribution $p_t(\phi)$, governed by the continuity equation
\begin{align}
    \partial_t p_t(\phi) + \frac{\delta}{\delta\phi}\cdot (p_t(\phi)v(\phi, t)) = 0.\label{eq:continuity_eq}
\end{align}
Here, $\delta/\delta\phi \cdot (p_tv)$ denotes $\int_x\, (\delta/\delta\phi_x) [p_t(\phi)v_x(\phi,t)]$, where $v_x(\phi,t)$ is the component of the velocity field at $x\in\Omega_L$. The flow~\eqref{eq:fm_ode} defines an invertible map between $\phi_{t_0}$ and $\phi_{t_1}$. Its trajectories $\{\phi_t\}_{t\in[t_0,t_1]}$ can be followed either forward or backward in time by integrating the ODE with the corresponding orientation of the time interval. Consequently, the probability flow $p_t$ generated by the velocity field $v(\phi,t)$ can also be traced in both time directions.

The FM \cite{lipman2023,albergo2023,liu2023a} provides a framework for generating data $\phi\sim p_{\rm data}$ through the ODE flow~\eqref{eq:fm_ode}. To this end, one constructs a velocity field $v(\phi,t)$ and a probability flow $p_t$ that connect the data distribution $p_{\rm data}(\phi)$ to a simple reference distribution $q(\phi)$, such as the standard normal distribution $\mathcal N(0,I)$. Specifically, one chooses $v(\phi_t,t)$ and $p_t$ so that $p_{t=0}\approx p_{\rm data}$ and $p_{t=1}\approx q$. By learning the velocity field $v(\phi,t)$ with a machine learning framework, one can generate data $\phi\sim p_{\rm data}$ by first sampling $\phi_{t=1}$ from $q$ and then solving the ODE~\eqref{eq:fm_ode} from $t=1$ to $t=0$. 

More specifically, in the FM one constructs the velocity field $v(\phi,t)$ and the probability path $p_t$ as follows. First, one introduces a conditional distribution $p_t(\phi|\phi_0)$, which gives the probability flow $p_t$ by
\begin{align}
    p_t(\phi) = \int d\phi_0\, p_t(\phi|\phi_0) p_0(\phi_0).\label{eq:fm_pt}
\end{align} 
Here, one needs to choose the conditional path $p_t(\phi|\phi_0)$ so that the conditions $p_{t=0}\!\approx\! p_{\rm data}$ and $p_{t=1}\!\approx\! q$ are satisfied. We also suppose that the evolution of the conditional distribution $p_t(\phi|\phi_0)$ is generated by a conditional velocity field $u_t(\phi|\phi_0)$, or equivalently, that it satisfies the continuity equation
\begin{align}
    \partial_t p_t(\phi|\phi_0) + \frac{\delta}{\delta\phi}\cdot(p_t(\phi|\phi_0)u_t(\phi|\phi_0))=0.\label{eq:cont_eq_fm}
\end{align}
Then, the velocity field $v(\phi,t)$ generating the flow of $p_t$ is obtained as the minimizer $v_\theta(\phi,t)$ of the cost function
\begin{widetext}
    \begin{align}
        \mathcal L_{\rm FM}(\theta) &= \mathbb E_{t,\phi_0\sim p_{\rm data}(\phi_0), \phi\sim p_t(\phi|\phi_0)} \left[||
        v_\theta(\phi,t) - u_t(\phi|\phi_0)
        ||_{\Omega_L}^2\right],\label{eq:fm_cost}
    \end{align}
\end{widetext}
which can be used as the training objective of flow matching. Here, $||\cdot||_{\Omega_L}$ denotes the $L_2$ norm of a field on $\Omega_L$, defined by $||f||_{\Omega_L} = (\int_{x\in\Omega_L} ||f_x||^2)^{1/2}$. We note that the minimizer $v_\theta^\star$ of Eq.~\eqref{eq:fm_cost} can be written explicitly as
\begin{align}
    v_\theta^\star(\phi,t) &= \int d\phi_0\, p_t(\phi_0|\phi) u_t(\phi|\phi_0),\label{eq:fm_v_explicit}
\end{align}
where $p_t(\phi_0|\phi)$ is the conditional distribution of $\phi_0$ when $\phi\sim \int d\phi_0\, p_t(\phi|\phi_0)p(\phi_0)$ is given.

One of the simplest FM constructions connects $p_{t=0}=p_{\rm data}$ and $p_{t=1}=\mathcal N(0,I)$ through the conditional path \cite{lipman2023}
\begin{align}
    p_t(\phi|\phi_0) = \mathcal N(\phi; (1-t)\phi_0, t^2I).\label{eq:cond_prob_fm}
\end{align}
In this case, the corresponding conditional velocity field $u_t(\phi|\phi_0)$ is given by 
\begin{align}
    u_t(\phi|\phi_0)=\frac{1}{t}(\phi-\phi_0),\label{eq:cond_vel_fm}
\end{align}
which is a simple linear combination of $\phi$ and $\phi_0$. Using the conditional velocity~\eqref{eq:cond_vel_fm} and the objective~\eqref{eq:fm_cost}, one can learn the velocity field $v(\phi,t)$ generating the path $p_t$, and subsequently generate $\phi\sim p_{\rm data}$ through the ODE flow~\eqref{eq:fm_ode}. Figure~\ref{fig:fm_rgfm}(b) illustrates the corresponding generative flow, in which the generative ODE transforms white Gaussian noise into data. Hereafter, we refer to the flow-matching scheme defined above as the standard FM.

For later use, in the standard FM, we rewrite the velocity field minimizing Eq.~\eqref{eq:fm_cost} in terms of the score $s_t(\phi) = (\delta/\delta\phi)\ln p_t(\phi)$ as
\begin{align}
        v_x^\star(\phi,t) &= -\frac{1}{1-t} \phi_x - \frac{t}{1-t} s_{tx}(\phi), \label{eq:vx_fm}
\end{align}
where the $x$-th component of the score $s_t(\phi)$ is given by $s_{tx}(\phi) = (\delta/\delta\phi_x)\ln p_t(\phi)$. To derive this relation, we use the conditional distribution~\eqref{eq:cond_prob_fm} and conditional velocity field~\eqref{eq:cond_vel_fm} of the standard FM. With Eqs.~\eqref{eq:cond_prob_fm} and \eqref{eq:cond_vel_fm}, the minimizing velocity field $v^\star$~\eqref{eq:fm_v_explicit} is expressed as 
\begin{align}
    v^\star(\phi,t) &= \mathbb E_{\phi_0\sim p_t(\phi_0|\phi)} [u_t(\phi|\phi_0)]\\
    &= \frac{1}{t} \phi - \frac{1}{t} \langle \phi_0 \rangle_{\phi_0\sim p_t(\phi_0|\phi)},\label{eq:vx_fm_der1}
\end{align}
whereas the score $s_t(\phi)$ can be written as 
\begin{align}
    s_t(\phi) &= \frac{1}{p_t(\phi)} \frac{\delta p_t(\phi)}{\delta\phi}\\
    &= \frac{1}{p_t(\phi)} \int d\phi_0\, \frac{\delta}{\delta\phi} p_t(\phi|\phi_0) p(\phi_0)\\
    &= \int d\phi_0\, \frac{(1-t)\phi_0 - \phi}{t^2} p_t(\phi_0|\phi)\\
    &= \frac{1-t}{t^2} \langle\phi_0\rangle_{\phi_0\sim p_t(\phi_0|\phi)} - \frac{\phi}{t^2}.\label{eq:vx_fm_der2}
\end{align}
Combining Eqs.~\eqref{eq:vx_fm_der1} and~\eqref{eq:vx_fm_der2}, we obtain Eq.~\eqref{eq:vx_fm}; this equation shows that $v_x(\phi,t)$ consists of a local contribution determined solely by $\phi_x$ and a generally nonlocal contribution that depends on the full field configuration $\phi$ through the score $s_t(\phi)$.

\subsection{Renormalization group flow matching}

As noted above, many natural datasets $p_{\rm data}$ exhibit an intrinsic hierarchy of scales, ranging from coarse structure to fine detail. Moreover, their power spectra often obey an approximate power law, ${\rm Var}[\phi_k] \sim |k|^{-\alpha}$ with $\alpha \sim 2$, reminiscent of field-theoretical models in statistical physics \cite{ruderman1994a,vanderschaaf1996a,saremi2013a}. This observation motivates the field-theoretical ansatz for the data action $S_{\rm data}=-\ln p_{\rm data}$ of the form
\begin{align}
S_{\rm data}(\phi) = \frac{1}{2}\int_x (\nabla\phi_x)^2 + U(\phi),
\label{eq:data_action}
\end{align}
where $U(\phi)$ is an interaction term that captures nonlocal structures in the data. Since data are generally defined on a lattice space with total linear size $L$, the data theory~\eqref{eq:data_action} has an ultraviolet (UV) wavenumber cutoff $\Lambda_{\rm UV}$ and an infrared (IR) wavenumber cutoff $\Lambda_{\rm IR}$, determined by the data lattice spacing $a$ and the linear system size $L$ as $\Lambda_{\rm UV}=\pi/a$ and $\Lambda_{\rm IR} = \pi/L$, respectively. For concreteness, we adopt the units with $a=1$ throughout the paper. For multiscale data described by Eq.~\eqref{eq:data_action}, the renormalization group (RG) \cite{wilson1975a,wilson1983,polchinski1984a} provides a systematic coarse-graining procedure that progressively integrates out modes from high to low wavenumbers as the cutoff decreases from $\Lambda_{\rm UV}$ to $\Lambda_{\rm IR}$. By choosing a probability path $p_t$ whose forward evolution from $t = 0$ to $t = 1$ follows the RG flow, one obtains a generative model that constructs data from coarse to fine scales by reversing this flow from $t = 1$ to $t = 0$.

To this end, we employ the framework of the exact RG \cite{polchinski1984a,wetterich1993a,morris1994a,kopietz2010}. We parameterize the RG wavenumber scale as $\Lambda(t)=\Lambda_{\rm UV}e^{-t/\tau}$ and choose $\tau$ such that $\Lambda(1)<\Lambda_{\rm IR}$, ensuring that all modes of $p_{\rm data}$ have been integrated out by $t=1$. For concreteness, in this work, we choose $\tau=c(\ln L)^{-1}$ with a constant $c$ independent of the linear system size $L$. Given a UV action, the exact RG provides an effective theory at the wavenumber scale $\Lambda$ in the form
\begin{align}
S^{\rm RG}_\Lambda(\phi) = \frac{1}{2} \int_k K_\Lambda^{-1}(k) (k^2+m^2) |\phi_k|^2 + U_\Lambda(\phi),
\label{eq:eff_action_erg}
\end{align}
where $\int_k = \int\frac{d^dk}{(2\pi)^d}$ denotes integration over wavenumber space, and we introduce a small mass parameter $m$ to regularize the theory in the limit $k \to 0$. Here, we introduce the RG cutoff function $K_\Lambda(k)$ by
\begin{align}
K_\Lambda(k) = \kappa\left(\frac{k^2+m^2}{\Lambda^2}\right),
\label{eq:K_Lam}
\end{align}
where $\kappa(x)$ is a monotonically decreasing function satisfying $\lim_{x\to 0}\kappa(x)=1$ and $\lim_{x\to\infty}\kappa(x)=0$.
Throughout this work, we use an exponential cutoff
\begin{align}
\kappa(x) = e^{-x},
\label{eq:kappax}
\end{align}
which corresponds to a heat-kernel RG cutoff and is one of the common choices in the functional RG \cite{litim2000a,litim2001a,berges2002}. In the effective theory~\eqref{eq:eff_action_erg}, fluctuations $\phi_k$ with $|k|\gtrsim\Lambda$ are suppressed and integrated out because $K_\Lambda(k)\sim 0$ in this regime. The resulting theory at scale $\Lambda$ therefore describes the remaining modes with $|k|\lesssim \Lambda$.

In Ref.~\cite{polchinski1984a}, Polchinski showed that correlations of the modes $\phi_k$ below the RG scale $\Lambda$ are exactly preserved if fluctuations near $k\sim \Lambda$ are incorporated into the interaction $U_\Lambda$ according to
\begin{align}
    \partial_\Lambda U_\Lambda = - \frac{1}{2} \int_k G_0(k)\partial_\Lambda K_\Lambda(k) \left(\frac{\delta^2U_\Lambda}{\delta\phi_k\delta\phi_{-k}} \!-\! \frac{\delta U_\Lambda}{\delta\phi_k}\frac{\delta U_\Lambda}{\delta\phi_{-k}} \right),\label{eq:pol}
\end{align}
where we introduce the bare Green function $G_0(k)=(k^2+m^2)^{-1}$. Based on the RG flow~\eqref{eq:pol}, one may, in principle, construct an FM generative model, whose probability path $p_t$ coincides with the RG effective distribution $p_\Lambda^{\rm RG} \propto e^{-S_\Lambda^{\rm RG}}$. However, along this path, the distributions of the eliminated modes converge to the singular zero-fluctuation distribution $\delta(|\phi|)$, which can cause numerical instabilities when learning the FM velocity field. Therefore, instead of the exact RG effective theory~\eqref{eq:eff_action_erg}, we use the rescaled action 
\begin{align}
    S_\Lambda(\phi) = S^{\rm RG}_{\Lambda}(\sqrt{K_{\Lambda}}\phi)\label{eq:eff_action}
\end{align}
and construct an FM model along the rescaled probability path $p_\Lambda\propto e^{-S_\Lambda}$. The rescaling factor $\sqrt{K_\Lambda}$ is chosen to keep the Gaussian part of $S^{\rm RG}_\Lambda$~\eqref{eq:eff_action_erg} invariant, which takes the form $S_{\rm GS}(\phi) = \frac{1}{2} \int_k (k^2+m^2)|\phi_k|^2$. Previous work \cite{masuki2025b} has shown that diffusion models based on the rescaled RG probability path $p_\Lambda\propto e^{-S_\Lambda(\phi)}$ can improve generative performance on multiscale data such as images and protein structures. We here extend this diffusion-model framework to the FM.

Combining Eqs.~\eqref{eq:eff_action_erg}, \eqref{eq:pol}, and~\eqref{eq:eff_action}, we find that the probability path $p_t\propto e^{-S_{\Lambda(t)}}$ obeys \cite{cotler2023b,masuki2025b}
\begin{align}
    \partial_t p_t &\!=\! -\frac{1}{2}\int_k \!\left[
        \frac{G_0(k)\partial_t K_{tk}}{K_{tk}} \frac{\delta^2 p_t}{\delta\phi_k\delta\phi_{-k}} \!+\! \frac{\delta}{\delta\phi_k}\! \left(\!\frac{\partial_t K_{tk}}{K_{tk}} \phi_k p_t\right)
    \!\right]\!\!.\label{eq:pt_diff}
\end{align}
For notational simplicity, we denote $p_{\Lambda(t)}$ and $K_{\Lambda(t)}(k)$ by $p_t$ and $K_{tk}$, respectively. For the typical behavior of the RG cutoff function $K_{tk}$, see Fig.~\ref{fig:fm_rgfm}(c). Because Eq.~\eqref{eq:pt_diff} describes a convection-diffusion process, the flow of $p_t$ can be generated by the conditional diffusion process \cite{masuki2025b}
\begin{align}
    p_t(\phi|\phi_0) &= \prod_{k} \mathcal N(\phi_k; \sqrt{\bar\alpha_{tk}}\phi_{0k},\bar\beta_{tk}),\label{eq:rg_diffusion}\\
    \bar\alpha_{tk} &= K_{tk}, \bar\beta_{tk} = G_0(k)(1-K_{tk}),
\end{align}
which we call the RG diffusion throughout this paper. 

In the RG diffusion~\eqref{eq:rg_diffusion}, the modes $\phi_k$ are successively transformed into Gaussian noise $\mathcal N(0,G_0(k))$, proceeding from high to low wavenumbers. Consequently, the distribution $p_t(\phi)$ at time $t$ possesses the scale-separation property across the momentum scale $\Lambda(t)$ as 
\begin{align}
    p_t(\phi) = p_{{\rm eff},t}(\phi_<) p_{\rm GS}(\phi_>).\label{eq:scale_sep}
\end{align}
Here, the effective distribution $p_{{\rm eff},t}(\phi_<)$ is a nontrivial effective theory for the low-momentum modes $\phi_{k<\Lambda(t)}$, which exactly preserves the correlations in the data distribution $p_{t=0} \!=\! p_{\rm data}$, whereas $p_{\rm GS}(\phi_>)\propto e^{-\frac{1}{2}\int_{k>\Lambda(t)} (k^2+m^2)|\phi_k|^2}$ describes the Gaussian theory for the integrated-out high-momentum modes $\phi_{k>\Lambda(t)}$. We recall that the hyperparameter $\tau$ is chosen such that $p_{t=1}$ is sufficiently close to the Gaussian distribution $p_{\rm GS}(\phi)$. Strictly speaking, for a smooth cutoff such as the exponential regulator specified above, Eq.~\eqref{eq:scale_sep} should be understood as a controlled approximation rather than as an exact factorization at finite $\Lambda(t)$. 

We now define the renormalization group flow matching (RGFM) as the flow matching associated with the RG diffusion in Eq.~\eqref{eq:rg_diffusion}. The conditional velocity field $u_t(\phi|\phi_0)$ that corresponds to the RG diffusion~\eqref{eq:rg_diffusion} is given by
\begin{align}
    u_{tk}(\phi|\phi_0) &\!=\! -\frac{\partial_t K_{tk}}{2(1-K_{tk})}\left(\phi_k \!-\! \sqrt{K_{tk}}\phi_{0k}\right) \!+\! \frac{\partial_t K_{tk}}{2\sqrt{K_{tk}}}\phi_{0k}.\label{eq:cond_vel_rgfm}
\end{align}
Here, $u_{tk}$ denotes the $k$-th wavenumber component of the conditional velocity field $u_t$. This expression follows from the requirement that $p_t(\phi|\phi_0)$ and $u_t(\phi|\phi_0)$ satisfy the continuity equation~\eqref{eq:cont_eq_fm} for the conditional distribution. As in the standard FM, the RGFM velocity field $v(\phi,t)$ can be learned by minimizing Eq.~\eqref{eq:fm_cost} using the conditional velocity $u_t(\phi|\phi_0)$ in Eq.~\eqref{eq:cond_vel_rgfm}. Since the RGFM probability path converges to the Gaussian distribution $p_{\rm GS}(\phi)$ at $t=1$ (cf. Eq.~\eqref{eq:scale_sep}), data samples $\phi\sim p_{\rm data}$ can be generated by first sampling $\phi_{t=1}\sim p_{\rm GS}$ and then integrating the ODE~\eqref{eq:fm_ode} backward from $t=1$ to $t=0$. As illustrated in Fig.~\ref{fig:fm_rgfm}(b), the generative ODE starts from Gaussian fluctuations $p_{\rm GS}$ and progressively constructs data structure from long to short length scales.

As in the standard FM, the velocity field in the RGFM can be expressed explicitly in terms of the score $s_t(\phi) = (\delta/\delta\phi)\ln p_t$. Using Eqs.~\eqref{eq:rg_diffusion} and \eqref{eq:cond_vel_rgfm}, we find that its $k$-th wavenumber component is
\begin{align}
    v^\star_k(\phi,t) &= \frac{\partial_t K_{tk}}{2K_{tk}} \phi_{k} + G_0(k) \frac{\partial_t K_{tk}}{2K_{tk}} s_{tk}(\phi). \label{eq:vk_rgfm}
\end{align}
Here, we use the fact that the velocity field that minimizes the cost function~\eqref{eq:fm_cost} in the RGFM is given by $v^\star_k(\phi,t) = \mathbb E_{\phi_0\sim p_t(\phi_0|\phi)} [u_{tk}(\phi|\phi_0)]$, together with the relation
\begin{align}
    s_{tk}(\phi) &= 
    \frac{k^2+m^2}{1-K_{tk}}\left(
        - \phi_k + \sqrt{K_{tk}} \langle \phi_{0k}\rangle_{\phi_0\sim p_t(\phi_0|\phi)}
    \right),\label{eq:sk_rgfm}
\end{align}
which follows from the RG diffusion~\eqref{eq:rg_diffusion} and the conditional velocity field~\eqref{eq:cond_vel_rgfm} in the RGFM. Since $K_{tk}$ is given by $K_{tk} = \exp(-(k^2+m^2)/\Lambda(t)^2)$, we have $G_0(k)\partial_t K_{tk}/(2K_{tk})=-1/(\Lambda(t)^2\tau)$. Thus, the real-space velocity field $v(\phi,t)$ corresponding to Eq.~\eqref{eq:vk_rgfm} takes the simple form
\begin{align}
    v_x(\phi,t) &= -\frac{1}{\Lambda(t)^2\tau} \left[ (-\nabla^2 +m^2)\phi_x + s_{tx}(\phi) \right].\label{eq:vx_rgfm}
\end{align}
As in the standard FM, Eq.~\eqref{eq:vx_rgfm} shows that the real-space RGFM velocity $v_x(\phi,t)$ consists of a local contribution proportional to $(-\nabla^2+m^2)\phi_x$ and a generally nonlocal contribution that depends on the full field configuration through the score $s_{tx}(\phi)$.

\begin{figure*}
    \centering
    \includegraphics[width=17.9cm]{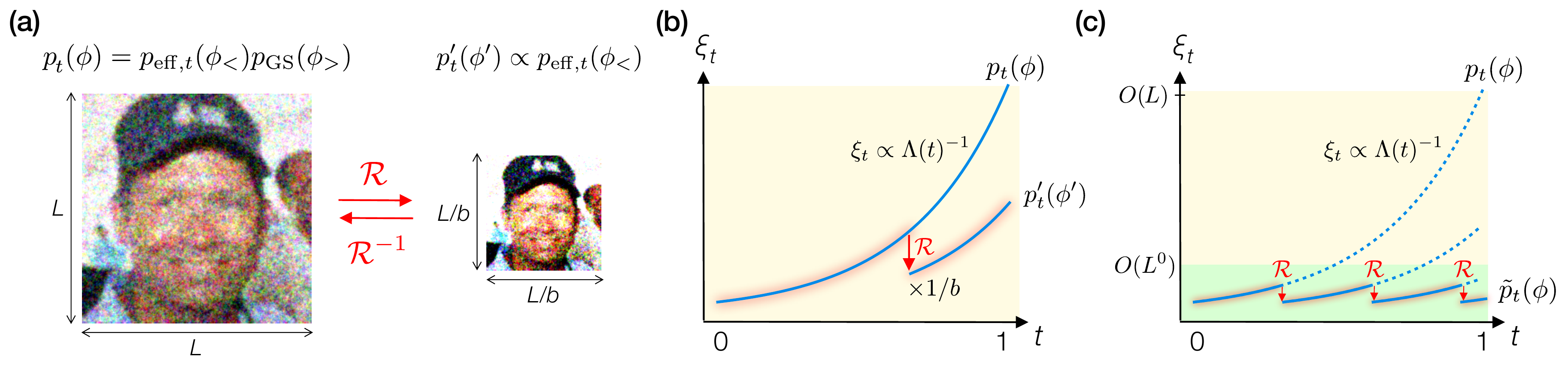}
    \caption{\label{fig:res_rgfm}
    (a) Site-decimation transformation $\mathcal R$ and its probabilistic inverse $\mathcal R^{-1}$. The coarse-grained field $\phi'$ is obtained by applying the discrete cosine transform (DCT) to $\phi$, discarding the high-wavenumber modes $\phi_>$ with $|k|>\Lambda(t)$, and transforming the retained modes $\phi_<$ back onto the decimated lattice. Conversely, $\phi$ is reconstructed probabilistically by sampling the missing modes from $p_{\rm GS}(\phi_>)$ and applying the inverse DCT to the combined modes.
    (b) Score-locality length $\xi_t$ along the RG diffusion path $p_t$. While $\xi_t$ grows as $c\Lambda(t)^{-1}$, the transformation $\mathcal{R}$, which reduces the linear lattice size from $L$ to $L/b$, restores the locality length from $\xi_t$ to $\xi_t/b$ in lattice units.
    (c) Successive site-decimation transformations define the rescaled probability path $\tilde p_t$, whose score-locality length remains $O(L^0)$ throughout the probability path from $t=0$ to $t=1$. Each segment can thus be approximately tracked by the ODE flow~\eqref{eq:fm_ode} generated by a local velocity field $v_{{\rm loc},t}$.}
\end{figure*}

\section{\label{sec:lgm_with_rgfm}Local generative modeling with RGFM}

In the previous section, we defined RGFM, whose probability path $p_t$ is based on the Polchinski RG flow~\eqref{eq:pol}.
Importantly, the RG-based probability path $p_t$ has two distinct features. The first is the scale-separation property of the RG~\eqref{eq:scale_sep}; the high-wavenumber modes $\phi_{k>\Lambda(t)}$ are integrated out of the effective theory and behave as independent Gaussian fluctuations. The second is the locality along the flow; if the probability flow starts from a physically reasonable data distribution, the locality length scale of $p_t$, which determines the patch size required for local generative modeling, scales linearly with the running RG length scale $\Lambda(t)^{-1}$. The latter property will be rigorously established in the subsequent sections. Intuitively, this locality of $p_t$ can be understood from the coarse-graining operation underlying the RG. As exemplified by Kadanoff blocking, the RG coarse-grains local regions of linear size $\sim \Lambda(t)^{-1}$. Thus, if the RG flow starts from a physically reasonable local theory~\eqref{eq:data_action}, the interaction range in the effective theory $p_t$ is expected to be at most on the order of the running RG length scale $\Lambda(t)^{-1}$.

Building on these two key features, we introduce a local generative model based on RGFM, which uses patches of linear size $O(\ln L)$ throughout the generative flow. 
To this end, we first introduce the site-decimation transformation of field configurations, which rescales the underlying lattice and restores the locality of the probability flow $p_t$. Because this transformation is reversible at the level of probability distributions owing to the scale-separation property of the RG, the rescaled flow can be traversed backward from $t=1$ to $t=0$, enabling local generative modeling. We then describe the training and sampling procedures within the local-patch framework.

\subsection{Rescaled RGFM flow}

When the data distribution possesses physically reasonable locality properties, the length scale $\xi_t$ along the RG diffusion scales approximately as $\xi_t\sim c\Lambda(t)^{-1}$ with a constant $c$; this notion of locality will be formulated precisely as score locality in Sec.~\ref{sec:loc_app}. Since $\Lambda(t)$ decreases from $\Lambda_{\rm UV}$ to $\Lambda_{\rm IR}$ along the RG flow, the locality length scale, which determines the patch size required for local generative modeling, grows from the microscopic scale $\Lambda_{\rm UV}^{-1}$ to the macroscopic scale $\Lambda_{\rm IR}^{-1}$. In particular, at $t\!=\!1$, one has $\xi_1\sim\Lambda_{\rm IR}^{-1}=O(L)$. A direct implementation of local generative modeling along this probability path would therefore require patches whose linear size grows with the system size, preventing the computational cost from scaling efficiently.

Importantly, the RGFM probability flow inherits the scale-separation property of the RG given in Eq.~\eqref{eq:scale_sep}. At each time $t$, the distribution $p_t(\phi)$ separates into a trivial Gaussian sector $p_{\rm GS}(\phi_>)$ for the integrated-out high-wavenumber modes $\phi_{k>\Lambda(t)}$, and a nontrivial effective distribution $p_{{\rm eff},t}(\phi_<)$ for the remaining low-wavenumber modes $\phi_{k<\Lambda(t)}$. By discarding the decoupled high-wavenumber modes $\phi_>$, one can therefore decimate the lattice and restore a microscopic score-locality length scale when measured in units of the coarse-grained lattice.

More specifically, suppose that $\Lambda(t)=\Lambda_{\rm UV}/b$. We first apply the discrete cosine transform (DCT) to $\phi$ and retain only the low-wavenumber modes $\phi_k$ satisfying $|k|<\Lambda(t)$. We then apply the inverse DCT to the retained coefficients on a reduced lattice, thereby defining the coarse-grained field $\phi'=\mathcal R_t(\phi)$ on a lattice whose number of sites is reduced from $L^d$ to $(L/b)^d$; see Fig.~\ref{fig:res_rgfm}(a). 
Conceptually, the transformation $\mathcal R_t$ removes the short-distance degrees of freedom and represents the remaining long-wavelength structure using fewer lattice sites. Since one lattice spacing of $\mathcal R_t(\phi)$ corresponds to $b$ lattice spacings of the original field $\phi$, a fixed physical length is reduced by a factor of $b$ when expressed in coarse-lattice units. Consequently, if the score-locality length scale of $p_t(\phi)$ is $c\Lambda(t)^{-1}$ before the transformation, that of the induced distribution $p'_t=(\mathcal R_t)_*p_t$ is approximately $c\Lambda(t)^{-1}/b=c\Lambda_{\rm UV}^{-1}=O(L^0)$ in coarse-grained lattice units; see Fig.~\ref{fig:res_rgfm}(b).

By applying this rescaling $O(\ln L)$ times during the RG diffusion, we can construct a probability path $\tilde p_t$ whose score-locality length scale remains $O(L^0)$ in the lattice units used at each stage of the flow; see Fig.~\ref{fig:res_rgfm}(c). Within each time interval between successive rescaling operations $\mathcal{R}_t$, the probability flow $\tilde p_t$ can then be approximately tracked by a flow generated by a local velocity field using patches of constant linear size. 

While the field transformation $\phi\mapsto\mathcal R_t(\phi)$ is many-to-one and hence not invertible for individual field configurations, the corresponding transformation of probability distributions, $p_t\mapsto (\mathcal R_t)_{*}p_t$, admits an exact probabilistic inverse owing to the scale-separation property~\eqref{eq:scale_sep}. 
Given a sample $\phi'$ on the coarse-grained lattice, one first reconstructs the low-momentum modes $\phi_<$ by applying the DCT to $\phi'$. One then independently samples the missing high-momentum modes $\phi_>$ from the Gaussian sector $p_{\rm GS}(\phi_>)$ and combines them with $\phi_<$. Applying the inverse DCT to the combined modes yields a field distributed according to the original distribution $p_t(\phi)$. 
One can therefore traverse the rescaled probability path $\tilde p_t$ from $t\!=\!1$ to $t\!=\!0$ by alternating this probabilistic coarse-to-fine inverse with local reverse-ODE evolution between successive site-decimation transformations $\mathcal R_t$.
Thus, the rescaled probability path $\tilde p_t$ admits local generative modeling.

\begin{figure}[t]
    \centering
    \includegraphics[width=8.5cm]{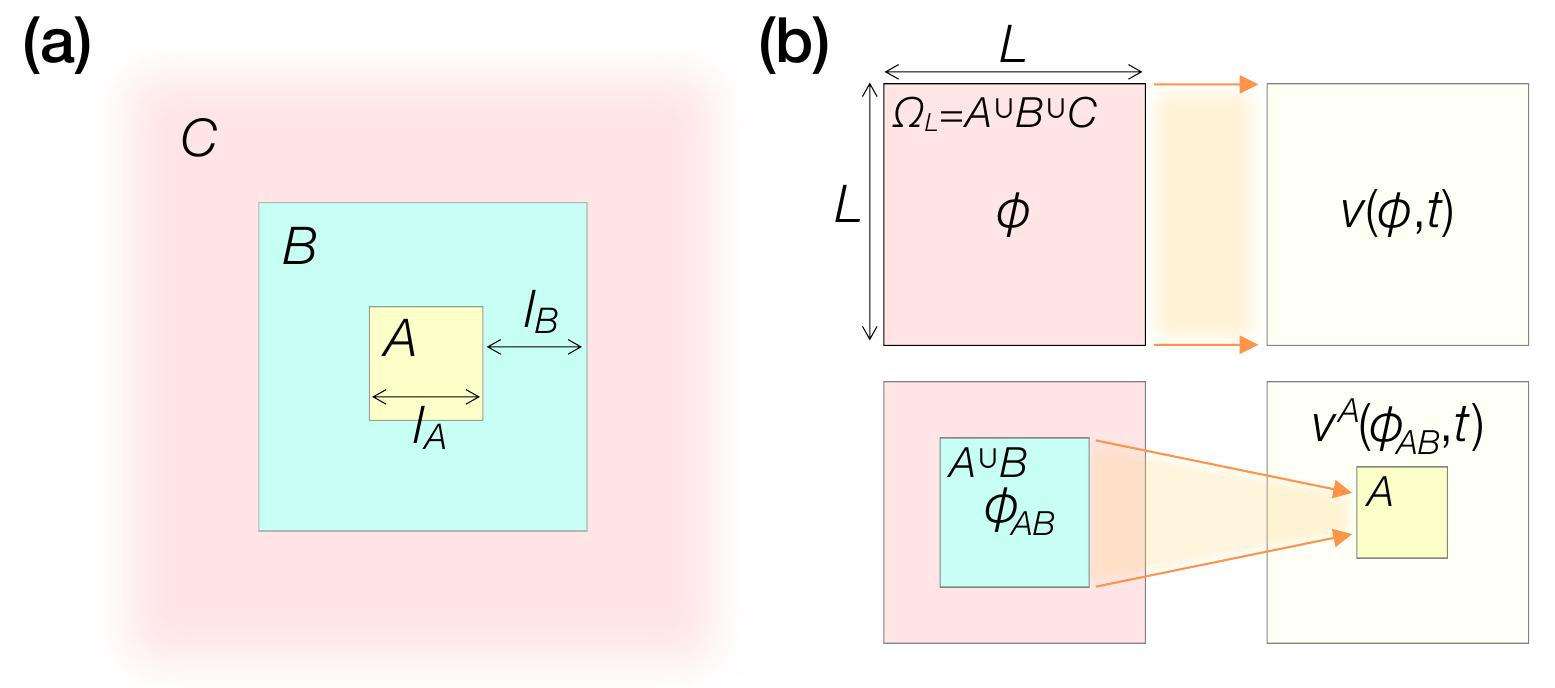}
    \caption{\label{fig:lrd}
    (a) Decomposition of the spatial domain $\Omega_L$ into local regions $A$ and $B$ and the remaining region $C=\overline{A\cup B}$.
    (b) Global and local velocity fields. In general, the velocity field $v(\phi,t)$ depends on the entire field configuration $\phi$. The local velocity field on region $A$, denoted by $v^A(\phi_{AB},t)$, depends only on the field configuration $\phi_{AB}$ on $A\cup B$.
    }
\end{figure}

\subsection{Training objective and sampling procedure}

In local generative modeling with RGFM, we learn a local velocity field along the rescaled probability path $\tilde p_t$ introduced above. 
Specifically, we apply the site-decimation transformation $\mathcal R_t$ at times $0<t_1<t_2<\cdots<t_N<1$ and denote the transformation associated with time $t$ by $\widetilde{\mathcal R}_t$. By definition, $\widetilde{\mathcal R}_t=\mathcal R_{t_n}$ for $t_n\leq t<t_{n+1}$.
At each scale, we partition the decimated lattice $\tilde{\Omega}_L$ into disjoint local regions $\{A_{it}\}$ of linear size $l_A$ and take a buffer region $B_{it}$ of common width $l_B$ around each $A_{it}$. The geometry of the local regions $A$, $B$, and $C=\overline{A\cup B}$ is illustrated in Fig.~\ref{fig:lrd}(a). 
The volumes of the local patches $A_{it}$ and $A_{it}\cup B_{it}$ are $l_A^d$ and $(l_A+2l_B)^d$, respectively.

Since the rescaling transformation $\widetilde{\mathcal R}_t$ is constant on each interval $[t_n,t_{n+1}]$, the velocity field $v(\phi,t)$ generating the rescaled probability flow $\tilde p_t$ can be obtained by minimizing the flow-matching objective (cf. Eq.~\eqref{eq:fm_cost})
\begin{widetext}
\begin{align}
    \mathcal L_{\rm RGFM}(\theta) &= \mathbb E_{t\sim{\rm Uniform}[0,1],\phi_0\sim p_{\rm data},\phi\sim p_t(\phi|\phi_0)} \left[
        \left\| v_\theta(\widetilde{\mathcal R}_t\phi,t) - \widetilde{\mathcal R}_tu_t(\phi|\phi_0) \right\|_{\tilde\Omega_L}^2
    \right].
\end{align}
\end{widetext}
Here, $u_t(\phi|\phi_0)$ is the conditional velocity field in Eq.~\eqref{eq:cond_vel_rgfm} associated with the RG diffusion path $p_t(\phi|\phi_0)$ in Eq.~\eqref{eq:rg_diffusion}, and $\widetilde{\mathcal R}_tu_t$ is the conditional velocity field rescaled by the site-decimation transformation $\widetilde{\mathcal R}_t$.

In local generative modeling, we approximate the velocity field $v(\phi,t)$ by a local field $v_{\rm loc}(\phi,t)$. Its component $v_{\rm loc}^{A_{it}}$ on region $A_{it}$ is determined by the field configuration on the local patch $A_{it}\cup B_{it}$, and is written as $v_{\rm loc}^{A_{it}}(\phi_{A_{it}B_{it}},t)$; see Fig.~\ref{fig:lrd}(b). 
The corresponding cost function for the local velocity field $v_{\rm loc}(\phi,t)$ is
\begin{widetext}
\begin{align}
    \mathcal L_{\rm RGFM}^{\rm loc}(\theta) = \mathbb E_{t\sim{\rm Uniform}[0,1],\phi_0\sim p_{\rm data},\phi\sim p_t(\phi|\phi_0)}\left[
    \sum_i \left\|
        v_{{\rm loc},\theta}^{A_{it}}((\widetilde{\mathcal R}_t\phi)_{A_{it}B_{it}},t) - (\widetilde{\mathcal R}_t u_t(\phi|\phi_0))_{A_{it}}
    \right\|_{A_{it}}^2
    \right].\label{eq:local_res_rgfm_cost}
\end{align}
\end{widetext}
Here, the local model $v_{{\rm loc},\theta}$ takes as input the restriction of $\widetilde{\mathcal R}_t\phi$ to $A_{it}\cup B_{it}$ together with the time $t$ and patch position $i$. The model predicts the velocity only on the target region $A_{it}$. Because the model is conditioned on time $t$ and patch position $i$, a single local neural network $v_{{\rm loc},\theta}$ can be shared across all patches.

To generate data, we alternate between reverse-ODE evolution under the learned local velocity field on each interval $[t_n,t_{n+1}]$ and stochastic inversion of the site-decimation transformations $\mathcal R_t$ at each $t_n$. Sampling starts from $\widetilde{\mathcal R}_1\phi_1$, where $\phi_1\sim p_{\rm GS}$, and the reverse ODE is integrated on each time interval between successive site-decimation transformations. At each scale-transition time, the current configuration is lifted to the finer lattice by retaining its low-momentum modes and sampling the missing high-momentum modes independently from the Gaussian sector $p_{\rm GS}(\phi_>)$. Repeating these reverse-flow and stochastic-lifting steps until $t=0$ yields a sample in the original data space. The training and sampling procedures are summarized in Algorithms~\ref{alg:local_rgfm_train} and \ref{alg:local_rgfm_sample}.

\begin{revtexalgorithm}
\algorithmcaption{\label{alg:local_rgfm_train}Training procedure for local generative modeling with RGFM}
\begin{algorithmic}[1]
    \State \textbf{Repeat}
    \State \hspace{12pt}Sample $\phi_0\sim p_{\rm data}(\phi_0)$.
    \State \hspace{12pt}Sample $t\sim {\rm Uniform}[0,1]$.
    \State \hspace{12pt}Sample $\phi_t \sim p_t(\phi_t|\phi_0)$ using the RG diffusion~\eqref{eq:rg_diffusion}.
    \State \hspace{12pt}Compute $u_t(\phi_t|\phi_0)$ using Eq.~\eqref{eq:cond_vel_rgfm}.
    \State \hspace{12pt}Compute $g = \nabla_\theta \mathcal L_{\rm RGFM}^{\rm loc}(\theta)$ using the objective~\eqref{eq:local_res_rgfm_cost}.
    \State \hspace{12pt}Update the model parameter $\theta$ using $\theta - \eta g$.
    \State \textbf{Until} converged.
    \State \textbf{Return} Local velocity field $v_{{\rm loc},\theta}$.
\end{algorithmic}
\end{revtexalgorithm}

\section{\label{sec:loc_app}Local approximability of the probability flow in the RGFM}

In the previous section, starting from the RGFM probability flow $p_t$, we defined the rescaled flow $\tilde p_t$ and formulated local generative modeling along it. This construction is motivated by the physical intuition that the locality length scale of $p_t$, which determines the patch size required to accurately approximate the ODE~\eqref{eq:fm_ode} using local velocity fields, scales proportionally to the running RG length scale $\Lambda(t)^{-1}$. In this section, we establish this intuition mathematically.

To this end, we first identify physically reasonable assumptions on the data distribution $p_{\rm data}$, as summarized in Sec.~\ref{sssec:assump_data}. In particular, we argue that a broad class of data distributions can be described by local or conditionally local actions. We then introduce assumptions on the RG evolution of such distributions, as summarized in Sec.~\ref{sssec:assump_rg}. We separately impose a regularity assumption on the local approximation of the velocity field in Sec.~\ref{ssec:la_of_v}. Under these assumptions, we establish the main result of this section: for any prescribed accuracy $\varepsilon$, the probability flow $p_t$ can be approximated by an ODE flow generated by local velocity fields whose receptive fields have linear size $O\left(\Lambda(t)^{-1}\ln\left(L^\alpha/\varepsilon\right)\right)$, where the exponent $\alpha$ satisfies $\alpha\leq 2d/3$.

\begin{revtexalgorithm}
\algorithmcaption{\label{alg:local_rgfm_sample}Sampling procedure for local generative modeling with RGFM}
\begin{algorithmic}[1]
    \Require Trained local velocity field $v_{{\rm loc},\theta}$.
    \Require Rescaling times $0=t_0<t_1<\cdots<t_N<1$, at which $\mathcal R_t$ is applied.

    \State Sample $\phi'_{1}=\widetilde{\mathcal R}_{1}\phi_1$ with $\phi_1\sim p_{\rm GS}(\phi_1)$.
    \State Evolve $\phi'_1$ from $t=1$ to $t_N$ by the ODE flow~\eqref{eq:fm_ode} with $v_{{\rm loc},\theta}$.
    \For{$n=N,N-1,\ldots, 1$}
    \State \hspace{12pt}Upsample $\phi'_{t_n}$ by sampling $\phi'_{k>\Lambda(t)}$ from $p_{\rm GS}$.
    \State \hspace{12pt}Evolve $\phi'_{t}$ from $t_n$ to $t_{n-1}$ by the ODE flow~\eqref{eq:fm_ode} with \Statex\hspace{27pt}$v_{{\rm loc},\theta}$.
    \EndFor
    \State \textbf{Return} Sampled data $\phi'_{t=0}$, which is at the original resolution.
\end{algorithmic}
\end{revtexalgorithm}

\begin{figure*}[t]
\centering
\begin{tikzpicture}[
    >=Latex,
    node distance=16mm and 28mm,
    box/.style={
        draw,
        rounded corners,
        minimum width=16mm,
        minimum height=8mm,
        align=center
    },
    lab/.style={
        font=\small,
        inner sep=1pt,
        fill=white
    }
]

\node[box] (A) {Assumption~\ref{assump:data_dist}(i)\\{\! }Local data distributions };
\node[box, right=25mm of A] (B) {Assumption~\ref{assump:data_dist}(ii)\\{\! }Conditionally local data distributions };
\node[box, below=12mm of A] (C) {Theorem~\ref{thm:LA_loc_data}\\{\! }Bound for local approximation error of $v_t$};
\node[box, below=11.8mm of B] (D) {Theorem~\ref{thm:LA_cloc_data}\\{\! }Bound for local approximation error of $v_t$};
\node[box, below=12mm of $(C)!0.5!(D)$] (E) {Main Theorem 1: \ref{thm:LA_rgfm}\\{\! }Unified bound for local approximation error of $v_t$};
\node[box, below=30mm of C] (F) {{\! }Main Theorem 2:~\ref{thm:rgfm_flow_stability}\\$W_2$ is small under local ODE\\with patch size $O(\Lambda(t)^{-1}\!\ln L)$};
\node[box, below=30mm of D] (G) {{\! }Theorem~\ref{thm:rgfm_flow_stability_sde}\\$D_{\rm KL}$ is small under local SDE\\with patch size $O(\Lambda(t)^{-1}\!\ln L)$};
\node[box, below=12mm of F] (H) {{\!} Local generative modeling with RGFM\\Data generation with local ODE with patch size $O(\ln L)$ (Sec.~\ref{sec:lgm_with_rgfm})};

\draw[->] (A) -- node[lab, right=1mm] {{\! }RG locality assumption~\ref{assump:rg_loc_int}} (C);
\draw[->] (B) -- node[lab, right=1mm, align=left] {{\! }RG locality assumption~\ref{assump:rg_loc_int}\\{\! }RG assumption for latent~\ref{assump:rg_loc_cmi}} (D);
\draw[->] (C) to ([xshift=5mm]E.north west);
\draw[->] (D) to ([xshift=-5mm]E.north east);
\draw[->] ([xshift=5mm]E.south west) -- node[lab, right=2mm, align=left] {ODE stability theorem~\ref{thm:ode_stability}\\Lipschitz continuity assumption~\ref{assump:Lip_conti}} (F);
\draw[->] ([xshift=-5mm]E.south east) -- node[lab, right=1mm, align=left] {SDE stability theorem~\ref{thm:sde_stability}} (G);
\draw[->, dashed] (F) -- node[lab, right=1mm] {Site-decimation rescaling transformation (Sec.~\ref{sec:lgm_with_rgfm})} (H);
\end{tikzpicture}
\caption{\label{flowchart_proofs}Flow chart of the main theorems and their proofs in Sec.~\ref{sec:loc_app}.}
\end{figure*}
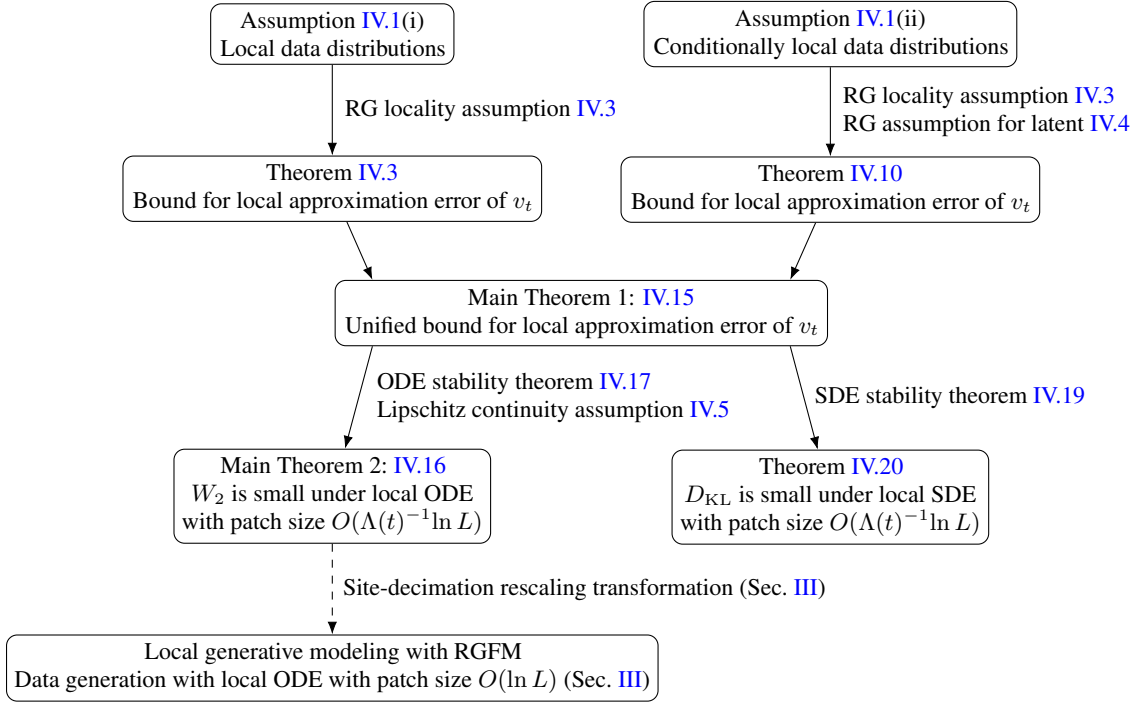

The proof proceeds as follows; see Fig.~\ref{flowchart_proofs} for an overview. In Sec.~\ref{ssec:assumptions}, we summarize the assumptions for data distributions and the properties of their RG flow. Then, in Sec.~\ref{ssec:la_of_v}, we define the local approximation $v_{{\rm loc},t}$ of the RGFM velocity field $v_t$ using local patches with buffer width $l_B$, together with the corresponding local approximation error. We show, in particular, that $v_{{\rm loc},t}$ is obtained as the minimizer of a patchwise flow-matching objective. We also introduce the Lipschitz continuity assumption on $v_{{\rm loc},t}$. In Sec.~\ref{ssec:lability_of_v}, based on the RG locality assumptions, we prove Theorems~\ref{thm:LA_loc_data} and \ref{thm:LA_cloc_data}, which establish that, for local and conditionally local data distributions, the local approximation error of $v_t$ is bounded by $e^{-c\Lambda l_B}L^\alpha\ln L$ with $\alpha\leq 2d/3$. Consequently, an approximation error smaller than $\varepsilon$ can be achieved by a buffer width of order $O\left(\Lambda(t)^{-1}\ln\left(L^\alpha/\varepsilon\right)\right)$, as stated in the main theorem~\ref{thm:LA_rgfm}. 

Combining this bound with the ODE stability theorem for the probability flow in FM~\ref{thm:ode_stability} and the regularity assumption on $v_{{\rm loc},t}$~\ref{assump:Lip_conti}, we obtain the main theorem~\ref{thm:rgfm_flow_stability}: in terms of the 2-Wasserstein distance, the probability flow $p_t$ of RGFM is accurately approximated by the flow generated by the local velocity field $v_{{\rm loc},t}$ with a receptive field of linear size $O\left(\Lambda(t)^{-1}\ln\left(L^\alpha/\varepsilon\right)\right)$. Finally, we discuss an equivalent SDE formulation~\ref{prop:equiv_sde}, in which the probability flow can also be locally approximated in terms of the KL divergence with local patches of size $O\left(\Lambda(t)^{-1}\ln\left(L^\alpha/\varepsilon\right)\right)$, as stated in Theorem~\ref{thm:rgfm_flow_stability_sde}.

For clarity and to make the logical dependencies among the propositions more transparent, we defer some of the technical proofs to Appendix~\ref{app:proofs}.

\subsection{\label{ssec:assumptions}Assumptions on the data distribution and its RG flow}

For concreteness, we briefly summarize the assumptions introduced below. First, we assume that the data distribution is either local or conditionally local (Assumption~\ref{assump:data_dist}). In the latter case, the action is local for each fixed latent variable $z$, uniformly in $z$. We further assume that the latent information can be approximately inferred mostly from a sufficiently large local neighborhood (Assumption~\ref{assump:latent_cmi}). Second, we assume that these locality properties are preserved under the RG flow, with the characteristic length scale growing as $\Lambda^{-1}$ (Assumption~\ref{assump:rg_loc_int}). For conditionally local distributions, we assume that the local predictability of the latent variable is also preserved along the flow (Assumption~\ref{assump:rg_loc_cmi}). In the remainder of this subsection, we motivate and formulate these assumptions precisely. The Lipschitz-continuity assumption on the local approximations of the velocity fields (Assumption~\ref{assump:Lip_conti}) is separately introduced in Sec.~\ref{ssec:la_of_v}, after we define the local approximation.

\subsubsection{\label{sssec:assump_data}Local and conditionally local data distributions}

As noted above, natural data often exhibit an intrinsic hierarchy of scales, ranging from coarse global structures to fine local details. In particular, the power spectra of natural data, such as images, empirically display an approximate power-law behavior, $\mathbb E[|\phi_k|^2]\sim |k|^{-\alpha}$ with $\alpha\sim 2$ \cite{ruderman1994a,vanderschaaf1996a,saremi2013a}, which is reminiscent of that observed in statistical-mechanical systems and field theories governed by local actions. It therefore motivates the local-action ansatz~\eqref{eq:data_action} for data distributions. 

For real-world data, however, the situation is more subtle, and a purely local description may be insufficient; natural data often exhibit strong long-range dependencies that enforce global coherence. In images of humans, for example, attributes such as age, pose, and identity must remain consistent across spatially separated regions. Such dependencies cannot, in general, be represented solely through local interactions among nearby degrees of freedom.

\begin{figure}[t]
    \centering
    \includegraphics[width=8.5cm]{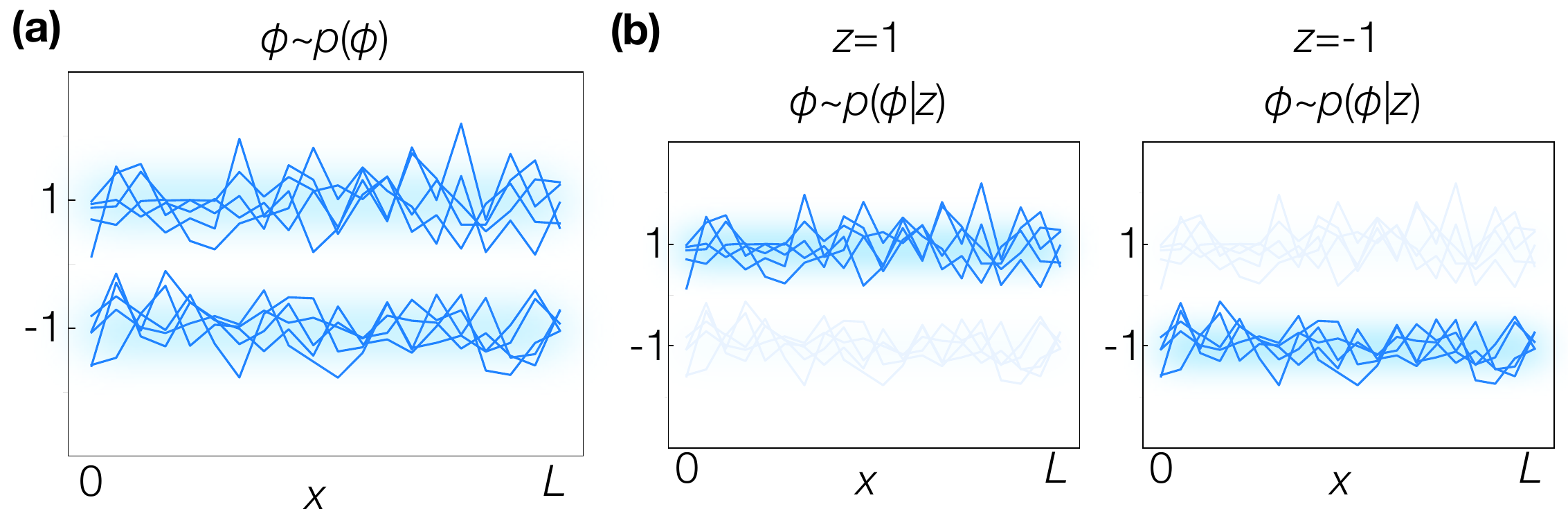}
    \caption{\label{fig:toy_cld}
    Toy example of a conditionally local distribution $p(\phi) = \int dz\, p(\phi|z)p(z)$, where $p(\phi|z) = \prod_{x=1}^L \mathcal N(\phi_x; z, \sigma^2)$ and $p(z) = {\rm Uniform}\{\pm 1\}$.
    (a) After marginalizing over $z$, $p(\phi)$ exhibits nonlocal global structure.
    (b) For a fixed $z$, the conditional distribution $p(\phi|z)$ is a local Gaussian distribution.
    }
\end{figure}

Motivated by these observations, we expect that the nonlocal structure of realistic data can be encoded by suitable latent variables $z$. More precisely, even when the marginal distribution $p_{\rm data}(\phi)$ exhibits long-range dependencies, we assume that these dependencies are mediated primarily by $z$. Once the latent variable is specified, the conditional distribution $p_{\rm data}(\phi|z)$ is expected to possess well-behaved locality properties. For image data, for instance, once global attributes such as age and pose are fixed, it is plausible that the remaining fluctuations can be described predominantly through local dependencies. As an illustration, we show a toy example of such a conditionally local distribution in Fig.~\ref{fig:toy_cld}. The marginal distribution $p(\phi)=\int dz\,p(z)p(\phi|z)$ exhibits long-range correlations because all spatial regions depend on the common latent variable $z\in\{-1,+1\}$. For fixed $z$, however, the conditional distribution $p(\phi|z)=\prod_x \mathcal N(\phi_x; z, \sigma^2)$ describes independent local fluctuations around the mean $z$.

In this work, based on these considerations, we study two classes of data distributions. The first consists of distributions whose actions are local. The second consists of distributions that admit a latent variable $z$ such that the corresponding conditional action is local for each fixed $z$. To formulate these properties precisely, we first introduce Definitions~\ref{def:Lqp_norm}, \ref{def:trun_interpolation}, and \ref{def:local_action}, regarding the locality of an action. 

\begin{definition}\label{def:Lqp_norm}
$L_q(p)$ norm of a functional $f(\phi)$.---
For any functional $f(\phi)$ of a field configuration $\phi$, we define its $L_q(p)$ norm with respect to a probability distribution $p(\phi)$ by
\begin{align}
    \left\|f(\phi)\right\|_{L_q(p)} = \left(\int d\phi\, p(\phi) |f(\phi)|^q\right)^{\frac{1}{q}}.
\end{align}

\end{definition}

We next introduce an interpolation that continuously suppresses fluctuations outside a prescribed local region.

\begin{definition}\label{def:trun_interpolation}
Truncation interpolation of a field $\phi$.---
For a probability distribution $p(\phi)$ and a local-region decomposition of $\Omega_L$ into $A$, $B$, and $C$ in Fig.~\ref{fig:lrd}, we define the truncation interpolation $T_{ABC}^\lambda$, with $0\leq\lambda\leq1$, by
\begin{align}
    T_{ABC}^\lambda \phi &= \phi_{AB} + \lambda \phi_C + (1-\lambda)\langle \phi_C\rangle_p.
\end{align}

\end{definition}

Here, $T_{ABC}^{\lambda=1}\phi=\phi$, whereas $T_{ABC}^{\lambda=0}\phi=\phi_{AB} + \langle \phi_C\rangle_p$, which is obtained by setting the field components in $C$ to their expectation values $\langle\phi_C\rangle_p$. Thus, $T_{ABC}^\lambda$ interpolates between the original configuration and a configuration in which fluctuations outside $A\cup B$ are removed.

Based on Definitions~\ref{def:Lqp_norm} and \ref{def:trun_interpolation}, we define the locality of a functional.

\begin{definition}\label{def:local_action}
Locality of a functional.---
Let $p(\phi)$ be the probability distribution of a field $\phi$. We say that a functional $U(\phi)$ is local with length scale $\xi$ if there exists a constant $\gamma$ such that, for any local-region decomposition of $\Omega_L$ into $A$, $B$, and $C$, any $x\in A$, $y\in C$, and any $0\leq\lambda\leq1$,
\begin{align}
    \left\|\left[\frac{\delta^2U(\psi)}{\delta\psi_x\delta\psi_y}\right]_{\psi=T_{ABC}^\lambda\phi}\right\|_{L_4(p)} \leq \left\|\frac{\delta U(\phi)}{\delta\phi_x}\right\|_{L_2(p)} \gamma e^{-|x-y|/\xi}.\label{eq:def:local_action}
\end{align}

\end{definition}

Intuitively, this condition states that the sensitivity of the interaction force $\delta U(\phi)/\delta\phi_x$ to the field $\phi_y$ decays exponentially with their separation $|x-y|$, relative to the typical magnitude of the force. The use of the interpolated configuration $T_{ABC}^\lambda\phi$ requires this decay to hold uniformly as the fluctuations in the exterior region $C$ are continuously removed. We note that, in Eq.~\eqref{eq:def:local_action}, the Hessian is computed before the substitution $\psi=T_{ABC}^\lambda\phi$. Since $T_{ABC}^\lambda$ only specifies the field configuration at which the already-computed Hessian is evaluated, we do not expect this prescription itself to impose an essential restriction on the class of local or quasi-local interactions considered here.

We now state our assumption on the data distribution.

\begin{assumption}\label{assump:data_dist}
    Assumption on the data distribution.---
    We assume that the data distribution $p_{\rm data}$ belongs to one of the following classes.

    \noindent (i) Local-action data distributions.

    The action $S_{\rm data}(\phi)=-\ln p_{\rm data}(\phi)$ can be written as
    \begin{align}
        S_{\rm data}(\phi) = \frac{1}{2}\int_k (k^2+m^2)|\phi_k|^2 + U_{\rm loc}(\phi),\label{eq:local_data_action}
    \end{align}
    where the functional $U_{\rm loc}(\phi)$ is local as in Definition~\ref{def:local_action} with a length scale of $O(L^0)$.

    \noindent (ii) Conditionally local data distributions.

    The data distribution admits a latent-variable representation
    $p_{\rm data}(\phi)=\int dz\,p(z)p_{\rm data}(\phi|z)$.
    For each $z$, the conditional action
    $S_{{\rm data},z}(\phi)=-\ln p_{\rm data}(\phi|z)$
    can be written as
    \begin{align}
        S_{{\rm data},z}(\phi) = \frac{1}{2}\int_k (k^2+m^2)|\phi_k|^2 + U_{{\rm loc},z}(\phi),\label{eq:cond_local_data_action}
    \end{align}
    where the functional $U_{{\rm loc},z}(\phi)$ is local as in Definition~\ref{def:local_action} with a length scale of $O(L^0)$. We assume that the parameters $\xi$ and $\gamma$ in Definition~\ref{def:local_action} can be chosen independently of the latent variable $z$.

    For both cases (i) and (ii), we also assume that $p_{\rm data}$ has a uniform $L_4$ bound: there exists a constant $M$ such that, for any $x\in\Omega_L$, 
    \begin{align}
        \left\|\phi_x\right\|_{L_4(p_{\rm data})} \leq M.\label{eq:phi_x_L4_bound}
    \end{align} 
\end{assumption}

As detailed below, for the conditionally local distributions, the locality of each $U_{{\rm loc},z}$ alone is not sufficient to guarantee that the RGFM of the marginal distribution $p_{\rm data}(\phi)$ can be locally tracked. We therefore also require that the latent information is essentially contained in a local configuration $\phi_{AB}$. To precisely formulate this notion, we first introduce the conditional mutual information.

\begin{definition}\label{def:cmi}
Conditional mutual information (CMI).---
Let $p_{\rm data}(\phi)=\int dz\, p(z)p_{\rm data}(\phi|z)$ be a conditionally local data distribution.
Consider a local-region decomposition of $\Omega_L$ into $A$, $B$, and $C$. We define the conditional mutual information (CMI) between the latent variable $z$ and region $C$, conditioned on the field in $A\cup B$, as
\begin{align}
    I_{\rm data}(Z\!:\!C|AB) &= H(AB, Z) + H(AB, C)\nonumber\\
    & \qquad - H(AB,C,Z) - H(AB),
\end{align}
where $H(\alpha)$ denotes the entropy of variable $\alpha$ under the joint distribution $p_{\rm data}(\phi_{AB},\phi_C,z)= p_{\rm data}(\phi|z)p(z)$.

\end{definition}

We assume that the additional information about $z$ contained in the exterior region decreases exponentially with the buffer width $l_B$.

\begin{assumption}\label{assump:latent_cmi}
Latent predictability.---
For a conditionally local data distribution with latent variable $z$, $p_{\rm data}(\phi) \!=\! \int dz\,p(z)p_{\rm data}(\phi|z)$, we assume that there exist constants $\gamma$ and $\xi$ such that, for any local regions $A$, $B$, and $C=\overline{A\cup B}$,
\begin{align}
    I_{\rm data}(Z\!:\!C|AB) \leq \gamma e^{-l_B/\xi}.
\end{align}

\end{assumption}

Intuitively, this assumption means that, once the field in $A\cup B$ is observed, the exterior region $C=\overline{A\cup B}$ provides only exponentially small additional information about the latent variable $z$. In other words, as the buffer width increases, the local configuration $\phi_{AB}$ becomes approximately sufficient for the latent inference in that the access to the exterior region $C$ does not substantially improve the inference.

\subsubsection{\label{sssec:assump_rg}Properties of the RG flow for local distributions}

As described in Sec.~\ref{sec:rgfm}, the effective interaction $U_\Lambda$ in the Polchinski RG equation~\eqref{eq:pol} incorporates the effects of the short-wavelength fluctuations integrated out between the UV scale $\Lambda_{\rm UV}$ and the running scale $\Lambda$. Therefore, starting from a local interaction $U_{\rm loc}$, one naturally expects $U_\Lambda$ to remain quasi-local: its dependence on fields separated by distances much larger than the running RG length scale $\Lambda^{-1}$ should be strongly suppressed. Based on this observation, we impose the following assumption.

\begin{assumption}\label{assump:rg_loc_int}
Locality of the interaction during the RG flow.---
Let $p_\Lambda$ be the probability distribution obtained by applying the RG diffusion to a local data distribution $p_{\rm data}(\phi)$ satisfying Assumption~\ref{assump:data_dist}. We denote the interaction part of $p_\Lambda$ by $V_\Lambda(\phi)$, which is related to the interaction $U_\Lambda(\phi)$ generated by the Polchinski RG equation~\eqref{eq:pol} through $V_\Lambda(\phi)=U_\Lambda(\sqrt{K_\Lambda}\phi)$.

We assume that the rescaled interaction $V_\Lambda(\phi)$ is local on a length scale of order $\Lambda^{-1}$. More precisely, we assume that there exist constants $\gamma$ and $c$, independent of $L$ and $\Lambda$, such that
\begin{align}
    \left\|\left[\frac{\delta^2V_\Lambda(\psi)}{\delta\psi_x\delta\psi_y}\right]_{\psi=T_{ABC}^\lambda\phi}\right\|_{L_4(p_\Lambda)} \!\!\!\!\leq \left\|\frac{\delta V_\Lambda(\phi)}{\delta\phi_x}\right\|_{L_2(p_\Lambda)} \!\!\gamma e^{-c\Lambda|x-y|}.\label{eq:rg_loc_assumption}
\end{align}

\end{assumption}

\noindent\textit{Remark.}
For the conditionally local distributions satisfying Assumption~\ref{assump:data_dist}, the conditional interaction $U_{{\rm loc},z}(\phi)$ is assumed to be local for each latent variable $z$. Therefore, the rescaled interaction $V_{\Lambda,z}(\phi)$ obtained by the RG flow of $p_{\rm data}(\phi|z)$ is assumed to satisfy Eq.~\eqref{eq:rg_loc_assumption}. In this paper, we further assume that $V_{\Lambda,z}(\phi)$ satisfies the locality bound uniformly in $z$; namely, the constants $\gamma$ and $c$ in Eq.~\eqref{eq:rg_loc_assumption} can be chosen independently of $z$.

Here, the locality in Assumption~\ref{assump:rg_loc_int} is given in terms of the rescaled interaction $V_\Lambda(\phi)=U_\Lambda(\sqrt{K_\Lambda}\phi)$, not the Polchinski RG interaction $U_\Lambda$ itself. This is because the interaction entering the RGFM probability flow is the rescaled one, $U_\Lambda(\sqrt{K_\Lambda}\phi)$ (see Eq.~\eqref{eq:eff_action}). We note that the effect of the rescaling $\sqrt{K_\Lambda}$ is given by a smoothing convolution with characteristic length scale $\Lambda^{-1}$:
\begin{align}
    \left(\sqrt{K_\Lambda}\phi\right)_x &= \int_{\Omega_L} d^dy\, R_\Lambda(x-y) \phi_y,\\
    R_\Lambda(x) &= \frac{\Lambda^d}{(2\pi)^{\frac{d}{2}}} \exp\left(-\frac{m^2}{2\Lambda^2}-\frac{\Lambda^2x^2}{2}\right).
\end{align}
Since $U_\Lambda$ is expected to be quasi-local on the scale $\Lambda^{-1}$ due to the quasi-locality of the exact RG, this additional smoothing would not introduce interactions over longer distances than $\Lambda^{-1}$. We therefore argue that assuming quasi-locality of $V_\Lambda$ with length scale $O(\Lambda^{-1})$ is physically reasonable and does not constitute a strong additional restriction. 

It is worthwhile to comment on quasi-locality in the exact RG \cite{morris2000,morris2001,kopper2007,kopper2007a,rosten2012}. In exact RG, quasi-locality for local UV actions is typically formulated by requiring the effective vertex functions to admit a Taylor expansion in the dimensionless external momenta $p_i/\Lambda$ around zero momentum, corresponding to an all-orders derivative expansion \cite{morris1994a}. In exact RG analyses, such quasi-locality is generally assumed, while it has been established in certain cases. For massive scalar theories, perturbative results establish the quasi-locality through bounds on momentum derivatives and the decay of effective vertices in position space \cite{kopper2007,kopper2007a}. Assumption~\ref{assump:rg_loc_int} strengthens this general picture in two respects. First, it imposes a uniform exponential bound directly on the Hessian in position space. Second, it requires this bound for the rescaled interaction $V_\Lambda$ relevant to RGFM, rather than only for the Wilsonian interaction $U_\Lambda$.

Lastly, for the conditionally local distributions, we additionally assume that the latent variable remains locally predictable along the RG flow, with the associated length scale $\Lambda^{-1}$. We formulate this requirement in terms of the CMI between the latent variable and the exterior region.

\begin{assumption}\label{assump:rg_loc_cmi}
    Latent predictability during the RG flow.---
    Let $p_\Lambda(\phi)$ be the distribution obtained by RG diffusion from a conditionally local data distribution with latent variable $z$, and let $p_\Lambda(\phi,z)$ denote the corresponding joint distribution. We define $I_\Lambda(Z\!:\!C|AB)$ as the CMI between $z$ and the field in region $C$, conditioned on the field in $A\cup B$, evaluated with respect to $p_\Lambda(\phi,z)$.

    We assume that there exist constants $\gamma$ and $c$, independent of $\Lambda$ and $L$, such that, for any local-region decomposition into $A$, $B$, and $C$,
    \begin{align}
        I_\Lambda(Z\!:\!C|AB) \leq \gamma e^{-c\Lambda l_B}.
    \end{align}

\end{assumption}

\noindent\textit{Remark.}
We remark that this assumption does not require the marginalized distribution $p_{\Lambda}(\phi)$ itself to have a finite Markov length during the flow. More precisely, the length scale  $(c\Lambda)^{-1}$ characterizes the local predictability of the latent variable $z$ and should not be confused with the Markov length defined through the marginal distribution $p_{\Lambda}$.
 The latter may become macroscopic or even diverge in the middle of the flow without violating Assumption~\ref{assump:rg_loc_cmi}~\cite{hu2026}.

\subsection{\label{ssec:la_of_v}Local approximation of velocity fields}

As described in Sec.~\ref{sec:lgm_with_rgfm}, in local generative modeling, we replace the RGFM velocity field $v(\phi,t)$~\eqref{eq:vx_rgfm} with a local velocity field $v_{{\rm loc},t}(\phi)$, whose component at $x\in\Omega_L$ depends only on the field configuration within a local neighborhood of position $x$. To formulate this notion precisely, in this section, we define the local approximation of a velocity field under a probability distribution and quantify its error. Then, we show that the resulting local velocity field coincides with the minimizer of a patchwise flow-matching objective, thereby providing a direct connection between the local approximation and flow matching in practice. Finally, we introduce a Lipschitz-continuity assumption on $v_{{\rm loc},t}$, which is used in Sec.~\ref{ssec:lability_of_flow} to control the propagation of the local approximation error along the generative ODE flow in Eq.~\eqref{eq:fm_ode}.

First, we define the local approximation of the velocity field and its error on local region $A$ as follows. 

\begin{definition}\label{def:la_of_velocity}
    Approximation of a velocity field on a local region.---
    Let $p(\phi)$ be a probability distribution over fields $\phi$ on $\Omega_L$, and let $v(\phi)$ be a velocity field on $\Omega_L$ that depends on $\phi$. We define the local approximation of $v(\phi)$ on region $A$, $v_{\rm loc}^A$, which is the velocity field on $A$ that depends on the local field configuration $\phi_{AB}$, as 
    \begin{align}
        v_{{\rm loc}}^A(\phi_{AB}) 
        &= \mathbb E_{\phi_C\sim p(\phi_C|\phi_{AB})} [v^A(\phi_{AB},\phi_C)].
    \end{align}
    Here, $v^A$ is the restriction of the velocity field $v$ to region $A$.
\end{definition}

\begin{definition}\label{def:la_error_A}
    Approximation error on a local region.---
    Let $p$ be a distribution over $\phi$ and $v$ be a velocity field.  
    We define the local approximation error of $v$ on region $A$ by 
    \begin{align}
        \Delta_A(p, v; l_B) := \left(\mathbb E_{\phi\sim p(\phi)}[||v^A(\phi) - v_{\rm loc}^A(\phi_{AB})||_A^2]\right)^\frac{1}{2},
    \end{align}
    where $v_{\rm loc}^A$ is the local approximation of $v$ on region $A$ defined in Def.~\ref{def:la_of_velocity}. Here, $\|\cdot\|_A$ denotes the $L_2$ norm of a field on region $A$ defined as $\|f\|_A=\left(\int_A d^dx |f_x|^2\right)^{\frac{1}{2}}$.
\end{definition}

To see that the local velocity field $v_{\rm loc}^A$ approximates the velocity field $v$, it is useful to establish the following proposition. In short, it states that $v_{\rm loc}^A$ is the local field on region $A$ that minimizes the $L_2(p)$ distance from $v^A$.

\begin{proposition}\label{prop:la_minimizer}
    Property of the local approximation.---
    Let $p$ be a distribution over $\phi$, $v$ be a velocity field, and $v_{\rm loc}^A$ be its local approximation on region $A$. Then, for any local velocity field $w^A(\phi_{AB})$ on region $A$ that depends only on the local configuration $\phi_{AB}$, the following inequality holds:
    \begin{align}
        \mathbb E_{p(\phi)}\left[\left\|v^A-v_{\rm loc}^A\right\|_A^2\right] \leq \mathbb E_{p(\phi)}\left[\left\|v^A-w^A\right\|_A^2\right].\label{eq:vloc_ineq}
    \end{align}
\end{proposition}

\begin{proof}
    To see this, we rewrite the right-hand side of Eq.~\eqref{eq:vloc_ineq} as 
    \begin{widetext}
    \begin{align}
        \mathbb E_{p(\phi)}\left[\left\|v^A-w^A\right\|_A^2\right] 
        &= \int d\phi\, p(\phi) \left(\left\|v^A(\phi)\right\|^2_A -2 \langle v^A(\phi)\cdot w^A(\phi_{AB}) \rangle_A + \left\|w^A(\phi_{AB})\right\|^2_A \right)
        \\
        &= \int d\phi\, p(\phi)\left\|v^A(\phi)\right\|_A^2 + \int d\phi_{AB}\, p(\phi_{AB}) \left\|w^A(\phi_{AB})-v_{\rm loc}^A(\phi_{AB})\right\|_A^2\nonumber\\
        &\qquad - \int d\phi_{AB}\,p(\phi_{AB})  \left\|v_{\rm loc}^A(\phi_{AB}) \right\|_A^2. 
    \end{align}
    \end{widetext}
    Here, $\langle \cdot,\cdot\rangle_A$ is the inner product of two fields on region $A$ defined as $\langle f,g\rangle_A = \int_{A} d^dx\, f_x g_x$. From this expression, it is clear that the expectation value $\mathbb E_{p(\phi)}\left[\left\|v^A-w^A\right\|_A^2\right]$ is minimized when the local field $w^A$ coincides with the local approximation $v_{\rm loc}^A$.
\end{proof}

Having defined the local velocity field on local region $A$, $v_{\rm loc}^A$, we define the local approximation $v_{\rm loc}$ and its error on the full domain $\Omega_L$ by patching together the local velocity fields.

\begin{definition}\label{def:la_vel}
    Local approximation of a velocity field on $\Omega_L$.---
    Unless otherwise specified, we assume that the full domain $\Omega_L$ is partitioned into disjoint local regions ${A_i}$ of equal size as $\Omega_L=\bigsqcup_i A_i$. We assume that each $A_i$ is surrounded by a local buffer region $B_i$ of width $l_B$ as in Fig.~\ref{fig:lrd}(a). Then, we define the local velocity field $v_{\rm loc}(\phi)$ on the full domain $\Omega_L$ that approximates $v(\phi)$ by patching together the local approximations $v_{\rm loc}^{A_i}$ of $v(\phi)$ on all regions $A_i$.
\end{definition}

\begin{definition}\label{def:la_error}
    Local approximation error on the full domain $\Omega_L$.---
    Let $p$ be a distribution over $\phi$, $v$ be a velocity field, and $v_{\rm loc}$ be its local approximation on $\Omega_L$ defined in Definition~\ref{def:la_vel}. We define the local approximation error of $v$ as  
    \begin{align}
        \Delta_{\Omega_L}(p,v;l_B) &= \left(\mathbb E_{\phi\sim p(\phi)}\left[ ||v(\phi) - v_{\rm loc}(\phi)||_{\Omega_L}^2 \right]\right)^{\frac{1}{2}}\\
        &= \left(\sum_{A_i} \Delta_{A_i}^2(p,v;l_B)\right)^{\frac{1}{2}}.\label{eq:loc_approx_error_global}
    \end{align}
    Here, $\Delta_{A_i}(p,v;l_B)$ is the local approximation error on each local patch $A_i$. 
\end{definition}

To connect the local approximations to FM, we establish the following proposition. In short, the local approximation of the FM velocity field $v_t$ coincides with that used in local generative modeling. 

\begin{proposition}
    Relation between local approximation and local generative modeling.---
    Let $p_t$ be the probability flow associated with either the standard FM or RGFM, and let $v(\phi,t)$ be the corresponding velocity field. Recall that $v(\phi,t)$ is obtained by minimizing the FM training objective involving the conditional velocity field $u_t(\phi|\phi_0)$, as in Eq.~\eqref{eq:fm_cost}.

    Then, the local approximation $v_{{\rm loc}}(\phi,t)$ of $v(\phi,t)$ on the full domain coincides with the minimizer of the following local-FM patchwise training objective:
    \begin{widetext}
    \begin{align}
    \mathcal L_{\rm FM}^{\rm loc}(\theta) &= \mathbb E_{t,\phi_0\sim p_{\rm data}(\phi_0), \phi\sim p_t(\phi|\phi_0)} \left[ \sum_{A_i} \left\|
    v^{A_i}_{{\rm loc},\theta}(\phi,t) - u_t^{A_i}(\phi|\phi_0)
    \right\|_{A_i}^2\right],\label{eq:fm_la}
    \end{align}
    \end{widetext}
    where $u_t^{A_i}$ denotes the restriction of $u_t$ to region $A_i$.
\end{proposition}

\begin{proof}
    Since the training objective in Eq.~\eqref{eq:fm_la} is a patchwise objective, it is sufficient to prove the proposition for a fixed local patch $A$. From Proposition~\ref{prop:la_minimizer}, we have
    \begin{widetext}
    \begin{align}
        v_{\rm loc}^A 
        &= {\rm arg} \min_{w^A:{\rm local}} \mathbb E_{p_t(\phi)}\left[\left\| v^A(\phi,t) - w^A(\phi_{AB}) \right\|_A^2\right]\\
        &= {\rm arg} \min_{w^A:{\rm local}} \left( 
            \int d\phi\,p_t(\phi)||w^A(\phi_{AB})||_A^2 
            - 2\int d\phi \,p_t(\phi)\langle v^A(\phi,t),w^A(\phi_{AB})\rangle_A
        \right)\\
        &= {\rm arg} \min_{w^A:{\rm local}} \left(
            \int d\phi\,p_t(\phi)||w^A(\phi_{AB})||_A^2 
            - 2\int d\phi\,d\phi_0 \,p_t(\phi)p_t(\phi_0|\phi)\langle u_t^A(\phi|\phi_0),w^A(\phi_{AB})\rangle_A
        \right)\\
        &= {\rm arg} \min_{w^A:{\rm local}} \Bigl(
            \mathbb E_{\phi_0\sim p_{\rm data}(\phi_0), \phi\sim p_t(\phi|\phi_0)} \left[
                ||w^A(\phi_{AB})-u_t^A(\phi|\phi_0)||^2_A
            \right]
        \Bigr).
    \end{align}
    \end{widetext}
    Therefore, the local approximation of the FM velocity field $v(\phi,t)$ on local region $A$ coincides with the minimizer of $\mathbb E_{\phi_0\sim p_{\rm data}(\phi_0), \phi\sim p_t(\phi|\phi_0)}\left[\left\| w^A(\phi_{AB})\!-\!u_t^A(\phi|\phi_0)\right\|^2_A\right]$, which proves the proposition.
\end{proof}

\noindent\textit{Remark.}
As is clear from the training objective~\eqref{eq:local_res_rgfm_cost}, the local velocity field used in the local generative modeling with RGFM coincides with the local approximation of the rescaled velocity field $\widetilde{\mathcal R}_t v$ under the rescaled distribution $\tilde p_t$.

Lastly, we impose a regularity assumption on the local approximation of the RGFM velocity field.

\begin{assumption}\label{assump:Lip_conti}
    Lipschitz-continuity condition for the local approximation of RGFM velocity fields.---
    Let $p_{\rm data}$ be a local or conditionally local data distribution satisfying Assumption~\ref{assump:data_dist}, and let $p_\Lambda$ and $v_\Lambda$ denote the corresponding RGFM distribution and velocity field at RG wavenumber scale $\Lambda$. We assume that there exists a constant $M$ such that the local approximation $v_{{\rm loc},\Lambda}(\phi)$ of $v_\Lambda(\phi)$ is Lipschitz continuous in $\phi$ with Lipschitz constant $M$:
    \begin{align}
        \left\|v_{{\rm loc},\Lambda}(\phi) - v_{{\rm loc},\Lambda}(\phi')\right\|_{\Omega_L} 
        \leq M \left\|\phi-\phi'\right\|_{\Omega_L}.
    \end{align}
\end{assumption}

\noindent\textit{Remark.}
In Propositions~\ref{prop:v_derV} and \ref{prop:vLam_cond}, below, we show that the RGFM velocity field $v_\Lambda(\phi)$ is proportional, up to a $\Lambda$-dependent coefficient, to the functional derivative of the interaction part of $p_\Lambda$. Moreover, in Proposition~\ref{prop:bound3_loc_data_dist}, we will prove that its $L_2(p_\Lambda)$ norm is bounded uniformly in $\Lambda$ as $\mathbb E_{p_\Lambda}\left[\left|v_\Lambda(\phi)\right|_{\Omega_L}^2\right]\leq cL^d(\ln L)^2$, where $c$ is independent of $\Lambda$. These facts alone do not directly imply Lipschitz continuity of $v_{{\rm loc},\Lambda}$. Nevertheless, provided that the interaction part $V_\Lambda$ does not develop singular or strongly irregular dependence on the field configuration along the RG flow, it is reasonable to expect that $v_{{\rm loc},\Lambda}$ remains sufficiently regular, with a Lipschitz constant that can be chosen uniformly in $\Lambda$.

\subsection{\label{ssec:lability_of_v}Local approximability of velocity fields in the RGFM}

\subsubsection{\label{sssec:lability_of_v_loc}Local approximability of velocity fields in the RGFM for local data distributions}

In the preceding subsections, we formulated the assumptions required for our analysis and defined local approximations of the RGFM velocity fields $v_t$. In this subsection, we show that, for local data distributions, the corresponding approximation error $\Delta_{\Omega_L}(p_t, v_t; l_B)$ decays exponentially with the buffer width $l_B$, with a characteristic length scale set by the running RG scale $\Lambda(t)^{-1}$. More precisely, we establish the following theorem.

\begin{theorem}\label{thm:LA_loc_data}
    Bound on the local approximation error for local data distributions.---
    Let $p_{\rm data}$ be a local data distribution satisfying Assumption~\ref{assump:data_dist}, and define $p_t=p_\Lambda$ and $v_t=v_\Lambda$ as the probability distribution and the velocity field~\eqref{eq:vx_rgfm} for the RGFM at RG wavenumber scale $\Lambda(t)$. Then, there exist constants $\gamma$ and $c$, independent of $L$ and $t$, such that
    \begin{align}
        \Delta_{\Omega_L}(p_t, v_t; l_B) \leq \gamma e^{-c\Lambda l_B} L^{\frac{d}{2}} \ln L.\label{eq:thm:LA_loc_data}
    \end{align}
\end{theorem}

To prove Theorem~\ref{thm:LA_loc_data}, we begin with the following propositions.

\begin{proposition}\label{prop:v_derV}
    RGFM velocity field and the interaction functional.---
    Let $p_\Lambda$ be the probability distribution obtained by the RG diffusion of a local data distribution $p_{\rm data}$. We denote the interaction part of 
    $p_\Lambda$ by $V_\Lambda(\phi)$. Then, the velocity field $v_\Lambda$ for the RG diffusion can be written as 
    \begin{align}
        v_\Lambda(\phi) &= \frac{1}{\Lambda^2\tau} \frac{\delta V_\Lambda(\phi)}{\delta\phi}. \label{eq:v_derV}
    \end{align} 
\end{proposition}

\begin{proof}
    From Eqs.~\eqref{eq:eff_action_erg} and \eqref{eq:eff_action}, the score of the distribution $p_\Lambda(\phi)\propto e^{-S_\Lambda(\phi)}$ can be written as 
    \begin{align}
        \frac{\delta}{\delta\phi} \ln p_\Lambda(\phi) &= \frac{\delta}{\delta\phi}\left(- \frac{1}{2}\int_x \phi_x(-\nabla^2+m^2)\phi_x + V_\Lambda(\phi)\right)\\
        &= (\nabla^2-m^2)\phi - \frac{\delta V_\Lambda(\phi)}{\delta\phi},
    \end{align}
    where $V_\Lambda$ is related to the interaction part of the exact RG $U_\Lambda$ by $V_\Lambda(\phi) = U_\Lambda(\sqrt{K_\Lambda}\phi)$. Together with the expression of the velocity field $v(\phi,t)$ in Eq.~\eqref{eq:vx_rgfm}, we obtain the desired equality~\eqref{eq:v_derV}.
\end{proof}

\begin{proposition}\label{prop:phix_4_bound}
    Moment bound during the RG.---
    The moment $\left\|\phi_x - \langle \phi_x\rangle_{p_\Lambda}\right\|_{L_4(p_\Lambda)}$ is uniformly bounded. There exists a constant $M$ independent of $\Lambda$ such that, for any $x\in \Omega_L$, 
    \begin{align}
        \left\|\phi_x - \langle \phi_x\rangle_{p_\Lambda}\right\|_{L_4(p_\Lambda)} \leq M.
    \end{align}
\end{proposition}

\begin{proof}
    We provide the proof in Appendix~\ref{proof:prop:phix_4_bound}.
\end{proof}

With these preliminaries, we prove the following proposition.
\begin{proposition}\label{prop:bound1_loc_data_dist}
    Bound on the local approximation error on a local region.---
    Let $v_\Lambda$ be the RGFM velocity field for a local data distribution. 
    Then, there exist constants $\gamma$ and $c$, and a $(d-1)$-th degree polynomial ${\rm Poly}(x)$ independent of $L$ and $\Lambda$ such that, for any local regions $A$ and $B$, 
    \begin{align}
        \Delta_A(p_\Lambda, v_\Lambda; l_B) &\leq \gamma\, {\rm Poly}(c\Lambda l_B) e^{-c\Lambda l_B}\nonumber\\
        &\qquad \!\times\! \frac{1}{\Lambda^d} \left(\int_A d^dx\, \mathbb E_{\phi\sim p_\Lambda} \left[\left|v_{\Lambda,x}(\phi)\right|^2\right]\right)^{\frac{1}{2}}\!.\label{eq:bound1_loc_data_dist}
    \end{align}
\end{proposition}

\begin{proof}
    Here, we give the overview of the proof. See Appendix~\ref{proof:prop:bound1_loc_data_dist} for details. The key inequality is Eq.~\eqref{eq:rg_loc_assumption} in the RG locality assumption~\ref{assump:rg_loc_int}.

    Since $\delta V_\Lambda/\delta \phi\equiv u_\Lambda(\phi)$ is proportional to $v_\Lambda(\phi)$, it is sufficient to prove the following inequality:
    \begin{align}
        \Delta_A(p_\Lambda, u_\Lambda; l_B) &\leq \gamma\, {\rm Poly}(c\Lambda l_B) e^{-c\Lambda l_B}\nonumber\\
        &\qquad \!\times\!\frac{1}{\Lambda^d} \left(\int_A d^dx\, \mathbb E_{\phi\sim p_\Lambda} \left[\left|u_{\Lambda,x}(\phi)\right|^2\right]\right)^{\frac{1}{2}}\!.\label{eq:lae_uLam}
    \end{align}
    
    By considering a local velocity field on $A$ by setting $\phi_C=\langle \phi_C\rangle_{p_\Lambda}$, we obtain the bound on the local approximation error as 
    \begin{widetext}        
    \begin{align}
        \Delta^2_A\left(p_\Lambda, u_\Lambda; l_B\right)
        \leq \int_A d^dx \left(\int_C d^dy \int_0^1 d\lambda\,\left\| \left(\phi_y-\langle\phi_y\rangle_{p_\Lambda}\right) \left[\frac{\delta^2 V_\Lambda}{\delta\psi_x\delta\psi_y}\right]_{T_{ABC}^\lambda\phi}\right\|_{L^2(p_\Lambda)}\right)^{2}.
    \end{align}
    With Eq.~\eqref{eq:rg_loc_assumption} and Proposition~\ref{prop:phix_4_bound}, this bound implies
    \begin{align}
        \Delta^2_A\left(p_\Lambda, u_\Lambda; l_B\right) 
        &\leq \int_A d^dx \left(M\gamma \left\|\frac{\delta V_\Lambda}{\delta\phi_x}\right\|_{L_2(p_\Lambda)} \int_C dy\, e^{-c\Lambda|x-y|} \right)^2\\
        & \leq \int_A d^dx \left(M\gamma \left\|u_{\Lambda,x}\right\|_{L_2(p_\Lambda)} \int_{l_B}^\infty ds\, e^{-c\Lambda s} \, 2d\cdot(2s)^{d-1} \right)^2\\
        & = \int_A d^dx \left(
            M\gamma \left\|u_{\Lambda,x}\right\|_{L_2(p_\Lambda)} \frac{d\cdot 2^d}{c^d\Lambda^d} e^{-c\Lambda l_B} \int_0^\infty ds\, e^{-s}\left(s + c\Lambda l_B \right)^{d-1}
        \right)^2\\
        & = \left(\frac{\gamma'\, {\rm Poly}\left(c\Lambda l_B\right)e^{-c\Lambda l_B}}{\Lambda^d}\right)^2 \int_A d^dx\, \mathbb E_{\phi\sim p_\Lambda}\left[|u_{\Lambda,x}|^2\right].
    \end{align}
    \end{widetext}
    Here, we bound the integral with respect to $y\in C$ by integrating over cubic shells centered at $x\in A$. We also define the polynomial ${\rm Poly}(x)$ by ${\rm Poly}(x) = \int_0^\infty ds\,e^{-s}(s+x)^{d-1}$ and absorb constants independent of $L$ and $\Lambda$ into $\gamma'$. This leads to the inequality~\eqref{eq:lae_uLam}.
\end{proof}

\begin{corollary}\label{cor:bound2_loc_data_dist}
    Bound on the local approximation error on the full domain.---
    Let $v_\Lambda$ be the velocity field of the RGFM for local data distributions. Then, $v_\Lambda$ satisfies
    \begin{align}
        \Delta_{\Omega_L}(p_\Lambda, v_\Lambda; l_B) &\leq \gamma\, {\rm Poly}(c\Lambda l_B) e^{-c\Lambda l_B}\nonumber\\
        &\quad\, \times\!\frac{1}{\Lambda^d} \left(\int_{\Omega_L}\! d^dx\, \mathbb E_{\phi\sim p_\Lambda} \left[\left|v_{\Lambda,x}(\phi)\right|^2\right]\right)^{\frac{1}{2}}\!\!,\label{eq:LAE_pv_loc_dist1}
    \end{align}
    where the constants $\gamma$ and $c$ and the polynomial ${\rm Poly}(x)$ are independent of $L$ and $\Lambda$.
\end{corollary}

\begin{proof}
    We obtain the inequality by adding the square of the inequality~\eqref{eq:bound1_loc_data_dist} for disjoint local regions $A_i$ with $\Omega_L = \sqcup_i A_i$.
\end{proof}

To evaluate the right-hand side of Eq.~\eqref{eq:LAE_pv_loc_dist1}, we first prove the following lemma.

\begin{lemma}\label{lem:vk_phi0k}
    Expectation representation of the RGFM velocity field.---
    Let $v_\Lambda(\phi)$ be the velocity field of the RGFM for any data distribution $p_{\rm data}$ at RG wavenumber scale $\Lambda(t)$. Then, the $k$-th wavenumber component $v_{\Lambda,k}$ can be expressed as 
    \begin{align}
        v_{\Lambda,k} (\phi) = \frac{\partial_t K_{tk}}{2(1-K_{tk})}\left\langle-\phi_k + \frac{1}{\sqrt{K_{tk}}} \phi_{0k}\right\rangle_{\phi_0\sim p_t(\phi_0|\phi)},\label{eq:vk_rgfm_for_bound}
    \end{align}
    where $p_t(\phi|\phi_0)$ is the conditional probability for the RG diffusion~\eqref{eq:rg_diffusion}.
\end{lemma}

\begin{proof}
    Combining Eqs.~\eqref{eq:vk_rgfm} and \eqref{eq:sk_rgfm}, we obtain Eq.~\eqref{eq:vk_rgfm_for_bound}.
\end{proof}

Then, we obtain the following bound on the right-hand side of Eq.~\eqref{eq:LAE_pv_loc_dist1}. 

\begin{proposition}\label{prop:bound3_loc_data_dist}
    Bound on $L_2(p_\Lambda)$-norm of RGFM velocity field.---
    Let $v_\Lambda(\phi)$ be the velocity field of the RGFM for a local data distribution $p_{\rm data}$. 
    Then, there exists a constant $\gamma$ independent of $L$ and $\Lambda$ such that 
    \begin{align}
        \frac{1}{\Lambda^d} \left(\int_{\Omega_L} d^dx\, \mathbb E_{\phi\sim p_\Lambda} \left[ |v_{\Lambda,x}(\phi)|^2\right]\right)^{\frac{1}{2}} 
        \leq \gamma L^{\frac{d}{2}} \ln L.
    \end{align}
\end{proposition}

\begin{proof}
    Applying conditional Jensen's inequality to Eq.~\eqref{eq:vk_rgfm_for_bound} in Lemma~\ref{lem:vk_phi0k}, we bound the second moment of $v_{\Lambda,k}(\phi)$ as 
    \begin{widetext}
    \begin{align}
        \mathbb E_{\phi\sim p_\Lambda} \left[\left\|v_{\Lambda,k}(\phi)\right\|^2\right]
        &\leq \frac{(\partial_t K_{tk})^2}{4(1-K_{tk})^2} 
        \mathbb E_{\phi_0\sim p_{\rm data}, \phi\sim p_t(\phi|\phi_0)}\!\left[\left\|-\phi_k \!+\! \frac{\phi_{0k}}{\sqrt{K_{tk}}}\right\|^2\right]\\
        &= \frac{(\partial_t K_{tk})^2}{4K_{tk}}\left(\mathbb E_{\phi_0\sim p_{\rm data}}\left[\|\phi_{0k}\|^2\right] + \frac{K_{tk}}{1-K_{tk}}\frac{1}{k^2+m^2}\right).\label{eq:vk_2mom_bound}\\ 
        &= \left(\frac{k^2+m^2}{\tau\Lambda^2}\right)^2 e^{-\frac{k^2+m^2}{\Lambda^2}} \left\|\phi_{0k}\right\|_{L_2(p_{\rm data})}^2 + \left(\frac{k^2+m^2}{\tau\Lambda^2}\right)^2 \frac{e^{-2\frac{k^2+m^2}{\Lambda^2}}}{1-e^{-\frac{k^2+m^2}{\Lambda^2}}}\frac{1}{k^2+m^2},
    \end{align}
    where we use the expression of the RG diffusion~\eqref{eq:pt_diff} and $K_{tk}=\exp[-(k^2+m^2)/\Lambda(t)^2]$. Therefore, we obtain the bound
    \begin{align}
        \frac{1}{\Lambda^d} \left(\int_{\Omega_L} d^dx\, \mathbb E_{\phi\sim p_\Lambda} \left[ |v_x(\phi)|^2\right]\right)^{\frac{1}{2}} 
        &= \frac{1}{\Lambda^d} \left( L^d \int_{|k|<\pi} \frac{d^dk}{(2\pi)^d}\, \mathbb E_{\phi\sim p_\Lambda} \left[ |v_k(\phi)|^2\right] \right)^{\frac{1}{2}}\\
        &\leq \frac{L^{\frac{d}{2}}}{\tau} \left(
            \int_{|k|<\pi} \frac{d^dk}{(2\pi)^d} \left[  \frac{f_1\left(\frac{k^2+m^2}{\Lambda^2}\right)\left\|\phi_{0k}\right\|_{L^2(p_{\rm data})}^2}{(k^2+m^2)^d} +  \frac{f_2\left(\frac{k^2+m^2}{\Lambda^2}\right)}{(k^2+m^2)^{d+1}} \right]
            \right)^\frac{1}{2}.\label{eq:integral_k_1}
    \end{align}
    \end{widetext}
    Here, we define dimensionless functions $f_1(x)=x^{d+2}e^{-x}$ and $f_2(x)=x^{d+2}e^{-2x}/(1-e^{-x})$. We note that $f_1(x)$ and $f_2(x)$ are bounded for $x\geq 0$. Since the integral domain is restricted to $|k|<\pi$, the integral in Eq.~\eqref{eq:integral_k_1} gives $O(L^0)$ contributions. Together with the fact that $\tau^{-1} = O(\ln L)$, we obtain
    \begin{align}
        \frac{1}{\Lambda^d} \left(\int_{\Omega_L} d^dx\, \mathbb E_{\phi\sim p_\Lambda} \left[ |v_x(\phi)|^2\right]\right)^{\frac{1}{2}} 
        \leq \gamma L^{\frac{d}{2}} \ln L,
    \end{align}
    where $\gamma$ is a constant independent of $L$ and $\Lambda$. This completes the proof.
\end{proof}

Finally, we prove the main theorem. 
\begin{proof}[Proof of Theorem~\ref{thm:LA_loc_data}.]
    From Corollary~\ref{cor:bound2_loc_data_dist} and Proposition~\ref{prop:bound3_loc_data_dist}, we obtain the following bound on $\Delta_{\Omega_L}(p_t, v_t; l_B)$ as 
    \begin{align}
        \Delta_{\Omega_L}(p_t, v_t; l_B) \leq \gamma {\rm Poly}(c\Lambda l_B) e^{-c\Lambda l_B} L^{\frac{d}{2}}\ln L,
    \end{align}
    where the constants $\gamma$ and $c$ and the polynomial ${\rm Poly}(x)$ are independent of $L$ and $\Lambda$. Here, by slightly weakening the exponential factor by taking $c'<c$, one can bound $\gamma {\rm Poly}(c\Lambda l_B)e^{-c\Lambda l_B}$ as 
    \begin{align}
        \gamma {\rm Poly}(c\Lambda l_B) e^{-c\Lambda l_B} \leq \gamma' e^{-c'\Lambda l_B}, 
    \end{align}
    where $\gamma'$ is independent of $L$ and $\Lambda$. Therefore, we obtain the bound for $\Delta_{\Omega_L}(p_t, v_t; l_B)$ as 
    \begin{align}
        \Delta_{\Omega_L}(p_t, v_t; l_B) \leq \gamma' e^{-c'\Lambda l_B} L^{\frac{d}{2}} \ln L,
    \end{align}
    where $\gamma'$ and $c'$ are independent of $L$ and $\Lambda$. This proves Eq.~\eqref{eq:thm:LA_loc_data} in Theorem~\ref{thm:LA_loc_data}.
\end{proof}

\noindent\textit{Remark.}
By construction, the constants $\gamma$ and $c$ in the bound~\eqref{eq:thm:LA_loc_data} in Theorem~\ref{thm:LA_loc_data} depend only on the locality parameters $c$ and $\gamma$ in the RG locality assumption~\ref{assump:rg_loc_int} and the uniform bound for the moment $\|\phi_{x}\|_{L_4(p_{\rm data})}$. We use this fact to prove the local approximability of the velocity fields for conditionally local data distributions in the next section.

\subsubsection{\label{sssec:lability_of_v_cloc}Local approximability of velocity fields in the RGFM for conditionally local data distributions}

In the preceding subsection, we showed that, for local data distributions, the local approximation error of the RGFM velocity field decays exponentially with the buffer width as in Theorem~\ref{thm:LA_loc_data}. In this subsection, we extend this result to the conditionally local data distributions and show that similar exponential decay persists despite the presence of latent-variable-induced nonlocal correlations. More precisely, we establish the following theorem.

\begin{theorem}\label{thm:LA_cloc_data}
    Bound on the local approximation error for conditionally local data distributions.---
    Let $p_{\rm data}$ be a conditionally local data distribution satisfying Assumption~\ref{assump:data_dist}. Let $p_\Lambda = p_t$ and $v_\Lambda = v_t$ denote the RGFM probability distribution and velocity field~\eqref{eq:vx_rgfm}, respectively, at RG wavenumber scale $\Lambda(t)$. Then, there exist constants $\gamma$ and $c$, independent of $L$ and $t$, such that
    \begin{align}
        \Delta_{\Omega_L}(p_t, v_t; l_B) \leq \gamma e^{-c\Lambda l_B} L^{\frac{2d}{3}} \ln L.\label{eq:thm:LA_cloc_data}
    \end{align}
\end{theorem}

\noindent\textit{Remark. } As shown below, the contribution $L^{2d/3}$ is a rough evaluation and can be replaced by $L^{d/2}$ for translation invariant systems. 

To prove the theorem, we begin with the following propositions.
\begin{proposition}\label{prop:vLam_cond}
    RGFM velocity field equality for conditionally local data distributions.---
    Let $p_{\rm data}(\phi)$ be a conditionally local data distribution with latent variable $z$. Let $p_{\Lambda}(\phi)$ and $p_{\Lambda}(\phi|z)$ be the distributions obtained by applying the RG diffusion to $p_{\rm data}$ and $p_{\rm data}(\phi|z)$, and let $v_\Lambda(\phi)$ and $v_\Lambda(\phi|z)$ be the corresponding RGFM velocity fields, respectively. Then, $p_\Lambda(\phi)$ is given by 
    \begin{align}
        p_\Lambda(\phi) &= \int dz\, p(z) p_{\Lambda}(\phi|z),\label{eq:pLam_cond}
    \end{align}
    and the following equality holds
    \begin{align}
        v_\Lambda(\phi) &= \mathbb E_{z|\phi} \left[v_\Lambda(\phi|z)\right].\label{eq:vLam_cond}
    \end{align}
    Here, $z|\phi$ denotes that $z$ is sampled from the conditional distribution $p_\Lambda(z|\phi)=p_\Lambda(\phi,z)/p_\Lambda(\phi) = p_\Lambda(\phi|z)p(z)/p_\Lambda(\phi)$.
\end{proposition}

\begin{proof}
    Since the RG diffusion~\eqref{eq:rg_diffusion} is independent of the latent $z$, $\phi\sim p_\Lambda(\phi)$ can be generated by applying the RG diffusion to $\phi_0 \sim p_{\rm data}(\phi|z)$ with $z\sim p(z)$. This shows that the joint distribution of $\phi$ and $z$ at RG scale $\Lambda$ is 
    \begin{align}
        p_\Lambda(\phi,z) &= p_\Lambda(\phi|z) p(z),
    \end{align}
    which proves Eq.~\eqref{eq:pLam_cond}.

    From Eq.~\eqref{eq:pLam_cond}, we have the following score equality.
    \begin{align}
        \frac{\delta}{\delta\phi} \ln p_\Lambda(\phi) &= \frac{1}{p_\Lambda(\phi)}\frac{\delta}{\delta\phi} \int dz\,p(z)p_\Lambda(\phi|z)\\
        &= \int dz\, \frac{p(z)p_\Lambda(\phi|z)}{p_\Lambda(\phi)} \frac{\delta}{\delta\phi} \ln p_\Lambda(\phi|z)\\
        &= \mathbb E_{z|\phi} \left[\frac{\delta}{\delta\phi} \ln p_\Lambda(\phi|z)\right].\label{eq:score_eq}
    \end{align}
    As shown in the proof of Proposition~\ref{prop:v_derV}, the scores of $p_\Lambda(\phi)$ and $p_\Lambda(\phi|z)$ are related to $v_\Lambda(\phi)$ and $v_\Lambda(\phi|z)$ as 
    \begin{align}
        v_\Lambda(\phi) &= \frac{1}{\Lambda^2\tau}\left( (\nabla^2-m^2)\phi - \frac{\delta}{\delta\phi}\ln p_\Lambda(\phi) \right),\\
        v_\Lambda(\phi|z) &= \frac{1}{\Lambda^2\tau}\left((\nabla^2-m^2)\phi - \frac{\delta}{\delta\phi}\ln p_\Lambda(\phi|z)\right),
    \end{align}
    respectively. Together with the score equality~\eqref{eq:score_eq}, we obtain Eq.~\eqref{eq:vLam_cond}.
\end{proof}

Using the velocity-field identity~\eqref{eq:vLam_cond}, we obtain the following bound on the local approximation error $\Delta_A(p_\Lambda, v_\Lambda; l_B)$.

\begin{proposition}\label{prop:LAE_A_condloc_1}
    Bound on the local approximation error on a local region.---
    Let $A$, $B$, and $C$ be a local-region decomposition of $\Omega_L$. We denote by $v_{\Lambda,{\rm loc}}^A(\phi_{AB}|z)$ the local approximation of the velocity field $v_\Lambda(\phi|z)$ under $p_\Lambda(\phi|z)$. We denote the local approximation error of $v_\Lambda(\phi|z)$ under $p_\Lambda(\phi|z)$ on region $A$ by $\Delta_A(p_\Lambda(\cdot|z), v_\Lambda(\cdot|z); l_B)$. Then, the local approximation error of $v_\Lambda$ under $p_\Lambda$ satisfies 
    \begin{widetext}
    \begin{align}
        \Delta_A(p_\Lambda, v_\Lambda; l_B)^2 &\leq 2 \mathbb E_{p(z)}\left[
            \Delta_A(p_\Lambda(\cdot|z), v_\Lambda(\cdot|z);l_B)^2
        \right] + 2 \mathbb E_{p_\Lambda(\phi)} \left[\left\|
            \mathbb E_{z|\phi}[v_{\Lambda,{\rm loc}}^A(\phi_{AB}|z)]
            - \mathbb E_{z|\phi_{AB}}[v^A_{\Lambda,{\rm loc}}(\phi_{AB}|z)]
        \right\|_A^2\right].\label{eq:LAE_condloc}
    \end{align}
    \end{widetext}
\end{proposition}

\begin{proof}
    We provide the proof in Appendix~\ref{proof:prop:LAE_A_condloc_1}. 
    Here, we note that the first term in the right-hand side of Eq.~\eqref{eq:LAE_condloc} represents the average local approximation error of $v_\Lambda(\phi|z)$ under $p_\Lambda(\phi|z)$. On the other hand, the second term represents the error incurred when estimating $\mathbb E_{z|\phi}[v_{\Lambda,{\rm loc}}^A]$ from $\phi_{AB}$ alone.
\end{proof} 

By the RG locality assumption~\ref{assump:rg_loc_int}, the first term in the right-hand side of Eq.~\eqref{eq:LAE_condloc} can be bounded by an exponential factor $e^{-c\Lambda l_B}$ as in Proposition~\ref{prop:bound1_loc_data_dist}. Under the RG assumption on the latent variable~\ref{assump:rg_loc_cmi}, the second term can be bounded as follows. 

\begin{proposition}\label{prop:bound_latent_pred}
    Cost for the latent estimation error.--- 
    For any $q>2$, the second term in the right-hand side of Eq.~\eqref{eq:LAE_condloc} can be bounded as 
    \begin{align}
    &\mathbb E_{p_\Lambda(\phi)} \left[\left\|
        \mathbb E_{z|\phi}[v_{\Lambda,{\rm loc}}^A(\phi_{AB}|z)]
        - \mathbb E_{z|\phi_{AB}}[v_{\Lambda,{\rm loc}}^A(\phi_{AB}|z)]
    \right\|_A^2\right]\nonumber\\
    &\qquad \leq \gamma^2 e^{-2c\Lambda l_B} \int_A d^dx\, \left(\mathbb E_{\phi,z} \left[ \left| v_{\Lambda,x}(\phi|z)\right|^q \right] \right)^{\frac{2}{q}},\label{eq:bound_latent_pred}
    \end{align}
    where the constants $\gamma$ and $c$ are independent of $L$ and $\Lambda$. 
\end{proposition}

\begin{proof}
    Since the proof is technical, we provide the proof in Appendix~\ref{proof:prop:bound_latent_pred}.
\end{proof}

From Propositions~\ref{prop:LAE_A_condloc_1} and \ref{prop:bound_latent_pred}, we obtain the following corollary.

\begin{corollary}\label{cor:LAE_cond_loc_cor}
    Bound on the local approximation error on the full domain.---
    For $q>2$, there exist constants $\gamma_1$, $\gamma_2$, $c_1$, and $c_2$, independent of $L$ and $\Lambda$, such that 
    \begin{widetext}
    \begin{align}
        \Delta_{\Omega_L}(p_\Lambda, v_\Lambda; l_B)^2 
        &\leq \left(\gamma_1 e^{-c_1\Lambda l_B} L^{\frac{d}{2}}\ln L\right)^2 + \gamma_2^2 e^{-2c_2\Lambda l_B} \int_{\Omega_L} d^dx\, \left(\mathbb E_{\phi,z} \left[ \left| v_{\Lambda,x}(\phi|z)\right|^q \right] \right)^{\frac{2}{q}}.\label{eq:LAE_cond_loc_cor}
    \end{align}
    \end{widetext}
\end{corollary}

\begin{proof}
    By adding the inequality~\eqref{eq:LAE_condloc} for all disjoint local regions $A_i$ with $\Omega_L=\sqcup_i A_i$ and using the inequality~\eqref{eq:bound_latent_pred}, we obtain 
    \begin{widetext}
    \begin{align}
        \Delta_{\Omega_L}(p_\Lambda, v_\Lambda; l_B)^2
        & \leq \mathbb E_{p(z)} \left[\Delta_{\Omega_L}^2(p_\Lambda(\cdot|z), v_\Lambda(\cdot|z); l_B)\right] + \gamma_2^2 e^{-2c_2\Lambda l_B} \int_{\Omega_L} d^dx\, \left(\mathbb E_{\phi,z} \left[ \left| v_{\Lambda,x}(\phi|z)\right|^q \right] \right)^{\frac{2}{q}},
    \end{align}
    \end{widetext}
    where $\gamma_2$ and $c_2$ are constants independent of $L$ and $\Lambda$.

    Since $p_\Lambda(\phi|z)$ is obtained by the RG diffusion of the local data distribution $p_{\rm data}(\phi|z)$, the results of Theorem~\ref{thm:LA_loc_data} apply. Namely, $\Delta_{\Omega_L}(p_\Lambda(\cdot|z), v_\Lambda(\cdot|z); l_B)$ can be bounded by $\gamma_1 e^{-c_1\Lambda l_B} L^{\frac{d}{2}}\ln L$ with constants $\gamma_1$ and $c_1$ independent of $L$ and $\Lambda$. Further, due to Assumption~\ref{assump:rg_loc_int}, we can choose these constants independently of the latent $z$ (cf. Remark below Theorem~\ref{thm:LA_loc_data}). This leads to the inequality~\eqref{eq:LAE_cond_loc_cor}.
\end{proof}

Finally, we prove the main theorem~\ref{thm:LA_cloc_data}. 

\begin{proof}[Proof of Theorem~\ref{thm:LA_cloc_data}.]
    We provide the proof in Appendix~\ref{proof:thm:LA_cloc_data}. We note that the scaling $L^{2d/3}$ arises from the estimate of the integral in Eq.~\eqref{eq:LAE_cond_loc_cor}, which can be evaluated in a manner similar to the proof of Proposition~\ref{prop:bound1_loc_data_dist}. For translation invariant systems, this integral scales as $O(L^d)$. In general, however, an $O(L^d)$ bound cannot be guaranteed, and we instead use an $O(L^{2d(1-1/q)})$ bound with $2<q\leq 4$ for the integral. Setting $q=3$ then yields the bound in Theorem~\ref{thm:LA_cloc_data}.
\end{proof}

Combining Theorems~\ref{thm:LA_loc_data} and \ref{thm:LA_cloc_data}, we obtain the following statement, which is one of the main theorems in this paper.

\begin{theorem}\label{thm:LA_rgfm}
    Unified bound on the local approximation of the RGFM velocity fields.---
    Let $p_{\rm data}$ be either a local or conditionally local data distribution satisfying Assumption~\ref{assump:data_dist}. 
    Denote the local approximation error for the velocity field in the RGFM at time $t$ by $\Delta_{\Omega_L}(p_t, v_t; l_{B,t})$. 

    Then, there exist constants $\gamma$ and $c$ independent of $L$ and $t$, such that 
    \begin{align}
        \Delta_{\Omega_L}(p_t, v_t; l_{B,t}) \leq \gamma e^{-c\Lambda l_{B,t}} L^{\alpha} \ln L.\label{eq:LA_rgfm}
    \end{align}
    Here, $\alpha=d/2$ for local data distributions and $\alpha=2d/3$ for conditionally local distributions.

    In particular, for any $\varepsilon>0$, one can choose $l_{B,t} = O(\Lambda(t)^{-1}\ln(L^\alpha/\varepsilon))$ so that
    \begin{align}
        \Delta_{\Omega_L}(p_t, v_t; l_{B,t}) \leq \varepsilon.
    \end{align} 
\end{theorem}

\begin{proof}
    From Theorems~\ref{thm:LA_loc_data} and \ref{thm:LA_cloc_data}, the local approximation error $\Delta_{\Omega_L}(p_t, v_t; l_{B,t})$ can be bounded as 
    \begin{align}
        \Delta_{\Omega_L}(p_t, v_t; l_{B,t}) \leq \gamma e^{-c\Lambda l_{B,t}} L^{\alpha} \ln L,\label{eq:LA_rgfm_prf1}
    \end{align} 
    where $\gamma$ and $c$ are independent of $L$ and $\Lambda$. Here, we recall that $\alpha=d/2$ for a local data distribution and $\alpha=2d/3$ for a conditionally local distribution in general. 
    
    Taking 
    \begin{align}
        l_{B,t} \geq \frac{1}{c\Lambda(t)} \ln\left(\frac{\gamma L^\alpha \ln L}{\varepsilon}\right),
    \end{align}
    which is $O(\Lambda(t)^{-1}\ln(L^\alpha/\varepsilon))$, we obtain $\Delta_{\Omega_L}(p_t,v_t;l_{B,t})\leq \varepsilon$.
\end{proof}

\subsection{\label{ssec:lability_of_flow}Local approximability of probability flows in the RGFM}

As established in Theorem~\ref{thm:LA_rgfm}, for any prescribed accuracy $\varepsilon>0$, the RGFM velocity field $v_t$ admits a local approximation $v_{{\rm loc},t}$ constructed using patches with buffer width $O\left(\Lambda(t)^{-1}\ln(L^\alpha/\varepsilon)\right)$. In this subsection, we show that this local velocity field $v_{{\rm loc},t}$ can also be used to approximate the generative ODE flow in the RGFM. Specifically, let $p_{\rm data}^{\rm loc}$ denote the distribution generated by the ODE flow~\eqref{eq:fm_ode} obtained by replacing $v_t$ with $v_{{\rm loc},t}$. Then, we prove that $p_{\rm data}^{\rm loc}$ remains close to the target distribution $p_{\rm data}$ in terms of the 2-Wasserstein distance. Below, we prove the following theorem, which is one of the main results in this work.

\begin{theorem}\label{thm:rgfm_flow_stability}
    Local approximability of the generative ODE flow in the RGFM.---
    Let $p_{\rm data}$ be either a local or conditionally local data distribution satisfying Assumption~\ref{assump:data_dist}. Let $(p_t)_{t\in[0,1]}$ be the corresponding RGFM probability path generated by the velocity field $v_t$, with $p_0=p_{\rm data}$ and $p_1=p_{\rm GS}$. At each time $t$, choose a buffer width $l_{B,t}$ and let $v_{{\rm loc},t}$ be the corresponding local approximation of $v_t$. Suppose that $v_{{\rm loc},t}$ satisfies the Lipschitz-continuity condition in Assumption~\ref{assump:Lip_conti} with Lipschitz constant $M$.

    Starting from $\phi_1\sim p_1=p_{\rm GS}$, evolve the local ODE
    \begin{align}
        \frac{d\phi_t}{dt}=v_{{\rm loc},t}(\phi_t)
    \end{align}
    backward from $t=1$ to $t=0$, and denote the resulting distribution of $\phi_0$ by $p_{\rm data}^{\rm loc}$. 
    
    Then, for any prescribed accuracy $\varepsilon>0$, one can choose $l_{B,t} = O(\Lambda(t)^{-1}\ln(L^\alpha/\varepsilon))$ so that the 2-Wasserstein distance between $p_{\rm data}$ and $p_{\rm data}^{\rm loc}$ satisfies
    \begin{align}
        W_2\left(p_{\rm data},p_{\rm data}^{\rm loc}\right)
        \leq \varepsilon.\label{eq:local_rgfm_w2_bound}
    \end{align}
    Here, $\alpha=d/2$ for local data distributions and $\alpha=2d/3$ for conditionally local distributions.
\end{theorem}

To prove this theorem, we first prove the stability of the probability flow $p_t$ generated by the ODE flow~\eqref{eq:fm_ode}.

\begin{theorem}\label{thm:ode_stability}
    2-Wasserstein bound on the probability flow with local approximation.---
    Suppose that the time evolution of $p_t$ is generated by the ODE flow~\eqref{eq:fm_ode} associated with $v_t$. At each time $t$, we choose the width of the buffer region $B$ to be $l_{B,t}$ and define the corresponding local velocity field $v_{{\rm loc},t}$. We denote by $p_t^{\rm loc}$ the time evolution of the same initial distribution $p_{t=0}$ generated by the ODE flow~\eqref{eq:fm_ode} associated with $v_{{\rm loc},t}$.

    Assume that, for each $t$, $v_{{\rm loc},t}(\phi)$ is Lipschitz continuous with respect to $\phi$, with Lipschitz constant $M(t)$. Then, the 2-Wasserstein distance between $p_t$ and $p_t^{\rm loc}$ is bounded in terms of the local approximation error $\Delta_{\Omega_L}(p_t,v_t;l_{B,t})$ as
    \begin{align}
        W_2(p_t, p_t^{\rm loc}) \leq \int_0^t ds\, e^{\int_s^t dr\, M(r)} \Delta_{\Omega_L}(p_s, v_s; l_{B,s}).
    \end{align}
\end{theorem}

\begin{proof}
    We give the proof in Appendix~\ref{proof:thm:ode_stability}.
\end{proof}

In the generative process of RGFM, one first samples $\phi_{t=1}$ from the Gaussian distribution $p_{\rm GS}$. One then evolves the sample backward from $t=1$ to $t=0$ along the ODE flow~\eqref{eq:fm_ode} generated by the velocity field $v(\phi,t)$ given in Eqs.~\eqref{eq:vx_fm} or \eqref{eq:vx_rgfm}. This procedure produces a sample $\phi_{t=0}$ following the data distribution $p_{\rm data}$. These backward evolutions can be expressed as a forward probability flow. If we define $p'_t = p_{1-t}$, then, $p'_0=p_{\rm GS}$, whereas $p'_1=p_{\rm data}$. In particular, the time evolution of $p'_t$ from $t=0$ to $t=1$ is generated by the velocity field $v'_t(\phi)=-v(\phi,1-t)$. Theorem~\ref{thm:ode_stability} can therefore be applied directly to the generative probability flow of RGFM.

Combining Theorem~\ref{thm:ode_stability} with the local-approximation bound established in Theorem~\ref{thm:LA_rgfm}, we obtain the main result, Theorem~\ref{thm:rgfm_flow_stability} as follows.

\begin{proof}[Proof of Theorem~\ref{thm:rgfm_flow_stability}.]
    As shown in the proof of Theorem~\ref{thm:LA_rgfm}, the local approximation error $\Delta_{\Omega_L}(p_t, v_t; l_{B,t})$ can be bounded as 
    \begin{align}
        \Delta_{\Omega_L}(p_t, v_t; l_{B,t}) \leq \gamma e^{-c\Lambda(t) l_{B,t}} L^{\alpha}\ln L,
    \end{align} 
    where $\gamma$ and  $c$ are independent of $L$ and $\Lambda$, and $\alpha=d/2 (2d/3)$ for local (conditionally local) distributions. Also, from the Lipschitz-continuity assumption~\ref{assump:Lip_conti}, $v_{{\rm loc},t}(\phi)$ is Lipschitz continuous with Lipschitz constant $M$. 

    Therefore, from the flow stability theorem~\ref{thm:ode_stability}, the 2-Wasserstein distance between $p_{\rm data}$ and $p_{\rm data}^{\rm loc}$ can be bounded as 
    \begin{align}
        W_2\left(p_{\rm data},p_{\rm data}^{\rm loc}\right)
        &\leq  \int_0^t ds\, e^{\int_s^t dr\, M} \gamma e^{-c\Lambda(s) l_{B,s}} L^{\alpha} \ln L\\
        &\leq \left(\gamma e^M \cdot L^\alpha\ln L \right) \int_0^t ds e^{-c\Lambda(s) l_{B,s}}.
    \end{align}
    Thus, by taking $l_{B,t}$ as 
    \begin{align}
        l_{B,t} \geq \frac{1}{c\Lambda(t)} \ln \left(
            \gamma e^M \cdot\frac{L^\alpha\ln L}{\varepsilon}
        \right),
    \end{align}
    we obtain $W_2(p_{\rm data}, p_{\rm data}^{\rm loc})\leq \varepsilon$, completing the proof.
\end{proof}

Lastly, we note that the flow stability results in Theorems~\ref{thm:ode_stability} and \ref{thm:rgfm_flow_stability} can be extended to the SDE formulation. We give a brief overview and defer all the proofs to the Appendices~\ref{proof:prop:equiv_sde}, \ref{proof:thm:sde_stability}, and \ref{proof:thm:rgfm_flow_stability_sde}. First, the deterministic ODE flow~\eqref{eq:fm_ode} admits an equivalent stochastic differential equation (SDE) \cite{song2021b,albergo2025}. Specifically, the following proposition holds.

\begin{proposition}\label{prop:equiv_sde}
    Equivalent SDE formulation of the ODE time evolution.---
    Suppose that the time evolution of $p_t$ is generated by the ODE flow~\eqref{eq:fm_ode} associated with a velocity field $v_t(\phi)$. Then, the following SDE for the field $\phi_t$ generates the same marginal probability path $p_t$:
    \begin{align}
        d\phi_t = \left[v_t(\phi_t) + \frac{g^2(t)}{2} s_t(\phi_t)\right] dt + g(t) dw_t \label{eq:sde1}
    \end{align}
    Here, $s_t(\phi)=(\delta/\delta\phi)\ln p_t(\phi)$ is the score of $p_t$, $g(t)\geq 0$ is an arbitrary time-dependent diffusion coefficient, and $dw_t$ is a Wiener process satisfying $\mathbb E[dw_t]=0$ and $\mathbb E[dw_{t,x}dw_{t,x'}]=dt\,\delta(x-x')$.
\end{proposition}

In this SDE formulation, one can locally approximate the SDE~\eqref{eq:sde1} by replacing $v_t$ and $s_t$ with their local approximations, $v_{{\rm loc},t}$ and $s_{{\rm loc},t}$, respectively. In fact, the KL divergence between $p_t$ and its locally approximated counterpart $\hat p_t^{\rm loc}$ can be bounded by the locality of $v_t$ and $s_t$ as follows. 

\begin{theorem}\label{thm:sde_stability}
    KL divergence bound for the locally approximated SDE.---
    Suppose that the time evolution of $p_t$ is generated by the SDE~\eqref{eq:sde1}, and assume that $g(t)>0$ on the time interval under consideration.
    At each time $t$, choose the width of the buffer region $B$ to be $l_{B,t}$, and denote the corresponding local approximations of $v_t$ and $s_t$ under the distribution $p_t$ by $v_{{\rm loc},t}$ and $s_{{\rm loc},t}$, respectively.

    Let $\hat p_t^{\rm loc}$ be the marginal distribution generated from the same initial distribution $p_0$ and SDE by replacing $v_t$ and $s_t$ with $v_{{\rm loc},t}$ and $s_{{\rm loc},t}$, respectively. Then, by choosing a suitable diffusion coefficient $g(t)$, the KL divergence between $p_t$ and $\hat p_t^{\rm loc}$ is bounded as
    \begin{align}
        D_{\rm KL}(p_t||\hat p_t^{\rm loc}) &\leq \int_0^t dr\, 
        \Delta_{\Omega_L}(p_r,v_r;l_{B,r})\Delta_{\Omega_L}(p_r,s_r;l_{B,r}).\label{eq:sde2}
    \end{align}
\end{theorem}

Importantly, in the RGFM, the score $s_t$ coincides with the velocity field $v_t$ up to a local contribution $\propto (-\nabla^2+m^2)\phi$ (see, e.g., Eq.~\eqref{eq:vx_rgfm}). In particular, their local approximation errors are related to each other as 
\begin{align}
    \Delta_{\Omega_L}(p_t,s_t;l_{B,t}) = \Lambda(t)^2\tau \Delta_{\Omega_L}(p_t,v_t;l_{B,t}).\label{eq:st_vt_rgfm}
\end{align}
Therefore, combining with the flow stability theorem for SDE~\ref{thm:sde_stability}, we obtain the following theorem.

\begin{theorem}\label{thm:rgfm_flow_stability_sde}
    Local approximability of the generative SDE flow equivalent to RGFM.---
    Let $p_{\rm data}$ be either a local or conditionally local data distribution satisfying Assumption~\ref{assump:data_dist}. Let $(p_t)_{t\in[0,1]}$ be the corresponding RGFM probability path generated by the velocity field $v_t$, with $p_0=p_{\rm data}$ and $p_1=p_{\rm GS}$. At each time $t$, choose a buffer width $l_{B,t}$ and denote the local approximation of $v_t$ and $s_t$ by $v_{{\rm loc},t}$ and $s_{{\rm loc},t}$. 

    Let us define $s=1-t$ and $\tilde{\phi}_s=\phi_{1-s}$.  Starting from $\tilde{\phi}_0\sim p_{\rm GS}$, evolve the local backward SDE
    \begin{align}
        d\tilde{\phi}_s &= \left[-v_{{\rm loc},1-s}(\tilde{\phi}_s) + \frac{g^2(1-s)}{2} s_{{\rm loc},1-s}(\tilde{\phi}_s)\right] ds\nonumber\\
        &\qquad + g(1-s) dw_s \label{eq:sde1_local}
    \end{align}
    from $s=0$ to $s=1$, and denote the resulting distribution of $\tilde{\phi}_1$ by $\hat p_{\rm data}^{\rm loc}$. Here, we set $g(t)=\sqrt{2/(\Lambda(t)^2\tau)}$.
    
    Then, for any prescribed accuracy $\varepsilon>0$, one can choose $l_{B,t} = O(\Lambda(t)^{-1}\ln (L^\alpha/\varepsilon))$ so that the KL divergence between $p_{\rm data}$ and $\hat p_{\rm data}^{\rm loc}$ satisfies
    \begin{align}
        D_{\rm KL}\left(p_{\rm data}||\hat p_{\rm data}^{\rm loc}\right)
        \leq \varepsilon.\label{eq:local_rgfm_KL_bound}
    \end{align}
    Here, $\alpha=d/2$ for local data distributions and $\alpha=2d/3$ for conditionally local distributions.
\end{theorem}

\noindent\textit{Remark 1.}
In Theorem~\ref{thm:rgfm_flow_stability_sde}, the Lipschitz-continuity assumption~\ref{assump:Lip_conti} is not required to control the propagation of the trajectory error, and there seems to be an advantage to using the SDE over the ODE flow in data generation. Nevertheless, this advantage applies at the continuous-time level. In practical sampling, the SDE must be discretized, and poor regularity or a large effective Lipschitz constant of the drift may require a finer time discretization to maintain numerical stability and accuracy.

\noindent\textit{Remark 2.} 
In contrast to Theorem~\ref{thm:ode_stability}, the bound for the distance between $p_t$ and $\hat p_t^{\rm loc}$~\eqref{eq:sde2} can be obtained without assuming the Lipschitz continuity of $v_{{\rm loc},t}$ in Assumption~\ref{assump:Lip_conti}. We remark that the probability flow $\hat p_t^{\rm loc}$, which is obtained by replacing $v_t$ and $s_t$ in the SDE~\eqref{eq:sde1}, is different from $p_t^{\rm loc}$ in Theorem~\ref{thm:ode_stability}. Therefore, this theorem does not imply that the KL divergence between $p_t$ and $p_t^{\rm loc}$ in Theorem~\ref{thm:ode_stability} can also be bounded by a local approximation error without the Lipschitz-continuity condition.

\section{\label{sec:num_exp}Numerical experiments}

In this section, we numerically investigate local generative modeling with RGFM and compare it with conventional FM under the same local-network constraints. We first consider one-dimensional distributions with local and conditionally local structures and examine whether their long-range correlations can be reproduced using a fixed-size receptive field. We then apply the method to natural-image generation at resolutions of $64\times64$ and $256\times256$, demonstrating both the computational scalability of local-patch RGFM and its limitations. Detailed numerical settings, including the network architectures and training parameters, are provided in the Appendix~\ref{app:num_exp}. The code and trained models used in the numerical experiments can be found at \url{https://github.com/kantamasuki/Local_RGFM}.

\begin{figure*}
    \centering
    \includegraphics[width=17.9cm]{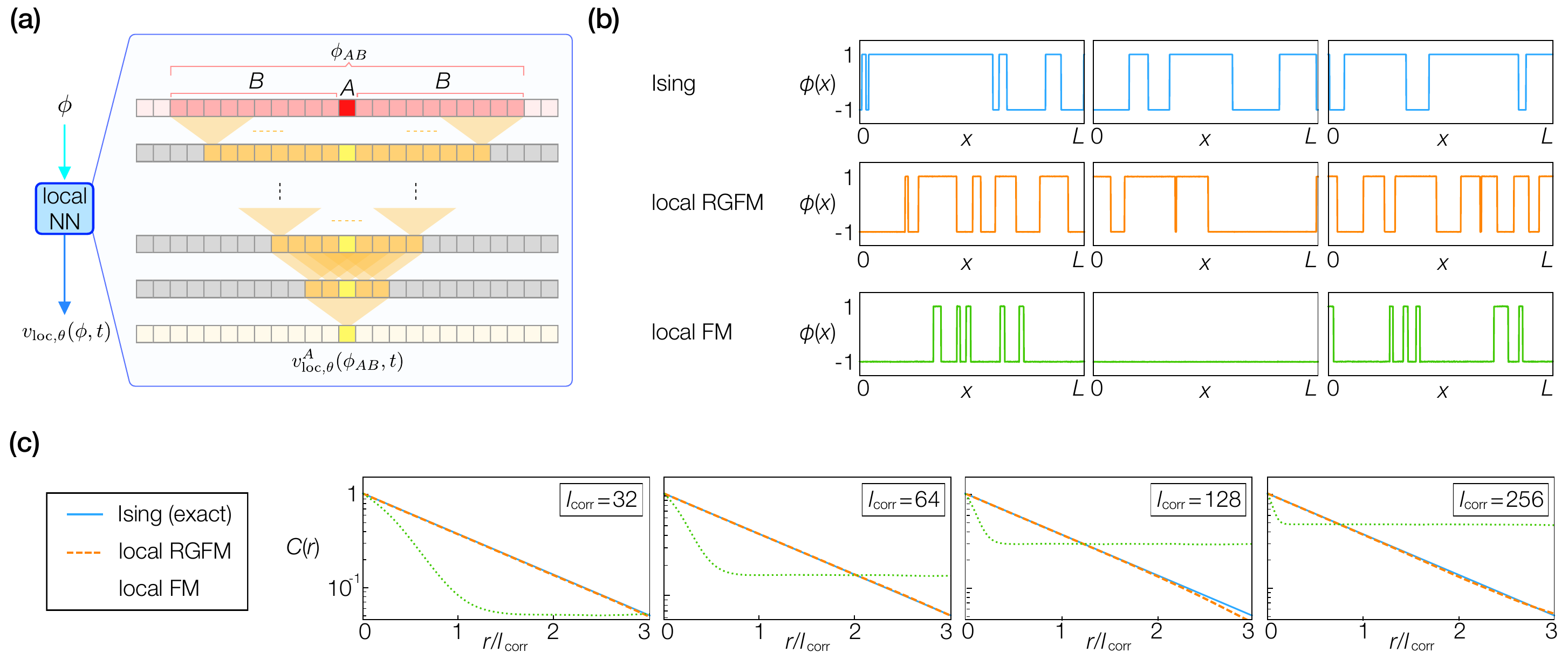}
    \caption{\label{fig:num_ising}
    (a) Architecture of the convolutional neural network (CNN) used to learn the local velocity field in Secs.~\ref{ssec:ne_ising} and \ref{ssec:ne_cond_loc}. The finite receptive field of the network is indicated schematically.
    (b) Representative configurations sampled from the one-dimensional Ising model~\eqref{eq:ising1} and generated by local-CNN RGFM and FM for the correlation length $l_{\rm corr}=64$.
    (c) Correlation function $C(r)=\langle\phi_x\phi_{x+r}\rangle$ computed from the generated samples, compared with the exact result in Eq.~\eqref{eq:ising_corr}. We set the system size $L=1024$ and the Ising coupling $K$ such that the correlation length takes the values $l_{\rm corr}=32$, $64$, $128$, and $256$. 
    }
\end{figure*}

\subsection{\label{ssec:ne_ising}One-dimensional local distribution: Ising model}

As a first demonstration, we consider the one-dimensional Ising model as an example of a data distribution specified by a strictly local action. A spin configuration is denoted by $\phi=(\phi_1,\phi_2,\ldots,\phi_L)$, where $\phi_x\in\{\pm 1\}$, and its probability distribution $p_{\rm Ising}(\phi)$ is given by
\begin{align}
    p_{\rm Ising}(\phi) &= \frac{1}{Z} e^{-S_{\rm Ising}(\phi)},\label{eq:ising1}\\
    S_{\rm Ising}(\phi) &= -K \sum_{x=1}^{L-1} \phi_{x}\phi_{x+1}.\label{eq:ising2}
\end{align}
Under open boundary conditions, the two-point correlation function $C(r)=\langle\phi_x\phi_{x+r}\rangle$ is exactly given by
\begin{align}
    C(r) = \exp\left(-\frac{r}{l_{\rm corr}}\right),\ l_{\rm corr}= - \frac{1}{\ln[\tanh(K)]}.\label{eq:ising_corr}
\end{align}
Although the Ising spins are discrete variables, we regard each configuration $\phi\in\{\pm 1\}^L$ as a point embedded in $\mathbb R^L$ and define a continuous probability flow in this ambient space. Since $S_{\rm Ising}$ contains only nearest-neighbor interactions, it is a strictly local action. The preceding analysis therefore suggests that the corresponding rescaled RGFM path $\tilde p_t$ can be accurately approximated using a local velocity field.

Here, we model the local velocity field $v_{{\rm loc},\theta}$ by a simple convolutional neural network (CNN) illustrated in Fig.~\ref{fig:num_ising}(a). The CNN consists of $N_{\rm layer}=6$ convolutional layers with kernel size $H=5$, resulting in a receptive field of $N_{\rm layer}(H-1)+1=25$ sites. This corresponds to a single-site target region, $l_A=1$, surrounded by a buffer of width $l_B=12$ on each side. We note that, although the full field configuration $\phi$ and time $t$ are provided as inputs to the CNN, the finite receptive field ensures that the velocity predicted at each site depends only on its local neighborhood. Moreover, since the Ising action is translation invariant in the bulk, positional information of each site is not supplied to the CNN; we model the CNN as $v_{\theta}^{\rm CNN}(\phi,t)$.

We set the original system size to $L\!=\!1024$ and train the RGFM models on the successively decimated lattices with $L\!=\!1024$, $512$, $256$, $128$, $64$, $32$, and $16$. For comparison, we train the standard FM model on the original lattice with $L\!=\!1024$ using the same architecture and receptive-field size. We choose the Ising coupling $K$ such that the correlation length takes the values $l_{\rm corr}=32$, $64$, $128$, and $256$. 
We numerically solve the ODE flow~\eqref{eq:fm_ode} defined by the trained local velocity field $v_{\theta}^{\rm CNN}(\phi,t)$ using the midpoint method implemented in the \texttt{odeint} function of the \texttt{torchdiffeq} package. We discretize the time interval $[0,1]$ into $200$ time steps (see Appendix~\ref{app:num_exp}).

Figure~\ref{fig:num_ising}(b) shows typical samples generated by local-CNN RGFM and local-CNN FM. Local-CNN RGFM reproduces configurations with correlations extending far beyond the receptive field of the CNN. By contrast, when restricted to the same fixed-size receptive field, local-CNN FM fails to reproduce the correct long-range structure. Figure~\ref{fig:num_ising}(c) compares the correlation functions computed from the generated samples with the exact result in Eq.~\eqref{eq:ising_corr}. For all correlation lengths considered, local-CNN RGFM accurately reproduces the expected correlation function, whereas local-CNN FM exhibits substantial deviations at distances beyond the local receptive field.

\begin{figure*}
    \centering
    \includegraphics[width=17.9cm]{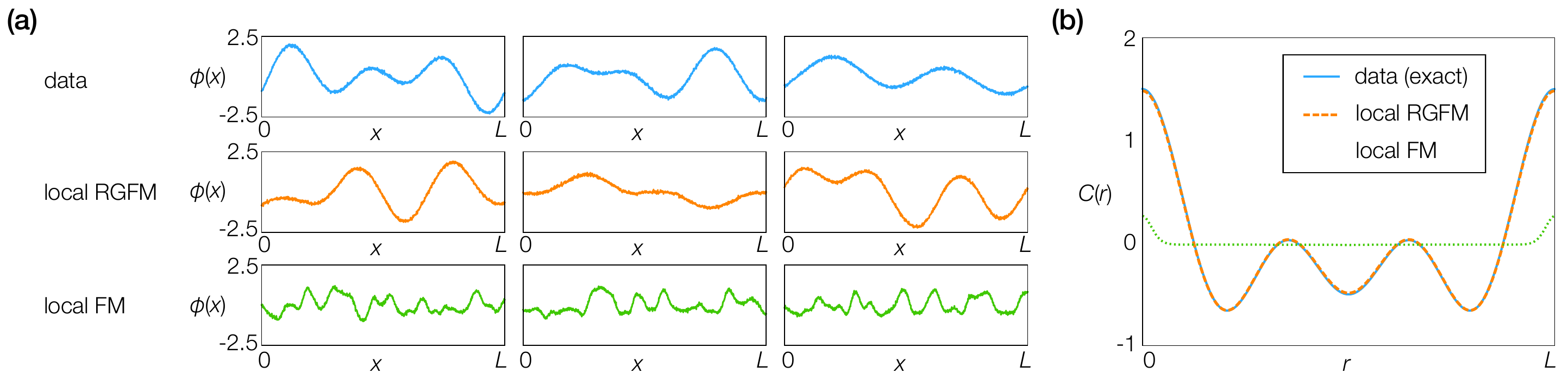}
    \caption{\label{fig:num_cld}
    (a) Representative configurations sampled from the conditionally local distribution in Eq.~\eqref{eq:cond_local_dist} and generated by local-CNN RGFM and FM. (b) Correlation function $C(r)=\langle\phi_x\phi_{x+r}\rangle$ computed from the generated samples, compared with the exact result in Eq.~\eqref{eq:cond_loc_cor}. The parameters are $L=1024$ and $\sigma=0.05$, and we use the same CNN as in Fig.~\ref{fig:num_ising}(a).
    }
\end{figure*}

\subsection{\label{ssec:ne_cond_loc}One-dimensional conditionally local distribution}

As a second demonstration, we consider a one-dimensional conditionally local distribution. Specifically, we consider a field $\phi=(\phi_1,\phi_2,\ldots,\phi_L)$ whose probability distribution is defined through the latent variable $z=(a_1,a_2,a_3,\eta_1,\eta_2,\eta_3)$ as
\begin{align}
    p_{\rm cond}(\phi) &= \int dz\, p(\phi|z) p(z),\label{eq:cond_local_dist}\\
    p(\phi|z) &= \prod_{x=1}^L \mathcal N(\phi_x; f_z(x), \sigma^2),\\
    f_z(x) &= \sum_{m=1}^3 a_m \cos\left(\frac{2m\pi x}{L} + \eta_m\right).
\end{align}
Here, the latent distribution $p(z)=p(a_1,a_2,a_3,\eta_1,\eta_2,\eta_3)$ is given by $a_m\sim\mathcal N(0,1)$ and $\eta_m\sim{\rm Uniform}[0,2\pi)$ for $m=1,2,3$. Since each phase $\eta_m$ is uniformly distributed, the marginal distribution $p_{\rm cond}(\phi)$ is invariant under lattice translations.

After integrating out $z$, the marginal distribution $p_{\rm cond}(\phi)$ possesses nonlocal correlations. In particular, its two-point correlation function $C(r)=\langle\phi_x\phi_{x+r}\rangle$ is given by
\begin{align}
    C(r) = \sum_{m=1}^3 \frac{1}{2}\cos\left(\frac{2m\pi r}{L}\right) + \delta_{0,r}\sigma^2.\label{eq:cond_loc_cor}
\end{align}
This correlation function oscillates without decaying as the separation $r$ increases, demonstrating that the marginal distribution contains long-range structure. Nevertheless, for fixed $z$, the field variables $\{\phi_x\}_{x=1,\ldots,L}$ are independent Gaussian variables with $\phi_x\sim\mathcal N(f_z(x),\sigma^2)$. The conditional distribution $p(\phi|z)$ is therefore strictly local. Moreover, when $\sigma$ is sufficiently small, it is reasonable to expect that the latent information can be inferred accurately from the waveform observed within a sufficiently large local region. Therefore, this distribution should provide a toy example of the conditionally local structure discussed in the preceding section, for which local generative modeling with RGFM is expected to be effective.

In the numerical experiments, we set $L\!=\!1024$ and $\sigma\!=\!0.05$ and use the same convolutional architecture and receptive-field size as in the Ising-model experiment (Fig.~\ref{fig:num_ising}(a)). Since $p_{\rm cond}(\phi)$ is translationally invariant, we again provide no explicit positional information to the CNN. For comparison, we train a conventional FM model using the same local architecture and receptive-field size.

Typical generated samples and the corresponding correlation functions are shown in Figs.~\ref{fig:num_cld}(a) and \ref{fig:num_cld}(b), respectively. Local-CNN RGFM reproduces both the smooth global waveform and the local Gaussian fluctuations using a fixed-size receptive field. It also accurately captures the oscillatory long-range correlation function of the marginal distribution. By contrast, local-CNN FM fails to maintain the global coherence of the waveform and consequently does not reproduce the correct long-range correlations.

\subsection{\label{ssec:ne_img}Image generation}

As an application to more realistic data, we apply local generative modeling with RGFM to image generation. We use images from the FFHQ dataset resized to $64\times64$ pixels. The local velocity field is modeled by a U-Net \cite{ronneberger2015a} equipped with self-attention at an intermediate resolution. Unlike the one-dimensional distributions considered above, the image distribution is not translationally invariant. To allow the local model to distinguish patches at different spatial positions, we augment each image patch with two additional channels encoding its spatial coordinates. Specifically, together with the three RGB channels of the image, we use the two spatial-coordinate channels as input, resulting in a five-channel input $(\phi^R,\phi^G,\phi^B,\phi^X,\phi^Y)$, where $\phi^X$ and $\phi^Y$ denote the $x$- and $y$-coordinates of each pixel, respectively. Further details of the network architecture, patch construction, and training procedure are provided in Appendix~\ref{app:num_exp}.

\begin{figure*}[t]
    \centering
    \includegraphics[width=17.9cm]{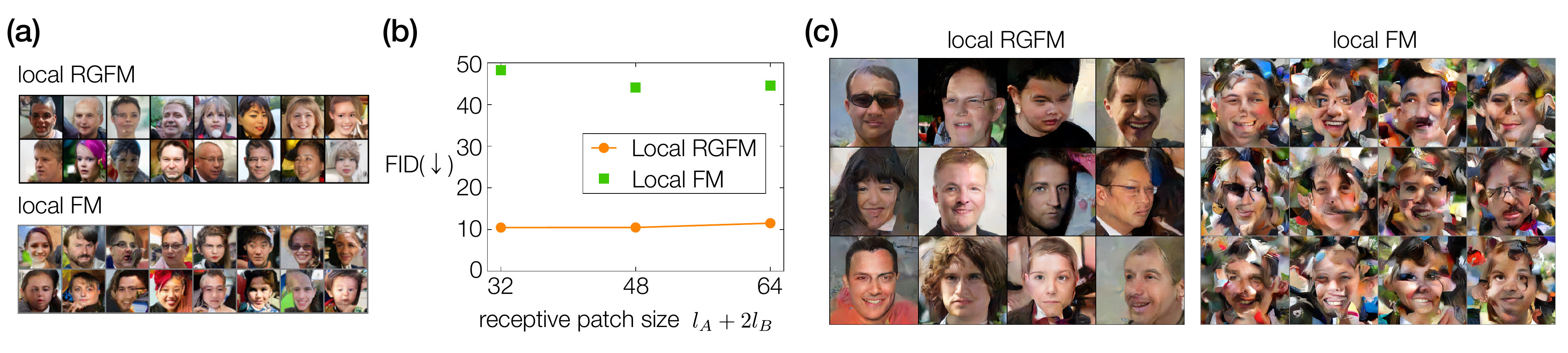}
    \caption{\label{fig:num_img}
    (a) Typical image samples generated by local generative modeling with the RGFM and the standard FM. The models are trained on the FFHQ dataset resized to a resolution of $64\times64$, using target patches of linear size $16$ and receptive patches of linear size $32$.
    (b) Fr\'echet inception distance (FID) scores \cite{heusel2017a,parmar2022} of the generated $64\times64$ images as a function of the receptive patch size $l_A+2l_B$ (see Fig.~\ref{fig:lrd} for the geometry). The target-patch size is fixed at $l_A=16$, while the receptive patch size is varied among $l_A+2l_B=32, 48$, and $64$. We generate $50{,}000$ images with each trained model.
    (c) Typical image samples generated by local generative modeling with the RGFM and the standard FM at a resolution of $256\times256$. The models are trained on a subset consisting of the first $5{,}000$ images in the FFHQ dataset, resized to $256\times256$, using local patches with receptive and target patch sizes of $32$ and $16$, respectively.
    }
\end{figure*}

Representative samples generated by local-patch RGFM and local-patch FM are shown in Fig.~\ref{fig:num_img}(a). The images generated by local-patch RGFM capture the global facial structure reasonably well, despite the fact that the velocity field at each position is predicted using only a local neighborhood. By contrast, local-patch FM under the same local-network constraint struggles to construct globally coherent facial images. We numerically solve the ODE flow~\eqref{eq:fm_ode} defined by the trained local velocity field $v_{\theta}^{\rm UNet}(\phi,t)$ using the midpoint method implemented in the \texttt{odeint} function of the \texttt{torchdiffeq} package. We discretize the time interval $[0,1]$ into $300$ time steps (see Appendix~\ref{app:num_exp}).

Figure~\ref{fig:num_img}(b) shows the Fr\'echet inception distance (FID) \cite{heusel2017a,parmar2022} as a function of the receptive patch size $l_A+2l_B$, with the target-patch size $l_A$ fixed. Increasing the receptive patch size does not substantially improve the performance of local FM, whereas local RGFM maintains a substantially lower FID over the entire range considered. Although a sufficiently large receptive patch should in principle contain enough information to approximate the conventional FM velocity field, the practical learnability of the velocity field can depend strongly on the choice of the probability path. The results suggest two complementary advantages of RGFM. First, the performance gap between RGFM and FM remains large even for the largest receptive patches considered, suggesting that the scale-by-scale generation scheme itself facilitates learning: large-scale structures are learned on coarse lattices before progressively finer degrees of freedom are introduced. Second, the performance of RGFM changes only weakly as the receptive patch size is reduced. This robustness is consistent with the locality properties established in Sec.~\ref{sec:loc_app}, where the characteristic locality length of the RGFM velocity field is shown to scale with the running RG length scale $\Lambda^{-1}$. Together, these results indicate that the scale-by-scale construction and the controlled locality of the RGFM probability path play distinct but complementary roles in local generative modeling.

We further apply local-patch RGFM and FM to the generation of $256\times256$ images. At this resolution, training a global model that evaluates the velocity field from the full configuration $\phi$ becomes prohibitively expensive. For example, in our implementation, introducing global self-attention at the first resolution of the U-Net requires more than $10\,{\rm GB}$ of GPU memory per image in a training batch. By contrast, the input and output dimensions of a local-patch model are determined by the sizes of the local fields $\phi_{\rm loc}$ and $v_{\rm loc}$ rather than by the full image size. The local formulation therefore makes it possible to train the velocity field without processing the entire high-resolution image simultaneously. In our numerical experiments, training requires approximately six days on a single NVIDIA RTX PRO 6000 Blackwell Max-Q GPU with 96 GB of memory.

Examples of the generated $256\times256$ images are shown in Fig.~\ref{fig:num_img}(c), where we solve the ODE~\eqref{eq:fm_ode} by discretizing the time interval $[0,1]$ into $300$ time steps (see Appendix~\ref{app:num_exp}). The sample quality of local-patch RGFM deteriorates relative to the $64\times64$ case, but the generated images remain substantially more coherent than those obtained with local-patch FM. The remaining failures of RGFM primarily appear as inconsistencies between facial components that are spatially separated, such as differences in the shape, orientation, or appearance of the two eyes. For high-resolution facial images, global attributes---including pose, illumination, overall facial geometry, and identity-related information---play an increasingly important role in determining the local image structure. Inferring such latent information reliably from a fixed-size local patch becomes more difficult as the image resolution increases. We therefore expect that the degradation observed at $256\times256$ arises, at least in part, because the local observations do not contain sufficient information to determine these global latent attributes consistently across distant regions.

\section{\label{sec:discussions}Discussion}
In this paper, we introduce RGFM as a framework for scalable local generative modeling based on the scale separation and quasi-locality of the RG. Under physically motivated assumptions on the data distribution and its RG flow, we show that the characteristic length scale controlling the local approximability of the RGFM velocity field grows proportionally to the running RG length scale $\Lambda(t)^{-1}$. Based on this result, we introduce successive site-decimation transformations that keep this locality length at $O(L^0)$ in the rescaled lattice units throughout the flow. The resulting probability path can therefore be tracked using local velocity fields with patches whose linear size grows only as $O(\ln L)$ for a prescribed target accuracy. We demonstrate this construction numerically for one-dimensional local and conditionally local distributions and for natural image generation. In these examples, local RGFM reproduces long-range correlations and global structures using receptive fields substantially smaller than the full system size, while conventional local FM exhibits pronounced errors in long-range structure. Since the resulting model evaluates the velocity field patchwise, its computational cost can scale near-linearly with the data dimension, up to polylogarithmic factors. These results suggest that controlling the locality of the probability path itself provides a route toward scalable generative modeling of high-dimensional spatially structured data.

In the image-generation experiments, RGFM retains a clear practical advantage even when the buffer width becomes comparable to the linear system size $L$, where local FM is no longer strongly limited by its receptive field. This suggests that the benefit of RGFM is not solely due to locality, but also to its scale-by-scale decomposition, which maps long-range structures in the original data space onto smaller effective systems through site decimation. This observation is consistent with the success of multiscale generative approaches, in which coarse structures are generated first and subsequently refined \cite{kadkhodaie2023,guth2022a,marchand2023,gerdes2024}. In particular, wavelet-based approaches \cite{guth2022a,marchand2023} have similarly connected scale-by-scale generative modeling to RG ideas, exploiting the fact that high-frequency fluctuations can remain short-range correlated even when the coarse field develops long-range correlations. In the present work, we directly construct the flow-matching probability path from an exact RG flow and analyze the locality of the resulting velocity field. In particular, under the assumptions considered here, we show that its locality length scales as $\Lambda^{-1}$, providing a theoretical basis for scale-by-scale local generative modeling. Moreover, each intermediate distribution along the direct RGFM path has a natural interpretation as an effective theory at the running RG scale $\Lambda$. In this way, RGFM provides a systematic connection between coarse-to-fine generation and scale-dependent local effective descriptions in the RG.

At the same time, the present results also indicate a limitation of purely local modeling for high-resolution real-world data. In the $256\times256$ image-generation experiment, local RGFM produces substantially more coherent samples than local FM, but inconsistencies remain between spatially separated facial components. Within the framework of conditionally local distributions developed in this work, a natural interpretation is that a local patch no longer contains sufficient information to reliably infer the latent variables $z$ controlling global attributes such as identity, pose, illumination, or overall facial geometry. We therefore speculate that the degradation at higher resolutions reflects a breakdown, or at least a weakening, of the local-predictability assumption for the relevant latent variables, namely, Assumptions~\ref{assump:latent_cmi} and \ref{assump:rg_loc_cmi}, rather than a failure of the local description of the remaining fluctuations.

This limitation may be addressed by several extensions of the present framework. One possibility is to combine local RGFM with a separate model that infers a latent variable $z(\phi)$ and conditions the local velocity field on this information as $v_t(\phi,z)$. Once the relevant global information is supplied, the remaining generative dynamics may again admit an accurate local description, as suggested by our analysis of conditionally local distributions. If the latent-inference model can be kept sufficiently lightweight, such a hybrid approach may preserve much of the computational scalability of local-patch modeling while improving global consistency. A related direction is to perform local RGFM in a latent representation obtained from a pretrained encoder, as is widely done in high-resolution image or video generation \cite{rombach2022a,blattmann2023,ma2025,ma2025a}. More generally, it would be interesting to combine the probability flow of RGFM with architectures that model global information in coarse or latent representations \cite{ho2022c,peebles2023,gerdes2024}. Combining such global modeling with the locality-preserving flow of RGFM may provide a way to capture long-range structure while retaining the advantages of local generative modeling.

A further natural application of the present framework is the sampling or learning of physical many-body systems \cite{wang2024,singha2025,tan2026,hu2026a}. Many equilibrium statistical-mechanical models are defined by local Hamiltonians or actions and therefore fall naturally within the class of distributions considered in our theoretical analysis. A point emphasized by our analysis is that the locality relevant to local generative modeling is not the decay length of correlation functions themselves but the locality of the RG-evolved interaction $V_\Lambda$. Consequently, a distribution can possess correlations extending over very long distances while its RGFM velocity field remains locally approximable at each RG scale. This distinction is particularly relevant near criticality, where the physical correlation length becomes large even though the microscopic theory remains local. It therefore suggests that local RGFM may provide an efficient generative-sampling approach in regimes where conventional local sampling algorithms suffer from critical slowing down \cite{hohenberg1977,wolff1990}.

More broadly, it is interesting to ask how the present framework can be extended beyond systems governed by strictly or quasi-locally interacting effective theories. Long-range interacting statistical models \cite{dauxois2002,campa2009,defenu2023}, as well as nonequilibrium systems with intrinsically nonlocal collective effects \cite{toner1995,toner1998,bertini2015}, provide examples in which the exponential locality assumptions used in the present analysis may no longer hold. Nevertheless, in such cases, the nonlocal part of the dynamics may admit a lower-dimensional representation in terms of collective or latent variables. Combining a model that identifies or infers these nonlocal degrees of freedom with an RGFM-type local generative model for the remaining fluctuations could provide a route toward scalable modeling even when strict locality is absent. More generally, combining local computation with global representations learned by neural networks may provide a flexible framework to generate multiscale structures in physical systems. Extending the present approach along these directions may enable numerical studies of system sizes and parameter regimes that remain difficult to access with fully global generative models.

\begin{acknowledgments}
    We thank Yoshiyuki Kabashima, Kyogo Kawaguchi, Yang-yang Tan, and Lingxiao Wang for fruitful discussions.
    K.M. acknowledges support from JSPS KAKENHI Grant No.~24KJ0898. Y.A. acknowledges support from JST FOREST Program (Grant No. JPMJFR222U), JST CREST (Grant No. JPMJCR23I2), and JST Moonshot Research and Development Program (Grant No. JPMJMS256J).
\end{acknowledgments}

\section*{CODE AVAILABILITY}
Code, all the hyperparameters, and the trained models used in the numerical experiments can be found at \url{https://github.com/kantamasuki/Local_RGFM}.

\appendix

\section{\label{app:math}Mathematical lemmas used in the main text}
Here, we provide several mathematical lemmas and their proofs, which play an important role in the discussion in the main text.

\begin{lemma}\label{lem:gronwall}
    Gr\"onwall inequality \cite{gronwall1919}.

    Let a time-dependent function $a(t)$ satisfy $a(0)=0$ and
    \begin{align}
        \frac{da(t)}{dt} \leq f(t) + M(t)a(t).\label{eq:gron_asump}
    \end{align}
    Then, the following inequality holds.
    \begin{align}
        a(t) \leq \int_0^t ds\,f(s)\,e^{\int_s^t dr\, M(r)}.
    \end{align}
\end{lemma}

\begin{proof}
We define the integrating factor $N(t)$ for the differential equation~\eqref{eq:gron_asump} as $N(t)=\exp(-\int_0^t dr M(r))$, which satisfies $N'= -MN$. Therefore, we have
\begin{align}
    (Na)' &= - MN a + Na'\\
            &\leq - MN a + N(f + Ma)\\
            &= Nf.
\end{align}
Integrating both sides of this inequality and using $a(0)\!=\!0$, we obtain
\begin{align}
    N(t) a(t) &\leq \int_0^t ds\, N(s) f(s),\\
    \Leftrightarrow a(t) &\leq \int_0^t ds\, f(s)\,e^{\int_s^t dr\, M(r)},
\end{align}
which is the desired inequality.
\end{proof}

\begin{lemma}\label{lem:minkowski}
    Minkowski integral inequality \cite{folland1999}.
    
    Let $f(x)$ and $g(y)$ be nonnegative functions on $X$ and $Y$, respectively, such that $\int_X dx\,f(x)$ and $\int_Y dy\,g(y)$ are finite. Then, for any function $F:X\!\times\!Y\!\to\!\mathbb{R}_{\geq 0}$ and any $1\!\leq\! r\!<\!\infty$, the following inequality holds
    \begin{align}
        &\left(\int_X dx\, f(x) \left(\int_Y dy\, g(y) F(x,y)\right)^{r} \right)^\frac{1}{r} \nonumber\\
        & \quad \leq \int_Y dy\, g(y) \left(\int_X dx\, f(x) F(x,y)^r\right)^{\frac{1}{r}}.
    \end{align}
\end{lemma}

\begin{proof}
For $r\!=\!1$, the two sides differ only in the order of integration, so the statement follows immediately. We therefore consider the case $1\!<\!r\!<\!\infty$. Defining $F_x=\int_Y dy\,g(y)F(x,y)$, the $r$-th power of the left-hand side can be written as
\begin{align}
    \int_X dx\, f(x) F_x^r &= \int_X dx\, f(x) F_x^{r-1} \int_Y dy\, g(y) F(x,y)\\
    &= \int_Y dy\, g(y) \left(\int_X dx\, f(x) F_x^{r-1} F(x,y)\right),\label{eq:lemma_minkowski1}
\end{align}
where we exchange the order of integration in the second equality.

We now introduce the conjugate exponent $r'=r/(r-1)$, which satisfies $1/r+1/r'=1$. Applying H\"older's inequality to the integral over $X$, we obtain
\begin{align}
    &\int_X dx\, f(x) F_x^{r-1} F(x,y) \nonumber\\
    &\leq 
        \left(\int_X dx\, f(x) (F_x^{r-1})^{r'} \right)^\frac{1}{r'}
        \!\left(\int_X dx\, f(x) F(x,y)^r\right)^\frac{1}{r}\nonumber\\
    &= \left(\int_X dx\, f(x) F_x^r \right)^\frac{r-1}{r}
    \left(\int_X dx\, f(x) F(x,y)^r\right)^\frac{1}{r}.
\end{align}
Substituting this bound into Eq.~\eqref{eq:lemma_minkowski1} gives
\begin{align}
    \left(\int_X dx\, f(x) F_x^r\right)^\frac{1}{r}
    \!\leq\! \int_Y dy\, g(y) \left(\int_X dx\, f(x) F(x,y)^r\right)^\frac{1}{r}, 
\end{align}
which is precisely the desired Minkowski inequality.
\end{proof}

\onecolumngrid

\section{\label{app:proofs}Proofs in the main text}

\subsection{\label{proof:prop:phix_4_bound}Proof of Proposition~\ref{prop:phix_4_bound} in the main text}

From the triangle inequality, we have $\left\|\phi_x-\langle \phi_x\rangle_{p_\Lambda}\right\|_{L_4(p_\Lambda)}\leq \left\|\phi_x\right\|_{L_4(p_\Lambda)}$ + $\left\|\langle \phi_x\rangle_{p_\Lambda}\right\|_{L_4(p_\Lambda)}$. Further, from conditional Jensen's inequality, we have $\left\|\langle \phi_x\rangle_{p_\Lambda}\right\|_{L_4(p_\Lambda)} \leq \left\|\phi_x\right\|_{L_4(p_\Lambda)}$. Therefore, we obtain the inequality
\begin{align}
    \left\|\phi_x - \langle \phi_x\rangle_{p_\Lambda}\right\|_{L_4(p_\Lambda)} \leq 2 \left\|\phi_x\right\|_{L_4(p_\Lambda)}.\label{eq:phix_4_bound1}
\end{align}

By using the RG diffusion~\eqref{eq:rg_diffusion}, $\phi\sim p_\Lambda$ can be expressed as $\phi = \sqrt{K_\Lambda}\phi_0 + \sqrt{G_0(1-K_\Lambda)}\epsilon$ with $\phi_0\sim p_{\rm data}$ and $\epsilon\sim \mathcal{N}(0,I)$. Therefore, we can express $\phi_x$ as 
\begin{align}
    \phi_x &= \int_{\Omega_L} d^dy R_\Lambda(x-y) \phi_{0y} + \sum_k \frac{e^{ikx}}{\sqrt{L^d}} \sqrt{\frac{1-e^{-\frac{k^2+m^2}{\Lambda^2}}}{k^2+m^2}} \epsilon_k,
\end{align}
where $R_\Lambda(x) = (\Lambda^d/(2\pi)^\frac{d}{2})e^{-\frac{m^2}{2\Lambda^2}-\frac{\Lambda^2x^2}{2}}$ is the Fourier transform of $\sqrt{K_\Lambda(k)}$. Thus, one can bound $\left\|\phi_x\right\|_{L_4(p_\Lambda)}$ as follows:
\begin{align}
    \left\|\phi_x\right\|_{L_4(p_\Lambda)} &\leq \left\|
        \int_{\Omega_L} d^dy R_\Lambda(x-y) \phi_{0y}
    \right\|_{L_4(p_{\rm data})}
    + 
    \left\|
        \sum_k \frac{e^{ikx}}{\sqrt{L^d}} \sqrt{\frac{1-e^{-\frac{k^2+m^2}{\Lambda^2}}}{k^2+m^2}} \epsilon_k
    \right\|_{L_4(\mathcal N(0,I))}\\
    &\leq \int_{\Omega_L} d^dy R_\Lambda(x-y) \left\|\phi_{0y}\right\|_{L_4(p_{\rm data})} + \left(3 \left( \frac{1}{L^d} \sum_k \frac{1-e^{-\frac{k^2+m^2}{\Lambda^2}}}{k^2+m^2} \right)^2\right)^{\frac{1}{4}}\\
    &\leq \int_{\Omega_L} d^dy R_\Lambda(x-y) \left\|\phi_{0y}\right\|_{L_4(p_{\rm data})} + 3^{\frac{1}{4}} \left(\int_{|k|\leq \pi} \frac{d^dk}{(2\pi)^d} \frac{1}{k^2+m^2} \right)^{\frac{1}{2}}.
\end{align}
Since $\int_{\Omega_L} d^dy R_\Lambda(x-y)=e^{-m^2/2\Lambda^2}< 1$ and $\left\|\phi_{0y}\right\|_{L_4(p_{\rm data})}$ is uniformly bounded by a constant due to Assumption~\ref{assump:data_dist} for local data distributions, the right-hand side can be bounded by a constant independent of $x$ and $\Lambda$. Together with the inequality~\eqref{eq:phix_4_bound1}, we prove that $\left\|\phi_x - \langle \phi_x\rangle_{p_\Lambda}\right\|_{L_4(p_\Lambda)}$ is uniformly bounded.

\subsection{\label{proof:prop:bound1_loc_data_dist}Proof of Proposition~\ref{prop:bound1_loc_data_dist} in the main text}

In this proof, we denote the functional derivative $\delta V_\Lambda/\delta \phi$ by $u_\Lambda(\phi)$.  
Since the velocity field $v_\Lambda(\phi)$ is proportional to $u_\Lambda(\phi)$ with field-independent coefficient $1/\Lambda^2\tau$, as shown in Proposition~\ref{prop:v_derV}, it is sufficient to prove the following inequality:
\begin{align}
    \Delta_A(p_\Lambda, u_\Lambda; l_B) &\leq \gamma\, {\rm Poly}(c\Lambda l_B) e^{-c\Lambda l_B}\nonumber\\
    &\qquad \!\times\!\frac{1}{\Lambda^d} \left(\int_A d^dx\, \mathbb E_{\phi\sim p_\Lambda} \left[\left|u_{\Lambda,x}(\phi)\right|^2\right]\right)^{\frac{1}{2}}\!.
\end{align}

Given a field configuration $\phi_{AB}$ on region $A\cup B$, one can extend it to a field $\bar{\phi}_{AB}$ on $\Omega_L$ by setting $\phi_C=\langle \phi_C\rangle_{p_\Lambda}$. We then define a local field ${u'}_\Lambda^{A}$ on region $A$ by ${u'}_\Lambda^{A}(\phi_{AB})=u_\Lambda^A(\bar{\phi}_{AB})$, where $u_\Lambda^A$ is the restriction of $u_\Lambda$ on region $A$. Since ${u'}_\Lambda^A$ is a local field on $A$, Proposition~\ref{prop:la_minimizer} leads to the inequality 
\begin{align}
    \Delta^2_A\left(p_\Lambda, u_\Lambda; l_B\right) 
    &\leq \mathbb E_{\phi\sim p_\Lambda(\phi)} [|| u_\Lambda^A(\phi) - {u'}_\Lambda^A(\phi_{AB})||_A^2]\\
    &= \int_{A} dx\, \mathbb{E}_{\phi\sim p_\Lambda(\phi)} \left[||u_{\Lambda,x}^A(\phi) - u_{\Lambda,x}^A(\bar\phi_{AB})||^2\right].\label{eq:ql_sl_1}
\end{align}

For each $x\in A$, the difference between the original and truncated forces, $u_{\Lambda,x}^A(\phi)$ and $u_{\Lambda,x}^A(\bar\phi_{AB})$, can be expressed using the truncation interpolation $T_{ABC}^\lambda$ (see Def.~\ref{def:trun_interpolation}) as
\begin{align}
        u_{\Lambda,x}^A(\phi) - u_{\Lambda,x}^A(\bar\phi_{AB}) 
        &= u_{\Lambda,x}^A(T_{ABC}^{\lambda=1}\phi) - u_{\Lambda,x}^A(T_{ABC}^{\lambda=0}\phi)\\
        &= \int_0^1 d\lambda\, \frac{du_{\Lambda,x}^A(T_{ABC}^\lambda\phi)}{d\lambda}\\
        &= \int_0^1 d\lambda\, \int_C dy\, \left(\phi_y-\langle\phi_y\rangle_{p_\Lambda}\right) \frac{\delta u_{\Lambda,x}^A}{\delta \psi_y}\biggr|_{\psi=T_{ABC}^\lambda\phi}\\
        &= \int_0^1 d\lambda\, \int_C dy\, \left(\phi_y-\langle\phi_y\rangle_{p_\Lambda}\right) \left[\frac{\delta^2 V_\Lambda}{\delta\psi_x\delta \psi_y}\right]_{T_{ABC}^\lambda\phi}.
\end{align}
Then, using the Minkowski integral inequality in Lemma~\ref{lem:minkowski}, we obtain 
\begin{align}
    \left(\mathbb E_{\phi\sim p_\Lambda(\phi)} \left[||u_{\Lambda,x}^A(\phi) - u_{\Lambda,x}^A(\bar\phi_{AB})||^2 \right] \right)^{\frac{1}{2}}
    &= \left(\int d\phi\, p_\Lambda(\phi) \left(
        \int_C dy\, \int_0^1 d\lambda\, \left(\phi_y-\langle\phi_y\rangle_{p_\Lambda}\right) \left[\frac{\delta^2 V_\Lambda}{\delta\psi_x\delta\psi_y}\right]_{T_{ABC}^\lambda\phi}
    \right)^2 \right)^{\frac{1}{2}}\\
    &\leq \int_C dy\, \int_0^1 d\lambda\, \left( \int d\phi\, p_\Lambda(\phi) \left(\left(\phi_y-\langle\phi_y\rangle_{p_\Lambda}\right)\left[\frac{\delta^2 V_\Lambda}{\delta\psi_x\delta\psi_y}\right]_{T_{ABC}^\lambda\phi}\right)^2 \right)^{\frac{1}{2}} \\
    &= \int_C dy\, \int_0^1 d\lambda\,\left\| \left(\phi_y-\langle\phi_y\rangle_{p_\Lambda}\right) \left[\frac{\delta^2 V_\Lambda}{\delta\psi_x\delta\psi_y}\right]_{T_{ABC}^\lambda\phi}\right\|_{L^2(p_\Lambda)}.
\end{align}
Applying H\"older's inequality in the form $\|fg\|_{L^2(p)}\leq\|f\|_{L_4(p)} \cdot \|g\|_{L_4(p)}$ and using the RG locality assumption~\ref{assump:rg_loc_int} and the uniform bound on $\left\|\phi_y-\langle\phi_y\rangle_{p_\Lambda}\right\|_{L_4(p_\Lambda)}$ in Proposition~\ref{prop:phix_4_bound}, we obtain
\begin{align}
    \mathbb E_{\phi\sim p_\Lambda(\phi)} \left[||u_{\Lambda,x}^A(\phi) - u_{\Lambda,x}^A(\bar\phi_{AB})||^2 \right]
    & \leq \left(\int_C dy\, \int_0^1 d\lambda\, 
    \left\|\phi_y-\langle\phi_y\rangle_{p_\Lambda}\right\|_{L_4(p_\Lambda)} \cdot 
    \left\|\left[\frac{\delta^2 V_\Lambda}{\delta\psi_x\delta\psi_y}\right]_{T_{ABC}^\lambda\phi}\right\|_{L_4(p_\Lambda)} \right)^2\\
    & \leq \left(M\gamma \left\|\frac{\delta V_\Lambda}{\delta\phi_x}\right\|_{L_2(p_\Lambda)} \int_C dy\, e^{-c\Lambda|x-y|} \right)^2\\
    & \leq \left(M\gamma \left\|u_{\Lambda,x}\right\|_{L_2(p_\Lambda)} \int_{l_B}^\infty ds\, e^{-c\Lambda s} \, 2d\cdot(2s)^{d-1} \right)^2\\
    & = \left(
        M\gamma \left\|u_{\Lambda,x}\right\|_{L_2(p_\Lambda)} \frac{d\cdot 2^d}{c^d\Lambda^d} e^{-c\Lambda l_B} \int_0^\infty ds\, e^{-s}\left(s + c\Lambda l_B \right)^{d-1}
    \right)^2\\
    & = \left(\frac{\gamma'\, {\rm Poly}\left(c\Lambda l_B\right)e^{-c\Lambda l_B}}{\Lambda^d}\right)^2 \left\|u_{\Lambda,x}\right\|_{L_2(p_\Lambda)}^2.
    \label{eq:ql_sl_2}
\end{align}
Here, we bound the integral with respect to $y\in C$ by integrating over shells centered at $x\in A$.
We also define the polynomial ${\rm Poly}(x)$ by ${\rm Poly}(x) = \int_0^\infty ds\,e^{-s}(s+x)^{d-1}$ and absorb constants independent of $L$ and $\Lambda$ into $\gamma'$.

By integrating Eq.~\eqref{eq:ql_sl_2} with respect to $x\in A$ and using Eq.~\eqref{eq:ql_sl_1}, we obtain 
\begin{align}
    \Delta_A(p_\Lambda, u_\Lambda; l_B) \leq \gamma'\, {\rm Poly}(c\Lambda l_B) e^{-c\Lambda l_B} \frac{1}{\Lambda^d} \left(\int_A d^dx\, \mathbb E_{\phi\sim p_\Lambda} \left[\left|u_{\Lambda,x}(\phi)\right|^2\right]\right)^{\frac{1}{2}},
\end{align}
which completes the proof.

\subsection{\label{proof:prop:LAE_A_condloc_1}Proof of Proposition~\ref{prop:LAE_A_condloc_1} in the main text}

First, we note that $\mathbb E_{z|\phi_{AB}}[v_{\Lambda,{\rm loc}}^A(\phi_{AB}|z)]$ is a local field on region $A$ since it only depends on $\phi_{AB}$. Therefore, by definition, the local approximation error $\Delta_A(p_\Lambda, v_\Lambda; l_B)$ satisfies
\begin{align}
    \Delta_A(p_\Lambda, v_\Lambda; l_B)^2 
    &\leq \mathbb E_{p_\Lambda(\phi)} \left[\left\| v^A_\Lambda(\phi) - \mathbb E_{z|\phi_{AB}}[v_{\Lambda, {\rm loc}}^A(\phi_{AB}|z)] \right\|_A^2\right]\\
    &\leq \mathbb E_{p_\Lambda(\phi)} \left[\left\| \mathbb E_{z|\phi} [v_{\Lambda}^A(\phi|z)] - \mathbb E_{z|\phi_{AB}}[v_{\Lambda, {\rm loc}}^A(\phi_{AB}|z)] \right\|_A^2\right],
\end{align}
where $v_\Lambda^A(\phi)$ is the restriction of $v_\Lambda(\phi)$ on region $A$, and we use the velocity equality~\eqref{eq:vLam_cond}.

Then, using the inequality $||a+b||^2_A\leq 2||a||^2_A + 2||b||^2_A$ and applying conditional Jensen's inequality, we obtain 
\begin{align}
    \Delta_A(p_\Lambda, v_\Lambda; l_B)^2 
    &\leq 2\mathbb E_{p_\Lambda(\phi)} \left[\left\|
        \mathbb E_{z|\phi} [v_{\Lambda}^A(\phi|z)] - \mathbb E_{z|\phi}[v_{\Lambda,{\rm loc}}^A(\phi_{AB}|z)]
    \right\|_A^2\right]\nonumber\\
    & \qquad + 2\mathbb E_{p_\Lambda(\phi)} \left[\left\| \mathbb E_{z|\phi}[v_{\Lambda,{\rm loc}}^A(\phi_{AB}|z)] - \mathbb E_{z|\phi_{AB}}[v_{\Lambda,{\rm loc}}^A(\phi_{AB}|z)] \right\|_A^2\right]\\
    &\leq 2 \mathbb E_{p_\Lambda(\phi,z)}\left[\left\|v_{\Lambda}^A(\phi|z) - v_{\Lambda,{\rm loc}}^A(\phi_{AB}|z)\right\|_A^2\right] \!+\! 2 \mathbb E_{p_\Lambda(\phi)} \left[\left\|
        \mathbb E_{z|\phi}[v_{\Lambda,{\rm loc}}^A(\phi_{AB}|z)]
        - \mathbb E_{z|\phi_{AB}}[v_{\Lambda,{\rm loc}}^A(\phi_{AB}|z)]
    \right\|_A^2\right]\\
    &= 2 \mathbb E_{p(z)}\left[
        \Delta_A(p_\Lambda(\cdot|z),v_\Lambda(\cdot|z);l_B)^2
    \right] + 2 \mathbb E_{p_\Lambda(\phi)} \left[\left\|
        \mathbb E_{z|\phi}[v_{\Lambda,{\rm loc}}^A(\phi_{AB}|z)]
        - \mathbb E_{z|\phi_{AB}}[v_{\Lambda,{\rm loc}}^A(\phi_{AB}|z)]
    \right\|_A^2\right],
\end{align}
which proves Eq.~\eqref{eq:LAE_condloc}.

\subsection{\label{proof:prop:bound_latent_pred}Proof of Proposition~\ref{prop:bound_latent_pred} in the main text}

First, we define $h_\Lambda(\phi,z)$ by
\begin{align}
    h_\Lambda(\phi,z) 
    &= v_{\Lambda,{\rm loc}}^A(\phi_{AB}|z) 
    - 
    \mathbb E_{z'|\phi_{AB}}[v_{\Lambda,{\rm loc}}^A(\phi_{AB}|z')].
\end{align}
We then take an arbitrary positive constant $R$ and divide the configuration space $\mathcal C=\left\{(\phi,z)\right\}$ into 
\begin{align}
    C^< := \left\{(\phi,z)\in \mathcal C\,\bigr|\, \left\|h_\Lambda(\phi,z)\right\|_A \leq R\right\},\\
    C^> := \left\{(\phi,z)\in \mathcal C\,\bigr|\, \left\|h_\Lambda(\phi,z)\right\|_A > R\right\}.
\end{align}
Based on this decomposition, we decompose $h_\Lambda$ as $h_\Lambda = h_\Lambda^> + h_\Lambda^<$ with 
\begin{align}
    h_\Lambda^<(\phi,z) &= \bm 1_{C^<}(\phi,z) \cdot h_\Lambda(\phi,z),\\
    h_\Lambda^>(\phi,z) &= \bm 1_{C^>}(\phi,z) \cdot h_\Lambda(\phi,z).
\end{align}
Then, since $\mathbb E_{z|\phi_{AB}}[h_\Lambda(\phi,z)] = 0$, the left-hand side of Eq.~\eqref{eq:bound_latent_pred} can be bounded as 
\begin{align}
    \mathbb E_{p_\Lambda(\phi)} \left[\left\|
        \mathbb E_{z|\phi}[v_{\Lambda,{\rm loc}}^A(\phi_{AB}|z)]
        - \mathbb E_{z|\phi_{AB}}[v_{\Lambda,{\rm loc}}^A(\phi_{AB}|z)]
    \right\|_A^2\right]
    &= \mathbb E_{p_\Lambda(\phi)}\left[\left\|
        \mathbb E_{z|\phi} \left[h_\Lambda(\phi,z)\right]
        - \mathbb E_{z|\phi_{AB}} \left[h_\Lambda(\phi,z)\right]
    \right\|_A^2\right]\\
    &= \mathbb E_{p_\Lambda(\phi)}\left[\left\|
        (\mathbb E_{z|\phi}- \mathbb E_{z|\phi_{AB}}) \left[h^<_\Lambda\right]
        + 
        (\mathbb E_{z|\phi}- \mathbb E_{z|\phi_{AB}}) \left[h^>_\Lambda\right]
    \right\|_A^2\right]\\
    &\leq 2 \mathbb E_{p_\Lambda(\phi)}\left[\left\|
        (\mathbb E_{z|\phi}- \mathbb E_{z|\phi_{AB}}) \left[h^<_\Lambda\right]
    \right\|_A^2\right]\nonumber\\
    &\qquad + 2\mathbb E_{p_\Lambda(\phi)}\left[\left\|
                (\mathbb E_{z|\phi}- \mathbb E_{z|\phi_{AB}}) \left[h^>_\Lambda\right]
    \right\|_A^2\right],
\end{align}
where we use the inequality $||a+b||_A^2 \leq 2||a||_A^2 + 2||b||_A^2$.

For $h_\Lambda^<$, we have 
\begin{align}
    \left\|(\mathbb E_{z|\phi}- \mathbb E_{z|\phi_{AB}})[h^<_\Lambda(\phi,z)]\right\|_A 
    &= \left\|
        \int dz\, \left(p_\Lambda(z|\phi) - p_\Lambda(z|\phi_{AB}) \right)h_\Lambda(\phi,z) \cdot \bm 1_{C^<}
    \right\|_A\\
    &\leq \int dz\, 
        \left|p_\Lambda(z|\phi) - p_\Lambda(z|\phi_{AB}) \right| \left\| h_\Lambda(\phi,z) \cdot \bm 1_{C^<}
    \right\|_A\\
    &\leq R \int dz\, \left|p_\Lambda(z|\phi) - p_\Lambda(z|\phi_{AB})\right|\\
    &\leq \sqrt{2} R \sqrt{D_{\rm KL}(p_\Lambda(z|\phi)||p_\Lambda(z|\phi_{AB}))},
\end{align}
where we use Pinsker's inequality to bound the total variation $\int dz\,|p_1(z)-p_2(z)|$ by the KL divergence as $\int dz\,|p_1(z)-p_2(z)| \leq \sqrt{2 D_{\rm KL}(p_1||p_2)}$. Averaging over $p_\Lambda(\phi)$, we obtain 
\begin{align}
    \mathbb E_{p_\Lambda(\phi)}\left[\left\|
        (\mathbb E_{z|\phi}- \mathbb E_{z|\phi_{AB}}) \left[h^<_\Lambda\right]
    \right\|_A^2\right]
    &\leq 2R^2\, \mathbb E_{p_\Lambda(\phi)} \left[
        D_{\rm KL}(p_\Lambda(z|\phi)||p_\Lambda(z|\phi_{AB}))
    \right]\\
    &= 2R^2 I_\Lambda(Z\!:\!C|AB).\label{eq:bound_h_lt}
\end{align}

Using conditional Jensen's inequality, for $q>2$, we have
\begin{align}
    \mathbb E_{p_\Lambda(\phi)} \left[\left\|\mathbb E_{z|\phi} \left[h_\Lambda^>\right]\right\|_A^2\right]
    & \leq \mathbb E_{p_\Lambda(\phi,z)} \left[\left\|
        h_\Lambda \cdot \bm 1_{C^>}
    \right\|_A^2 \right]\\
    & \leq R^{2-q}\, \mathbb E_{p_\Lambda(\phi,z)} \left[\left\|
        h_\Lambda
    \right\|_A^q\right].
\end{align}
Similarly, since $h_\Lambda$ depends only on $\phi_{AB}$ and $z$, conditional Jensen's inequality leads to 
\begin{align}
    \mathbb E_{p_\Lambda(\phi)} \left[\left\|\mathbb E_{z|\phi_{AB}} \left[h_\Lambda^>\right]\right\|_A^2\right]
    & \leq \mathbb E_{p_\Lambda(\phi,z)} \left[\left\|
        h_\Lambda \cdot \bm 1_{C^>}
    \right\|_A^2 \right]\\
    & \leq R^{2-q}\, \mathbb E_{p_\Lambda(\phi,z)} \left[\left\|
        h_\Lambda
    \right\|_A^q\right].
\end{align}
Therefore, using the inequality $||a+b||_A^2 \leq 2||a||_A^2 + 2||b||_A^2$,
\begin{align}
    \mathbb E_{p_\Lambda(\phi)}\left[\left\|
                (\mathbb E_{z|\phi}- \mathbb E_{z|\phi_{AB}}) \left[h^>_\Lambda\right]
    \right\|_A^2\right] 
    &\leq 2 \mathbb E_{p_\Lambda(\phi)} \left[\left\|\mathbb E_{z|\phi} \left[h_\Lambda^>\right]\right\|_A^2\right] + 2 \mathbb E_{p_\Lambda(\phi)} \left[\left\|\mathbb E_{z|\phi_{AB}} \left[h_\Lambda^>\right]\right\|_A^2\right]\\
    &\leq 4 R^{2-q} \mathbb E_{p_\Lambda(\phi,z)} \left[\left\|
        h_\Lambda
    \right\|_A^q\right].\label{eq:bound_h_gt}
\end{align}

Combining Eqs.~\eqref{eq:bound_h_lt} and \eqref{eq:bound_h_gt}, we arrive at the inequality 
\begin{align}
    \mathbb E_{p_\Lambda(\phi)} \left[\left\|
        \mathbb E_{z|\phi}[v_{\Lambda,{\rm loc}}^A(\phi_{AB}|z)]
        - \mathbb E_{z|\phi_{AB}}[v_{\Lambda,{\rm loc}}^A(\phi_{AB}|z)]
    \right\|_A^2\right] \leq 4 R^2 I_\Lambda(Z\!:\!C|AB) + 8 R^{2-q} \mathbb E_{p_\Lambda(\phi,z)} \left[\left\|h_\Lambda\right\|_A^q\right].
\end{align} 
By taking $R$ as 
\begin{align}
    R=(q-2)^{\frac{1}{q}}I_\Lambda(Z\!:\!C|AB)^{-\frac{1}{q}}\left(\mathbb E_{p_\Lambda(\phi,z)} \left[\left\|h_\Lambda\right\|_A^q\right]\right)^\frac{1}{q},
\end{align}
we obtain the bound 
\begin{align}
    \mathbb E_{p_\Lambda(\phi)} \left[\left\|
        \mathbb E_{z|\phi}[v_{\Lambda,{\rm loc}}^A(\phi_{AB}|z)]
        - \mathbb E_{z|\phi_{AB}}[v_{\Lambda,{\rm loc}}^A(\phi_{AB}|z)]
    \right\|_A^2\right] 
    \leq \left[ 4q(q-2)^{\frac{2}{q}-1} \right] I_\Lambda(Z\!:\!C|AB)^{1-\frac{2}{q}} \left(\mathbb E_{\phi,z} \left[\left\|h_\Lambda\right\|_A^q\right]\right)^\frac{2}{q}.\label{eq:bound_IEhq1}
\end{align}

Finally, we bound $\left(\mathbb E_{\phi,z} \left[\left\|h_\Lambda\right\|_A^q\right]\right)^\frac{2}{q}$ in the right-hand side of Eq.~\eqref{eq:bound_IEhq1}. Using Minkowski's integral inequality, we obtain 
\begin{align}
    \left(\mathbb E_{\phi,z} \left[\left\| h_\Lambda\right\|_A^{q}\right]\right)^{\frac{2}{q}}
    &= \left(\int d\phi\, dz\, p_\Lambda(\phi,z) \left(\int_A d^dx\, |h_{\Lambda,x}(\phi,z)|^2\right)^{\frac{q}{2}}\right)^{\frac{2}{q}}\\
    &\leq \int_A d^dx\, \left(\int d\phi\,dz\, p_\Lambda(\phi,z) |h_{\Lambda,x}(\phi,z)|^{2\cdot\frac{q}{2}} \right)^{\frac{2}{q}}\\
    &= \int_A d^dx\, \left(\mathbb E_{\phi,z} \left[ |v_{\Lambda,{\rm loc},x}(\phi_{AB}|z) - \mathbb E_{z|\phi_{AB}}[v_{\Lambda,{\rm loc},x}(\phi_{AB}|z)]|^q \right]\right)^{\frac{2}{q}}\\
    &\leq \int_A d^dx\, \left(2^q 
        \mathbb E_{\phi,z} \left[ |v_{\Lambda,{\rm loc},x}(\phi_{AB}|z)|^q \right]
        \right)^{\frac{2}{q}}\\
    &\leq 4 \int_A d^dx\, \left(\mathbb E_{\phi,z} \left[ \left| v_{\Lambda,x}(\phi|z)\right|^q \right] \right)^{\frac{2}{q}}.\label{eq:bound_IEhq2}
\end{align}
In the last line of inequalities, we use the fact $\mathbb E_{\phi|z}\left[\left|v_{\Lambda,{\rm loc},x}(\phi_{AB}|z)\right|^q\right]\leq \mathbb E_{\phi|z}\left[\left|v_{\Lambda,x}(\phi|z)\right|^q\right]$. To see this, we recall that, by definition, the local approximation satisfies
\begin{align}
    v_{\Lambda,{\rm loc},x}(\phi_{AB}|z) &= \mathbb E_{p(\phi_C|\phi_{AB},z)}[v_{\Lambda,x}(\phi|z)].
\end{align}
Therefore, by conditional Jensen's inequality, 
\begin{align}
    \mathbb E_{\phi|z}\left[\left|
        v_{\Lambda,{\rm loc},x}(\phi_{AB}|z)
        \right|^q\right]
    &= \mathbb E_{\phi_{AB}|z}\left[\left| 
        v_{\Lambda,{\rm loc},x}(\phi_{AB}|z)
        \right|^q\right]\\
    &= \mathbb E_{\phi_{AB}|z}\left[\left|
        \mathbb E_{p(\phi_C|\phi_{AB},z)}[v_{\Lambda,x}(\phi|z)]
        \right|^q\right]\\
    &\leq \mathbb E_{p(\phi_{AB}|z), p(\phi_C|\phi_{AB},z)} \left[\left|
        v_{\Lambda,x}(\phi|z)
        \right|^q\right]\\
    &= \mathbb E_{\phi|z} \left[\left|
        v_{\Lambda,x}(\phi|z)
        \right|^q\right].
\end{align}

With Eqs.~\eqref{eq:bound_IEhq1} and \eqref{eq:bound_IEhq2} and using the RG assumption on the CMI, $I_\Lambda(Z\!:\!C|AB)\leq \gamma e^{-c\Lambda l_B}$, we obtain the bound 
\begin{align}
    \mathbb E_{\phi} \left[\left\|
        \mathbb E_{z|\phi}[v_{\Lambda,{\rm loc}}^A(\phi_{AB}|z)]
        - \mathbb E_{z|\phi_{AB}}[v_{\Lambda,{\rm loc}}^A(\phi_{AB}|z)]
    \right\|_A^2\right] \leq \gamma'^2 e^{-2c'\Lambda l_B} \int_A d^dx\, \left(\mathbb E_{\phi,z} \left[ \left| v_{\Lambda,x}(\phi|z)\right|^q \right] \right)^{\frac{2}{q}}.
\end{align}
Here, we absorb constants independent of $L$ and $\Lambda$ to $\gamma'$, and introduce $c'$ by $c'=((q-2)/2q)c$. This proves the inequality~\eqref{eq:bound_latent_pred}.

\subsection{\label{proof:thm:LA_cloc_data}Proof of Theorem~\ref{thm:LA_cloc_data} in the main text}

From Corollary~\ref{cor:LAE_cond_loc_cor}, it is sufficient to prove that the integral $\int_{\Omega_L}\! d^dx \left(\mathbb E_{\phi,z} \left[ \left| v_{\Lambda,x}(\phi|z)\right|^q \right] \right)^{\frac{2}{q}}$ is $O(L^{4d/3}(\ln L)^2)$. We note that, for a translation invariant system, the integral is $O(L^d(\ln L)^2)$, where the $O((\ln L)^2)$ contribution arises from the fact that the velocity field $v_{\Lambda,x}(\phi|z)$ is proportional to the inverse of the time scale $\tau^{-1} \propto \ln L$ as shown in the proof of Proposition~\ref{prop:bound3_loc_data_dist}. In general, we can bound the integral as follows. 

On a finite lattice, the integral can be bounded as 
\begin{align}
    \int_{\Omega_L} d^dx\, \left(\mathbb E_{\phi,z} \left[ \left| v_{\Lambda,x}(\phi|z)\right|^q \right] \right)^{\frac{2}{q}}
    & = \int_{\Omega_L} d^dx\, \left[\left(\mathbb E_{\phi,z} \left[ \left| v_{\Lambda,x}(\phi|z)\right|^q \right] \right)^{\frac{1}{q}}\right]^{2}\\
    &\leq (L^d)^{1-\frac{2}{q}} \left(\int_{\Omega_L} d^dx\, \mathbb E_{\phi,z} \left[ \left| v_{\Lambda,x}(\phi|z)\right|^q \right] \right)^{\frac{2}{q}}\\
    &= (L^d)^{1-\frac{2}{q}} \left(\mathbb E_{\phi,z} \left[\int_{\Omega_L} d^dx\,  \left| v_{\Lambda,x}(\phi|z)\right|^q \right] \right)^{\frac{2}{q}}\\
    &\leq (L^d)^{1-\frac{2}{q}} \left(\mathbb E_{\phi,z} \left[\left(\int_{\Omega_L} d^dx\,  \left| v_{\Lambda,x}(\phi|z)\right|^2 \right)^{\frac{q}{2}}\right] \right)^{\frac{2}{q}}\\
    &= (L^d)^{1-\frac{2}{q}} \left(\mathbb E_{\phi,z} \left[\left( L^d\int_{|k|\leq \pi} \frac{d^dk}{(2\pi)^d}\,  \left| v_{\Lambda,k}(\phi|z)\right|^2 \right)^{\frac{q}{2}}\right] \right)^{\frac{2}{q}}.\label{eq:vq_bound1}
\end{align}
In the first inequality, we use the inequality 
\begin{align}
    \int_{\Omega_L} d^dx\, |a_x|^2 \leq (L^d)^{1-2/q} \left(\int_{\Omega_L} d^dx\, |a_x|^q \right)^{\frac{2}{q}} 
\end{align}
with $a_x = \left(\mathbb E_{\phi,z}\left[\left|v_{\Lambda,x}(\phi|z)\right|^q\right]\right)^\frac{1}{q}$. We note that this inequality follows by applying H\"older's inequality to $|a_x|^2=|a_x|^2\cdot 1$ as 
\begin{align}
    \int_{\Omega_L} d^dx\, |a_x|^2 \leq \left(\int_{\Omega_L} d^dx\, |a_x|^{2\cdot\frac{q}{2}}\right)^{\frac{2}{q}} \left(\int_{\Omega_L} d^dx\, 1^{\frac{q}{q-2}}\right)^{\frac{q-2}{q}}. 
\end{align}
In the second inequality, we use the norm inequality on the finite lattice $\left(\int_{\Omega_L} d^dx\, |a_x|^q\right)^{\frac{1}{q}} \leq \left(\int_{\Omega_L} d^dx\, |a_x|^2\right)^{\frac{1}{2}}$ with $a_x=v_{\Lambda,x}(\phi|z)$. 

To evaluate Eq.~\eqref{eq:vq_bound1}, we recall Lemma~\ref{lem:vk_phi0k}. For the conditionally local distribution, $v_{\Lambda,k}(\phi|z)$ is expressed as 
\begin{align}
    v_{\Lambda,k} (\phi|z) = \frac{\partial_t K_{tk}}{2(1-K_{tk})}\left\langle-\phi_k + \frac{1}{\sqrt{K_{tk}}} \phi_{0k}\right\rangle_{\phi_0\sim p_t(\phi_0|\phi,z)}.
\end{align}
Then, by using the conditional Jensen's inequality twice, we obtain 
\begin{align}
    \mathbb E_{\phi,z} \left[\left(L^d\int_{|k|\leq \pi} \frac{d^dk}{(2\pi)^d}\,  \left| v_{\Lambda,k}(\phi|z)\right|^2 \right)^{\frac{q}{2}}\right]
    &= \mathbb E_{\phi,z} \left[\left(L^d\int_{|k|\leq \pi} \frac{d^dk}{(2\pi)^d}\,  \left| \frac{\partial_t K_{tk}}{2(1-K_{tk})}\left\langle-\phi_k + \frac{1}{\sqrt{K_{tk}}} \phi_{0k}\right\rangle_{\phi_0|\phi,z} \right|^2 \right)^{\frac{q}{2}}\right]\\
    &\leq \mathbb E_{\phi,z} \left[\left(L^d\int_{|k|\leq \pi} \frac{d^dk}{(2\pi)^d}\,  \mathbb E_{\phi_0|\phi,z} \left[\left| \frac{\partial_t K_{tk}}{2(1-K_{tk})}\left(-\phi_k + \frac{1}{\sqrt{K_{tk}}} \phi_{0k}\right) \right|^2\right] \right)^{\frac{q}{2}}\right]\\
    &= \mathbb E_{\phi,z} \left[\left(\mathbb E_{\phi_0|\phi,z}\left[L^d\int_{|k|\leq \pi} \frac{d^dk}{(2\pi)^d}\,  \left| \frac{\partial_t K_{tk}}{2(1-K_{tk})}\left(-\phi_k + \frac{1}{\sqrt{K_{tk}}} \phi_{0k}\right) \right|^2 \right] \right)^{\frac{q}{2}}\right]\\
    &\leq \mathbb E_{\phi_0,\phi,z} \left[\left(L^d\int_{|k|\leq \pi} \frac{d^dk}{(2\pi)^d}\,  \left| \frac{\partial_t K_{tk}}{2(1-K_{tk})}\left(-\phi_k + \frac{1}{\sqrt{K_{tk}}} \phi_{0k}\right) \right|^2 \right)^{\frac{q}{2}}\right].
\end{align}
By changing the variables as $\phi_{k}=\sqrt{K_{tk}}\phi_{0k} + \sqrt{1-K_{tk}}\epsilon_{k}/\sqrt{k^2+m^2}$ with  $\epsilon_k\sim \mathcal{N}(0,I)$, we have 
\begin{align}
    &\mathbb E_{\phi,z} \left[\left(L^d\int_{|k|\leq \pi} \frac{d^dk}{(2\pi)^d}\,  \left| v_{\Lambda,k}(\phi|z)\right|^2 \right)^{\frac{q}{2}}\right]\nonumber\\
    &\qquad \leq \mathbb E_{\phi_0,\phi,z} \left[\left(L^d\int_{|k|\leq \pi} \frac{d^dk}{(2\pi)^d}\,  \left| \frac{\partial_t K_{tk}}{2(1-K_{tk})}\left(\frac{1-K_{tk}}{\sqrt{K_{tk}}}\phi_{0k} - \frac{\sqrt{1-K_{tk}}}{\sqrt{k^2+m^2}}\epsilon_k\right) \right|^2 \right)^{\frac{q}{2}}\right]\\
    &\qquad = \mathbb E_{\phi_0,\phi,z} \left[\left(\frac{L^d}{\tau^2}\int_{|k|\leq \pi} \frac{d^dk}{(2\pi)^d}\,  \left| f_1\left(\frac{k^2+m^2}{\Lambda^2}\right)\phi_{0k} - f_2\left(\frac{k^2+m^2}{\Lambda^2}\right) \frac{\epsilon_k}{\sqrt{k^2+m^2}}\right|^2 \right)^{\frac{q}{2}}\right]\\
    &\qquad \leq \frac{1}{\tau^q}\mathbb E_{\phi_0,\phi,z} \left[ \left(\left\|f_1\phi_0\right\|_{\Omega_L} + \left\|\frac{f_2}{\sqrt{k^2+m^2}}\epsilon\right\|_{\Omega_L} \right)^q \right],
\end{align}
where we use $K_{tk}=e^{-(k^2+m^2)/\Lambda^2}$ and $\partial_t K_{tk}=-(2/\tau)K_{tk}(k^2+m^2)/\Lambda^2$. We also define $f_1(x)$ and $f_2(x)$ by $f_1(x)=xe^{-x/2}$ and $f_2(x)=xe^{-x}/\sqrt{1-e^{-x}}$, respectively. In the last inequality, we use the triangle inequality 
\begin{align}
    \left(L^d\int_{|k|\leq \pi} \frac{d^dk}{(2\pi)^d}\, |g_k-h_k|^2\right)^{\frac{1}{2}}
    &\leq \left(L^d\int_{|k|\leq \pi} \frac{d^dk}{(2\pi)^d}\, |g_k|^2\right)^{\frac{1}{2}}
    + \left(L^d\int_{|k|\leq \pi} \frac{d^dk}{(2\pi)^d}\, |h_k|^2\right)^{\frac{1}{2}}\\
    &= \left\|g_k\right\|_{\Omega_L} + \left\|h_k\right\|_{\Omega_L}.
\end{align}
Since $f_1(x)$ and $f_2(x)$ are bounded as $|f_1(x)|\leq a_*$ and $|f_2(x)|\leq b_*$, respectively, we arrive at
\begin{align}
    \mathbb E_{\phi,z} \left[\left(L^d\int_{|k|\leq \pi} \frac{d^dk}{(2\pi)^d}\,  \left| v_{\Lambda,k}(\phi|z)\right|^2 \right)^{\frac{q}{2}}\right] 
    &\leq \frac{1}{\tau^q} \mathbb E_{\phi_0,\phi,z} \left[ \left(a_*\left\|\phi_0\right\|_{\Omega_L} + \frac{b_*}{m} \left\|\epsilon\right\|_{\Omega_L}\right)^q \right].\label{eq:vq_bound2}
\end{align}

Combining Eq.~\eqref{eq:vq_bound2} with norm triangle inequality $\left(\mathbb E_{p}\left[|v+w|^q\right]\right)^{1/q} \leq \left(\mathbb E_{p}\left[|v|^q\right]\right)^{1/q} +  \left(\mathbb E_{p}\left[|w|^q\right]\right)^{1/q}$, we obtain the bound for Eq.~\eqref{eq:vq_bound1} as 
\begin{align}
    \int_{\Omega_L} d^dx\, \left(\mathbb E_{\phi,z} \left[ \left| v_{\Lambda,x}(\phi|z)\right|^q \right] \right)^{\frac{2}{q}}
    &\leq \frac{(L^d)^{1-\frac{2}{q}}}{\tau^2} \left(a_* \left(\mathbb E_{\phi_0,\phi,z}\left[\left\|\phi_0\right\|_{\Omega_L}^q\right]\right)^{\frac{1}{q}} + \frac{b_*}{m} \left(\mathbb E_{\phi_0,\phi,z}\left[\left\|\epsilon\right\|_{\Omega_L}^q\right]\right)^{\frac{1}{q}}\right)^2.\label{eq:vq_bound3}
\end{align}
For $2< q\leq 4$, the first term in the right-hand side can be bounded as 
\begin{align}
    \left(\mathbb E_{\phi_0,\phi,z}\left[\left\|\phi_0\right\|_{\Omega_L}^q\right]\right)^{\frac{1}{q}} 
    &\leq \left(\mathbb E_{\phi_0,\phi,z}\left[\left\|\phi_0\right\|_{\Omega_L}^4\right]\right)^{\frac{1}{4}} \\
    &\leq \left(\mathbb E_{\phi_0,\phi,z}\left[L^d\left\|\phi_0\right\|_4^4\right]\right)^{\frac{1}{4}} \\
    &= L^{\frac{d}{4}} \left(\mathbb E_{\phi_0,\phi,z} \left[ \int_{\Omega_L} d^dx\, |\phi_{0,x}|^4 \right] \right)^{\frac{1}{4}}\\
    &\leq M^{\frac{1}{4}} L^{\frac{d}{2}}, 
\end{align}
where we use the norm inequality $\left\|\phi_0\right\|_{\Omega_L} \leq L^{d/4}\left\|\phi_0\right\|_4$ and Assumption~\ref{assump:data_dist} to bound $\mathbb E_{\phi_0}\left[|\phi_{0,x}|^4\right]\leq M$ with a constant $M$.
Also, the second term can be bounded as 
\begin{align}
    \left(\mathbb E_{\phi_0,\phi,z}\left[\left\|\epsilon\right\|_{\Omega_L}^q\right]\right)^{\frac{1}{q}}
    &\leq \left(\mathbb E_{\phi_0,\phi,z}\left[\left\|\epsilon\right\|_{\Omega_L}^4\right]\right)^{\frac{1}{4}}\\
    &= \left(L^d(L^d+2)\right)^{\frac{1}{4}}.
\end{align}
Combining these bounds with Eq.~\eqref{eq:vq_bound3} and $\tau^{-1}=O(\ln L)$, we finally obtain the bound
\begin{align}
    \int_{\Omega_L} d^dx\, \left(\mathbb E_{\phi,z} \left[ \left| v_{\Lambda,x}(\phi|z)\right|^q \right] \right)^{\frac{2}{q}} 
    &\leq \gamma L^{2d\left(1-\frac{1}{q}\right)}(\ln L)^2,
\end{align}
where we absorb constants independent of $L$ and $\Lambda$ to $\gamma$.

Together with the results of Corollary~\ref{cor:LAE_cond_loc_cor}, we obtain
\begin{align}
    \Delta_{\Omega_L}(p_\Lambda, v_\Lambda; l_B)^2 
    \leq \left(\gamma_1 e^{-c_1\Lambda l_B} L^{\frac{d}{2}}\ln L\right)^2 + \left(\gamma_2 e^{-c_2\Lambda l_B} L^{d\left(1-\frac{1}{q}\right)} \ln L\right)^2,
\end{align}
where $c_1, c_2,\gamma_1, \gamma_2$ are independent of $L$ and $\Lambda$. If we take $q=3$, for example, this gives the bound in Theorem~\ref{thm:LA_cloc_data}. 

\subsection{\label{proof:thm:ode_stability}Proof of Theorem~\ref{thm:ode_stability} in the main text}
    
Let $\phi_0$ be a field configuration drawn from $p_{t=0}(\phi)$. We denote by $\phi_t$ and $\hat{\phi}_t$ the solutions obtained by evolving the same initial condition $\phi_0$ under the ODE flows~\eqref{eq:fm_ode} generated by $v_t(\phi)$ and $v_{{\rm loc},t}(\phi)$, respectively. Then, by construction, $\phi_t\sim p_t$ and $\hat{\phi}_t\sim p_t^{\rm loc}$.

Since the ODE flow can also be evolved backward in time, we may consider the map between $\phi_t\sim p_t$ and $\hat \phi_t \sim p_t^{\rm loc}$ by means of the two ODE flows, the backward ODE flow $\phi_t\to\phi_0$ and the forward ODE flow $\phi_0\to\hat\phi_t$. Since this map $\phi_t\mapsto\hat\phi_t$ defines a coupling between $p_t$ and $p_t^{\rm loc}$, the Wasserstein distance $W_2$ \cite{villani2009} can be bounded as  
\begin{align}
    W_2^2(p_t, p_t^{\rm loc}) &\leq \mathbb E_{\phi_0\sim p_0} [|| \phi_t - \hat \phi_t ||^2].\label{eq:prf_bound_2W_1}
\end{align}

To estimate the right-hand side, we define
$\delta_t=\phi_t-\hat{\phi}_t$, whose time evolution satisfies 
\begin{align}
    \frac{d\delta_t}{dt} = v_t(\phi_t) - v_{{\rm loc},t}(\hat\phi_t).
\end{align}
The norm of $\delta_t$ therefore satisfies
\begin{align}
    \frac{d}{dt} \left\|\delta_t\right\|_{\Omega_L} &\leq \left\|\frac{d\delta_t}{dt}\right\|_{\Omega_L} \\
    &= \left\| v_t(\phi_t) - v_{{\rm loc},t} (\hat\phi_t) \right\|_{\Omega_L}\\
    &\leq \left\| v_t(\phi_t) - v_{{\rm loc},t}(\phi_t) \right\|_{\Omega_L} + \left\|v_{{\rm loc},t}(\phi_t) - v_{{\rm loc},t} (\hat\phi_t) \right\|_{\Omega_L}\\
    &\leq \| v_t(\phi_t) - v_{{\rm loc},t}(\phi_t) \|_{\Omega_L} + M(t) \|\delta_t\|_{\Omega_L}.\label{eq:ddt_delta}
\end{align}
Here, we use the triangle inequality $\|a+b\|_{\Omega_L} \!\leq\! \|a\|_{\Omega_L} \!+\! \|b\|_{\Omega_L}$ and the assumption that $v_{{\rm loc},t}(\phi)$ is Lipschitz continuous with Lipschitz constant $M(t)$.

Applying the Gr\"onwall inequality~\ref{lem:gronwall} to Eq.~\eqref{eq:ddt_delta} yields the bound
\begin{align}
\left\|\delta_t\right\|_{\Omega_L}^2 
\leq \left(
    \int_0^t ds\, e^{\int_s^t dr\, M(r)} \left\|v_s(\phi_s) - v_{{\rm loc},s}(\phi_s)\right\|_{\Omega_L}
\right)^2.
\end{align}
Taking the expectation $\mathbb E_{\phi_0\sim p_{t=0}}[\cdot]$ of both sides and applying the Minkowski inequality in Lemma~\ref{lem:minkowski} with $r=2$ (we replace $f(x)$ and $g(y)$ in Lemma~\ref{lem:minkowski} by $p_0(\phi_0)$ and $e^{\int_s^t dr\, M(r)}$, respectively), we find
\begin{align}
    \left( \mathbb E_{\phi_0\sim p_0} [||\delta_t||^2] \right)^{\frac{1}{2}}
    &\leq \left(
        \int d\phi_0\, p_0(\phi_0) \left(
            \int_0^t ds\, e^{\int_s^t dr\, M(r)} ||v_s(\phi_s) - v_{{\rm loc},s}(\phi_s)||_{\Omega_L} 
        \right)^2
    \right)^{\frac{1}{2}}\\
    &\leq 
    \int_0^t ds\, e^{\int_s^t dr\, M(r)}
        \left( \int d\phi_0\, p_0(\phi_0)
            ||v_s(\phi_s) - v_{{\rm loc},s}(\phi_s)||_{\Omega_L}^2
        \right)^{\frac{1}{2}}\\
    &= 
    \int_0^t ds\, e^{\int_s^t dr\, M(r)}
        \left( \int d\phi_s\, p_s(\phi_s)
            ||v_s(\phi_s) - v_{{\rm loc},s}(\phi_s)||_{\Omega_L}^2
        \right)^{\frac{1}{2}}\\
    &= \int_0^t ds\, e^{\int_s^t dr\, M(r)} \Delta_{\Omega_L}(p_s, v_s; l_{B,s}).\label{eq:ddt_delta2}
\end{align}
In the third line, we use the fact $\phi_s\sim p_s$, and in the final line, we invoke the definition of the local approximation error $\Delta_{\Omega_L}(p_s,v_s;l_{B,s})$ in Def.~\ref{def:la_error}. Combining Eqs.~\eqref{eq:ddt_delta2} and \eqref{eq:prf_bound_2W_1}, we obtain the desired bound on the 2-Wasserstein distance between $p_t$ and $p_t^{\rm loc}$ in Theorem~\ref{thm:ode_stability}, completing the proof.

\subsection{\label{proof:prop:equiv_sde}Proof of Proposition~\ref{prop:equiv_sde} in the main text}

The marginal distribution associated with the Langevin-type SDE~\eqref{eq:sde1} satisfies the Fokker--Planck equation
\begin{align}
    \partial_t p_t(\phi) 
    &= - \frac{\delta}{\delta\phi} \!\cdot\! \left[p_t\left(v_t \!+\! \frac{g(t)^2}{2} \frac{\delta \ln p_t}{\delta\phi}\right)\right] \!+\! \frac{g(t)^2}{2} \frac{\delta^2 p_t}{\delta\phi^2}\\
    &= - \frac{\delta}{\delta\phi} \!\cdot\! (p_tv_t).
\end{align}
Here, we use $p_t(\delta\ln p_t/\delta\phi)=\delta p_t/\delta\phi$ in the second equality. The resulting equation coincides with the continuity equation for the ODE flow generated by $v_t$. Therefore, the SDE~\eqref{eq:sde1} generates the same time evolution of the marginal distribution $p_t$ as the original ODE flow.

\subsection{\label{proof:thm:sde_stability}Proof of Theorem~\ref{thm:sde_stability} in the main text}

As shown in the proof of Proposition~\ref{prop:equiv_sde}, the marginal distribution $p_t$ satisfies the continuity equation
\begin{align}
    \partial_t p_t(\phi) = - \frac{\delta}{\delta\phi} \cdot (p_t(\phi) v_t(\phi)).
\end{align}
On the other hand, the marginal distribution $\hat p_t^{\rm loc}$ satisfies the Fokker--Planck equation
\begin{align}
    \partial_t \hat p_t^{\rm loc}(\phi) 
    &= -\frac{\delta}{\delta\phi} \!\cdot\! \left[\hat p_t^{\rm loc}\!\left(v_{{\rm loc},t} \!+\! \frac{g(t)^2s_{{\rm loc},t}}{2}\right)\right] \!+\! \frac{g(t)^2}{2} \frac{\delta^2 \hat p_t^{\rm loc}}{\delta\phi^2}.
\end{align}
Using these equations and assuming that the boundary terms arising from integration by parts vanish, the time derivative of the KL divergence is expressed as
\begin{align}
    \frac{d}{dt}D_{\rm KL}(p_t||\hat p_t^{\rm loc}) 
    &= \mathbb E_{p_t} \left[ \left(
        v_t - v_{{\rm loc},t} + \frac{g(t)^2}{2}(s_t - s_{{\rm loc},t})\right)\cdot\frac{\delta}{\delta\phi} \ln \frac{p_t}{\hat p_t^{\rm loc}} \right] 
        - \frac{g(t)^2}{2} \mathbb E_{p_t}\left[\left\| \frac{\delta}{\delta\phi} \ln \frac{p_t}{\hat p_t^{\rm loc}} \right\|^2\right].\label{eq:sde3}
\end{align}
Applying Young's inequality,
$A\cdot G\leq |A|^2/[2g(t)^2]+g(t)^2|G|^2/2$,
with
$A=v_t-v_{{\rm loc},t}+g(t)^2(s_t-s_{{\rm loc},t})/2$
and
$G=(\delta/\delta\phi)\ln(p_t/\hat p_t^{\rm loc})$,
we obtain
\begin{align}
    \frac{d}{dt}D_{\rm KL}(p_t||\hat p_t^{\rm loc})
    &\leq \frac{1}{2g(t)^2}\mathbb E_{p_t} \left[
        \left\|
            v_t - v_{{\rm loc},t} + \frac{g(t)^2}{2}(s_t - s_{{\rm loc},t})
        \right\|^2\right].\label{eq:sde4}
\end{align}
We further use $|a+b|^2\leq 2|a|^2+2|b|^2$ to find
\begin{align}
    \frac{d}{dt}D_{\rm KL}(p_t||\hat p_t^{\rm loc})
    &\leq \frac{1}{g(t)^2} \Delta^2_{\Omega_L}(p_t,v_t;l_{B,t}) + \frac{g(t)^2}{4} \Delta^2_{\Omega_L}(p_t, s_t; l_{B,t})\\
    &\leq \Delta_{\Omega_L}(p_t,v_t;l_{B,t}) \Delta_{\Omega_L}(p_t, s_t; l_{B,t}).
    \label{eq:sde5}
\end{align}
where we minimized the right-hand side by taking 
\begin{align}
    g^2(t) = \frac{2\Delta_{\Omega_L}(p_t,v_t;l_{B,t})}{\Delta_{\Omega_L}(p_t, s_t; l_{B,t})}.\label{eq:am-gm-gt}
\end{align}
Since $p_0=\hat p_0^{\rm loc}$, we have
$D_{\rm KL}(p_0||\hat p_0^{\rm loc})=0$.
Integrating Eq.~\eqref{eq:sde5} over time therefore gives the bound in Eq.~\eqref{eq:sde2}. \hfill$\square$

\subsection{\label{proof:thm:rgfm_flow_stability_sde}Proof of Theorem~\ref{thm:rgfm_flow_stability_sde} in the main text}
In proving Theorem~\ref{thm:sde_stability}, we choose the diffusion coefficient $g(t)$ as in Eq.~\eqref{eq:am-gm-gt}. In the case of RGFM, the local approximation errors of $s_t$ and $v_t$ are related to each other by Eq.~\eqref{eq:st_vt_rgfm} in the main text. Therefore, by choosing the diffusion coefficient $g(t)$ as 
\begin{align}
    g(t) = \sqrt{\frac{2}{\Lambda(t)^2\tau}},
\end{align}
we obtain the bound 
\begin{align}
    D_{\rm KL}\left(p_{\rm data}||\hat p_{\rm data}^{\rm loc}\right)
    &\leq \tau \int_0^1 dr\, \Lambda(r)^2 \Delta_{\Omega_L}^2(p_r,v_r;l_{B,r})\\
    &\leq \tau \Lambda_{\rm UV}^2 \int_0^1 dr\, \Delta_{\Omega_L}^2(p_r,v_r;l_{B,r}).
\end{align}
From Theorem~\ref{thm:LA_rgfm}, for any prescribed accuracy $\varepsilon>0$, the local approximation error $\Delta_{\Omega_L}^2(p_t,v_t;l_{B,t})$ can be made arbitrarily small with local patches of linear size $O(\Lambda(t)^{-1}\ln(L^\alpha/\sqrt{\varepsilon}))$. This proves Theorem~\ref{thm:rgfm_flow_stability_sde}.

\twocolumngrid

\section{\label{app:num_exp}Details of the numerical experiments}
In this Appendix, we provide details of the numerical experiments presented in Sec.~\ref{sec:num_exp}. In the numerical experiments, we set a=1, as described in the main text, and use the discrete cosine transform (DCT) with orthonormal normalization. Also, samples are generated by numerically integrating the learned ODE flow using the midpoint method implemented in the \texttt{odeint} function of the \texttt{torchdiffeq} package in PyTorch.

\subsection{One-dimensional experiments}

\subsubsection{Neural-network architecture and optimization}

For both the one-dimensional Ising model in Sec.~\ref{ssec:ne_ising} and the conditionally local distribution in Sec.~\ref{ssec:ne_cond_loc}, we use the same local convolutional neural network. The network contains $N_{\rm layer}=6$ spatial convolutional layers with kernel size $H=5$ and hidden-channel dimension $64$, followed by a $1\times1$ output convolution. The first convolution maps the single input channel to $64$ hidden channels and is followed by five residual convolutional blocks. Each residual block consists of group normalization with eight groups, a SiLU activation, and a convolution with $64$ input and output channels. All convolutions with kernel size $H=5$ use reflection padding so that the spatial size is preserved. The output layer consists of group normalization, a SiLU activation, and a $1\times1$ convolution that maps the hidden representation back to a single output channel. The resulting receptive field contains
\begin{align}
    1+N_{\rm layer}(H-1)=25
\end{align}
lattice sites, corresponding to $l_A=1$ and $l_B=12$ in the notation of Sec.~\ref{sec:loc_app}.

The time variable is encoded using a $64$-dimensional sinusoidal embedding, followed by a two-layer multilayer perceptron with hidden dimension $64$ and a SiLU activation. The resulting time embedding is added to the hidden representation after the first convolution. The same architecture is used for RGFM and the standard FM.

We train the models using the Adam optimizer with learning rate $2\times10^{-4}$. The learning rate is increased linearly during the first $5{,}000$ optimization steps, and the gradient norm is clipped at $1$. For evaluation and sampling, we use an exponential moving average of the model parameters with decay rate $0.995$.

\subsubsection{RGFM scale decomposition}

For both RGFM experiments, the original lattice has size $L_0=1024$ and is successively decimated as
\begin{align}
    L=1024,\ 512,\ 256,\ 128,\ 64,\ 32,\ 16.
\end{align}
We choose the scale-transition times from the running RG scale $\Lambda(t)$ so that the modes discarded at each transition have already entered the Gaussian sector of the RGFM distribution. More specifically, the transition time $t_i$ from $L_{i-1}$ to $L_i$ is determined by
\begin{align}
    K_{t_i,L_i\pi/L_0}=K_{1,\pi/L_0},
\end{align}
where $K_{tk}$ is the RG cutoff introduced in Eq.~\eqref{eq:eff_action_erg}. The transition time in the numerical experiments is given by 
\begin{align}
    t_i \simeq 0.41,\ 0.50,\ 0.59,\ 0.69,\ 0.78,\ 0.87.
\end{align} 
A separate copy of the same local CNN is trained on each time interval between successive scale transitions.

In the one-dimensional RGFM experiments, we set the mass parameter in the RG cutoff $K_\Lambda(k)$ in Eq.~\eqref{eq:K_Lam} to $m=0.02$.
We parameterize the running RG scale as $\Lambda_t=\Lambda_0 e^{-t/\tau}$ and choose $\Lambda_0$ and $\tau$ such that
\begin{align}
    \bar\beta_{t=0,k_{\rm max}=\pi} &\ll 1,\\
    \bar\alpha_{t=1,k_{\rm min}=\pi/L_0} &\ll 1.
\end{align}
These conditions ensure that $p_{t=0}\simeq p_{\rm data}$ and $p_{t=1}\simeq p_{\rm GS}$. For both one-dimensional datasets, we use
\begin{align}
    \Lambda_0 &\simeq 10.22,\\
    \tau &\simeq 0.136,
\end{align}
which gives $\bar\beta_{t=0,k_{\rm max}=\pi}\simeq3.7\times10^{-6}$ and $\bar\alpha_{t=1,k_{\rm min}}\simeq 7.1\times10^{-5}$. Also, in numerical experiments, we take $G_0(k)$ in the RG diffusion~\eqref{eq:rg_diffusion} as $m_0^2/(k^2+m^2)$ with $m_0=\sqrt{(\pi/L)^2+m^2}$ so that the mode with $k=\pi/L$ has ${\rm Var}[\phi_k]\simeq 1$ in $p_{\rm GS}$.

We use $100{,}000$ optimization steps for the first and last scale intervals and $50{,}000$ steps for each of the five intermediate intervals, corresponding to $4.5\times10^5$ optimization steps in total. The batch sizes for $L=1024,512,256,128,64,32,16$ are, respectively,
\begin{align}
    100,\ 200,\ 400,\ 800,\ 800,\ 800,\ 800.
\end{align}
The training takes approximately three hours in total on a single NVIDIA RTX 6000 Ada GPU with 48 GB of memory. During sampling, the corresponding scale intervals are integrated using
\begin{align}
    50,\ 20,\ 20,\ 20,\ 20,\ 20,\ 50
\end{align}
midpoint steps, giving $200$ integration steps over the complete probability path. At each scale transition in the reverse generative process, the retained low-wavenumber DCT coefficients are lifted to the finer lattice, while the missing high-wavenumber coefficients are independently sampled from the Gaussian sector, as described in Sec.~\ref{sec:lgm_with_rgfm}.

For local generative modeling with the standard FM, we use the same neural-network architecture and train a single model on the $L=1024$ lattice for $5\times10^5$ optimization steps with batch size $100$. Samples are generated by integrating the ODE over $t\in[0,1]$ using $200$ equally spaced midpoint steps.

\subsubsection{Dataset-specific details and evaluation}

For the Ising-model experiment, we evaluate the two-point correlation function using $10{,}000$ generated samples and compare it with the exact expression in Eq.~\eqref{eq:ising_corr}, as shown in Fig.~\ref{fig:num_ising}(c).

The conditionally local dataset is generated directly from Eq.~\eqref{eq:cond_local_dist}.
For each sample, we independently draw $a_m\sim\mathcal N(0,1)$ and $\eta_m\sim{\rm Uniform}[0,2\pi)$ for $m=1,2,3$, construct the smooth waveform $f_z(x)$, and add independent Gaussian fluctuations with standard deviation $\sigma=0.05$ at each site. For the correlation-function evaluation shown in Fig.~\ref{fig:num_cld}(b), we generate $10{,}000$ samples and compare the resulting correlation function with the exact expression in Eq.~\eqref{eq:cond_loc_cor}.

\subsection{Image-generation experiments}

\subsubsection{Dataset and preprocessing}

We use images from the FFHQ dataset. For the $64\times64$ experiments, the original FFHQ images are resized to $64\times64$ pixels and their RGB values are normalized to the interval $[-1,1]$. For the $256\times256$ experiment, we use the first $5{,}000$ FFHQ images resized to $256\times256$ pixels and apply the same normalization.

\subsubsection{Local patches and positional channels}

At each scale, the image is divided into nonoverlapping target patches of linear size $l_A$, and the velocity on each target patch is predicted from a larger receptive patch of linear size $l_A+2l_B$. The stride is set equal to the target-patch size. Away from the image boundary, the target patch is centered inside the receptive patch. Near the boundary, the receptive patch is shifted so that it remains entirely inside the image.

Because the image distribution is not translationally invariant, we additionally provide the absolute pixel coordinates to the local model. For a lattice of linear size $L$, the horizontal and vertical coordinates are normalized to $[-1,1]$ according to
\begin{align}
    \phi^X_{ij}=\frac{2j}{L-1}-1,
    \qquad
    \phi^Y_{ij}=\frac{2i}{L-1}-1.
\end{align}
Together with the RGB channels, the input to the U-Net is therefore the five-channel field $(\phi^R,\phi^G,\phi^B,\phi^X,\phi^Y)$. The network predicts three output channels corresponding to the RGB components of the velocity field. The flow-matching loss is evaluated only on the target region $A$ inside each receptive patch.

Throughout the image experiments, the \emph{batch size} refers to the number of full images included in one optimization batch. Each full image is then decomposed into multiple local target patches before being passed to the U-Net. Thus, if a lattice of size $L\times L$ is divided into nonoverlapping target patches of size $l_A\times l_A$, one image produces $(L/l_A)^2$ local patches, and a batch of $B$ images provides
\begin{align}
    B\left(\frac{L}{l_A}\right)^2
\end{align}
local training patches to the U-Net. We use this distinction below when specifying the training batch sizes.

\subsubsection{U-Net architecture and optimization}

For both RGFM and FM, we use the same U-Net architecture. The base channel dimension is $128$, the channel multipliers are $[1,2,2,2]$, and two residual blocks are used at each resolution. Self-attention is included at the second resolution level of the U-Net, corresponding to a spatial resolution reduced by a factor of two from the input patch size, as well as in the middle block. The dropout probability is set to $0.1$. The input and output channel dimensions are $5$ and $3$, respectively.

We train the image models using Adam with learning rate $2\times10^{-4}$. The learning rate is increased linearly during the first $5{,}000$ optimization steps, the gradient norm is clipped at $1$, and an exponential moving average with decay rate $0.9999$ is used for sampling and evaluation.

\subsubsection{RG parameters}

In the image-generation experiments, we set the mass parameter in the RG cutoff $K_\Lambda(k)$ in Eq.~\eqref{eq:K_Lam} to $m=\pi/40$ for the $64\times64$ experiments and to $m=\pi/160$ for the $256\times256$ experiment. As in the one-dimensional experiments, we parameterize the RG scale as $\Lambda_t=\Lambda_0e^{-t/\tau}$ and choose the scale-transition times using the criterion described above. Also, in numerical experiments, we take $G_0(k)$ in the RG diffusion~\eqref{eq:rg_diffusion} as $m_0^2/(k^2+m^2)$ with $m_0=3\pi/20$ for both resolutions. The remaining RG parameters and the corresponding endpoint values of $\bar\alpha$ and $\bar\beta$ are summarized in Table~\ref{tab:rg_parameters_img}.

\begin{table}[t]
    \centering
    \caption{\label{tab:rg_parameters_img}
    Parameters of the RG schedule used in the image-generation experiments.
    }
    \begin{tabular}{c || c | c | c | c | c}
        \hline\hline
        Resolution & $m$ & $\Lambda_0$ & $\tau$ &
        $\bar\beta_{0,k_{\rm max}}$ & $\bar\alpha_{1,k_{\rm min}}$ \\
        \hline
        $64\times64$   & $\pi/40$  & $87.42$ & $0.1224$ & $2.9\times 10^{-5}$ & $8.2\times 10^{-7}$ \\
        $256\times256$ & $\pi/160$ & $21.90$ & $0.1224$ & $4.5\times 10^{-4}$ & $8.7\times 10^{-7}$ \\
        \hline\hline
    \end{tabular}
\end{table}

\subsubsection{\texorpdfstring{$64\times64$}{64x64} images}

For the representative $64\times64$ experiment in Fig.~\ref{fig:num_img}(a), the target-patch size is fixed at $l_A=16$, and the receptive-patch size is $l_A+2l_B=32$. For RGFM, we use three successive lattice sizes,
\begin{align}
    L=64,\ 32,\ 16,
\end{align}
with target-patch size $16$ at all scales and receptive-patch sizes $32$, $32$, and $16$, respectively. The scale-transition times are determined from the RG cutoff using the same criterion as in the one-dimensional experiments. The transition time in the numerical experiments is given by 
\begin{align}
    t_i \simeq 0.63,\ 0.72.
\end{align} 

The three U-Nets are trained for approximately $6\times10^5$, $2.5\times10^5$, and $1.5\times10^5$ optimization steps, respectively, corresponding to $10^6$ optimization steps in total. The numbers of full images per optimization batch are $16$, $64$, and $128$ for $L=64,32,16$, respectively. Since $l_A=16$, each full image at these resolutions produces $16$, $4$, and $1$ local target patches, respectively. The corresponding numbers of local patches supplied to the U-Net in one optimization batch are therefore $256$, $256$, and $128$. In our numerical experiments, the training takes approximately three days in total on a single NVIDIA RTX PRO 6000 Blackwell Max-Q GPU with 96 GB of memory. During generation, the three scale intervals are integrated using $160$, $40$, and $100$ midpoint steps, respectively, giving $300$ steps over the complete probability path.

The standard local FM baseline is trained directly at $64\times64$ resolution using the same U-Net architecture, target-patch size, and receptive-patch construction. For the model with receptive-patch size $32$, we use $10^6$ optimization steps and a batch size of $8$ full images, corresponding to $128$ local patches per optimization batch.

To obtain Fig.~\ref{fig:num_img}(b), we fix the target-patch size at $l_A=16$ and vary the receptive-patch size over
\begin{align}
    l_A+2l_B=32,\ 48,\ 64.
\end{align}
For all three receptive-patch sizes, the U-Net architecture and target-patch geometry are kept fixed. At the finest RGFM scale, $L=64$, the models with receptive-patch sizes $48$ and $64$ are trained with batch sizes of $8$ and $4$ full images, respectively, corresponding to $128$ and $64$ local patches per optimization batch. Both models are trained for $6\times10^5$ optimization steps at this scale. At the coarser scales, $L=32$ and $16$, we reuse the U-Nets trained for receptive-patch size $l_A+2l_B=32$.

\subsubsection{\texorpdfstring{$256\times256$}{256x256} images}

For the $256\times256$ experiment in Fig.~\ref{fig:num_img}(c), the RGFM generative path uses
\begin{align}
    L=256,\ 128,\ 64,\ 32,\ 16.
\end{align}
The target-patch size is fixed at $l_A=16$. The receptive-patch size is $32$ for $L=256,128,64,32$ and $16$ for the final $L=16$ scale. The transition time in the numerical experiments is given by 
\begin{align}
    t_i \simeq 0.46,\ 0.55,\ 0.63,\ 0.72.
\end{align} 

The corresponding RGFM models are trained for approximately
\begin{align}
    6\times10^5,\ 6\times10^5,\ 5\times10^5,\ 3\times10^5,\ 3\times10^5
\end{align}
optimization steps, corresponding to $2.3\times10^6$ optimization steps in total. In our numerical experiments, the training takes approximately six days on a single NVIDIA RTX PRO 6000 Blackwell Max-Q GPU with 96 GB of memory. The batch sizes, measured in numbers of full images, are
\begin{align}
    1,\ 4,\ 16,\ 64,\ 64,
\end{align}
respectively. For $l_A=16$, these correspond to
\begin{align}
    256,\ 256,\ 256,\ 256,\ 64
\end{align}
local patches supplied to the U-Net per optimization batch. During generation, the five scale intervals are integrated using
\begin{align}
    100,\ 50,\ 50,\ 50,\ 50
\end{align}
midpoint steps, respectively, giving $300$ integration steps in total. These settings are summarized in Table~\ref{tab:rgfm_256_settings}.

\begin{table}[t]
    \centering
    \caption{\label{tab:rgfm_256_settings}
    Training and sampling settings for the $256\times256$ RGFM experiment.
    The batch size denotes the number of full images, while the number of local patches counts the inputs supplied to the U-Net in one optimization batch.
    }
    \begin{tabular}{c || c | c | c | c}
        \hline\hline
        $L$ & Receptive size & Training steps & Batch size & Local patches \\
        \hline
        $256$ & $32$ & $6\times10^5$ & $1$  & $256$ \\
        $128$ & $32$ & $6\times10^5$ & $4$  & $256$ \\
        $64$  & $32$ & $5\times10^5$ & $16$ & $256$ \\
        $32$  & $32$ & $3\times10^5$ & $64$ & $256$ \\
        $16$  & $16$ & $3\times10^5$ & $64$ & $64$ \\
        \hline\hline
    \end{tabular}
\end{table}

The local FM baseline uses the same target- and receptive-patch sizes on the original $256\times256$ lattice.
We train the U-Net for the same total number of optimization steps, $2.3\times10^6$, using a batch size of one full image.
Since one $256\times256$ image contains $256$ nonoverlapping $16\times16$ target patches, each optimization batch therefore supplies $256$ local patches to the U-Net.
Samples are generated using $300$ equally spaced midpoint steps over $t\in[0,1]$.

\subsubsection{FID evaluation}

We evaluate the $64\times64$ image models using the Fr\'echet inception distance (FID) \cite{heusel2017a}.
The real-data statistics are computed from $50{,}000$ FFHQ images using the clean-FID preprocessing convention \cite{parmar2022}.
Both real and generated images are first mapped from $[-1,1]$ to $8$-bit RGB values in $[0,255]$ and resized to $299\times299$ pixels using the clean-FID resizer before the Inception features are extracted.
We then compute the empirical feature means and covariance matrices and evaluate the standard Fr\'echet distance between the generated and real distributions.

\bibliography{rgfm_refs}

\end{document}